\documentclass{article}

\PassOptionsToPackage{numbers, compress}{natbib}

\usepackage[main, final]{neurips_2026}

\usepackage[utf8]{inputenc} %
\usepackage[T1]{fontenc}    %
\usepackage{hyperref}       %
\usepackage{url}            %
\usepackage{booktabs}       %
\usepackage{amsfonts}       %
\usepackage{nicefrac}       %
\usepackage{microtype}      %
\usepackage{xcolor}         %

\usepackage{amsmath}
\usepackage{amssymb}
\usepackage{graphicx}
\usepackage{algorithm}
\usepackage{algorithmic}
\usepackage{multirow}
\usepackage{subcaption}
\usepackage{wrapfig}
\usepackage{amsthm}
\usepackage{tikz}
\usepackage{arydshln}
\usetikzlibrary{arrows.meta, positioning, fit, backgrounds, calc, decorations.pathreplacing}

\newtheorem{proposition}{Proposition}

\usepackage{enumitem}

\title{Neural Dual Bounds: Valid-by-Construction JGLP Warm-Starts for MAP and Constrained MAP}

\author{%
  Akshay Vyas \\
  Department of Computer Science \\
  The University of Texas at Dallas \\
  \texttt{akshay.vyas@utdallas.edu} \\
  \And
  Shivvrat Arya \\
  Department of Computer Science \\
  New Jersey Institute of Technology \\
  \texttt{shivvrat.arya@njit.edu} \\
  \AND
  Brij Malhotra \\
  Department of Computer Science \\
  The University of Texas at Dallas \\
  \texttt{brijgulsharan.malhotra@utdallas.edu} \\
  \And
  Tahrima Rahman \\
  Department of Computer Science \\
  The University of Texas at Dallas \\
  \texttt{tahrima.rahman@utdallas.edu} \\
  \AND
  Vibhav Gogate \\
  Department of Computer Science \\
  The University of Texas at Dallas \\
  \texttt{vibhav.gogate@utdallas.edu} \\
  \And
  Nicholas Ruozzi \\
  Department of Computer Science \\
  The University of Texas at Dallas \\
  \texttt{nicholas.ruozzi@utdallas.edu}
}

\begin{document}

\maketitle

\begin{abstract}

We present Neural Dual Bounds, a calibration-free framework that produces deterministic upper bounds for MAP inference in probabilistic graphical models using a single neural-network forward pass. The framework trains a network to output the edge messages of a mini-bucket style upper bound on a given join-graph decomposition in a self-supervised manner by directly minimizing a smooth dual objective.  The output of this network can be used at test time to warm-start a standard JGLP optimizer. Across diverse benchmarks, this warm start yields dual gaps that are orders of magnitude tighter than those from zero initialization, with the largest gains in the early-iteration regime. We also apply our framework for the constrained MAP (CMPE) task, where neural warm starts with test-time adaptation are comparable to the adaptive baseline on Bayesian networks and achieve the tightest dual gaps, by one to two orders of magnitude, on the grid and segmentation families.
\end{abstract}

\section{Introduction}

Many applications of probabilistic graphical models (PGMs) repeatedly solve closely related inference problems on a fixed graph. Branch-and-bound search expands one node per query, adversarial robustness certification verifies one input per query, constrained planning evaluates one constraint per query, and bound-based parameter learning computes a new bound at each gradient step. In these settings, per-query inference often dominates wall-clock time. Classical polynomial-time dual bounds, including join-graph linear programming (JGLP)~\citep{ihler2012join}, Max-Product Linear Programming (MPLP)~\citep{globerson2007fixing}, and TRW-BP~\citep{wainwright2005map}, provide valid upper bounds, but often require hundreds of coordinate-descent iterations per query. This repeated-query structure creates a natural opportunity for amortization, provided that amortization preserves the validity guarantees required by downstream algorithms.

Learned methods can amortize inference, but they typically sacrifice deterministic per-query validity. Learned bound predictors usually provide only statistical validity: if a predicted upper bound violates the true upper-bound property on even a small fraction of queries, branch-and-bound may prune the optimal solution and robustness certificates may become invalid~\citep{nair2020solving,gasse2019exact,khalil2017learning}. Conformal calibration~\citep{vovk2005algorithmic,angelopoulos2023conformal} can reduce this risk by adding a safety margin estimated from held-out data, but it still permits violations at the target residual failure rate. Learned inference networks based on graph neural networks~\citep{satorras2021neural,yoon2019inference,zhang2020factor} predict approximate inference outputs, but do not enforce an upper-bound certificate for every query. Existing learned methods therefore accelerate inference by weakening the certificate that makes classical dual bounds useful in exact and safety-critical pipelines.

This validity-speed tradeoff is especially important for constrained MAP inference, which seeks the most probable explanation subject to one or more budgets defined by a second log-linear model. This formulation extends standard MAP inference to settings with explicit side constraints and captures safety, fairness, adversarial, and resource-constrained reasoning problems~\citep{NEURIPS2020_8ab9bb97,rahman2021novel}. Existing upper-bounding methods for this setting optimize a single scalar Lagrange multiplier by subgradient descent~\citep{rahman2021novel}. They do not directly handle the vector-valued multipliers induced by multiple assurance requirements. With multiple constraints, projected gradient descent over the multiplier vector must solve an inner MAP-bound problem at every step, while queries may vary in evidence, costs, constraint budgets, or Lagrange multipliers. This nested optimization makes valid amortized bounds particularly valuable in the multi-constraint setting.

We introduce \emph{Neural Dual Bounds}, a framework for amortized inference that produces deterministically valid upper bounds from a single neural-network forward pass. Instead of learning bound values directly, the network outputs antisymmetric edge messages on a join-graph decomposition. This parameterization restricts every output to a feasible subspace of the JGLP bound, so the resulting bound is valid by construction rather than by training accuracy, calibration, or a held-out quantile estimate. Antisymmetric messages preserve feasibility because each message added to one side of an edge is subtracted from the other, leaving the decomposed upper-bound objective unchanged while reparameterizing local potentials. The guarantee holds for trained, untrained, and adversarial network outputs.

The same construction also provides effective warm-starts for iterative dual optimization. For MAP inference, we train the model end-to-end against a smooth objective, without an LP solver in the inner loop and without JGLP-converged labels. For constrained MAP inference, we supervise the network with LP-optimal edge messages for the per-$\mu$ Lagrangians, since this supervision is cheaper than self-supervised training to comparable tightness in this setting. We also show that an LP-free training variant can achieve similar bound quality. Thus, the framework supports both direct amortized certification and accelerated iterative tightening.
\paragraph{Contributions.} 
\begin{enumerate}[leftmargin=*,nosep] 
\item \textbf{Deterministically valid neural bounds.} We restrict the neural-network output to a structurally valid subspace of the JGLP upper bounds using an antisymmetric scatter map. This construction yields, to our knowledge, the first learned MAP upper-bound predictor whose validity is enforced by architecture rather than by post-hoc calibration.
\item \textbf{Self-supervised dual amortization without LP labels.} We train the model end-to-end against a smooth dual objective, with no LP solver in the inner training loop and no JGLP-converged labels. This objective removes the need to generate optimal targets for MAP training and makes the method applicable when LP solves are unavailable or prohibitively expensive.
\item \textbf{Extension to multi-constraint constrained MAP inference.} We apply the framework to vector-valued Lagrange multipliers using projected gradient descent, neural warm-starts, and replay-buffer test-time adaptation at each step. %
\item \textbf{Empirical validation on benchmark PGMs.} On 15 UAI benchmark PGMs spanning Bayesian networks, image segmentation, and grid Markov random fields,
neural warm-starts perform comparably to zero-initialized JGLP in the worst case and up to six orders of magnitude tighter in the best case, with the largest gains in the early-iteration regime. On constrained MAP inference, replay-buffer test-time adaptation gives the tightest dual gaps on the grid and segmentation families at every iteration budget, while on bayesian networks it is matched by a no-learning baseline that simply reuses the previous multiplier step's converged solution.

\item \textbf{Structural applicability beyond JGLP.} We identify a broader class of decomposed Lagrangian duals whose message structure admits the same antisymmetric parameterization, including dual decomposition, TRW-BP, weighted CSPs, and minimum-cost flow.
\end{enumerate}

\section{Background}
\label{sec:background}
A probabilistic graphical model $(G,\mathcal{F}, \theta)$ defines a factored distribution of $n$ discrete variables $\mathbf{x} = (x_1, \ldots, x_n)$ with domains $\mathcal{D}_i$ over a graph $G=(V,E)$ with cliques $\mathcal{F}$ as
\begin{equation}
    p(\mathbf{x} \mid \theta) \propto \prod_{\alpha \in \mathcal{F}} \exp(\theta_\alpha(\mathbf{x}_\alpha))%
\end{equation}
where $\mathcal{F}$ indexes cliques of $G$ with scopes $\mathbf{x}_\alpha \subseteq \mathbf{x}$ and $\theta_\alpha:D_\alpha \rightarrow \mathbb{R}$ are known as the log-potential functions.  

The MAP problem is to find an assignment that maximizes the probability distribution $p$ or, equivalently, maximizes the $\log$ probability
\begin{equation}
    \text{MAP}^*(\boldsymbol{\theta}) = \max_{\mathbf{x}} \sum_{\alpha \in \mathcal{F}} \theta_\alpha(\mathbf{x}_\alpha).
    \label{eq:map}
\end{equation}

Exact MAP is NP-hard in general~\citep{shimony1994finding}, and
standard methods designed for this task, such as variable elimination, scale exponentially in
the tree-width of $G$. As a result, a variety of approximation methods have been designed that approximately solve the MAP problem via upper bound minimization.  The
Mini-bucket
elimination~(MBE,~\citealp{dechter2003mini}), one such upper bound minimization approach, further regroups the clique potentials into clusters determined by a specified \emph{$i$-bound} parameter.
Specifically, the procedure induces a
\emph{join graph} $\mathcal{G} = (\mathcal{C},
\mathcal{E}, \theta)$ whose nodes $\mathcal{C} = \{C_1, \ldots, C_{|\mathcal{C}|}\}$ 
 correspond to a subset of variables $\alpha\subseteq V$ with $|\alpha| \leq i+1$ and edges in $\mathcal{E}$ connect
clusters that share variables. For each edge $(\alpha, \beta) \in
\mathcal{E}$, the \emph{separator} $S_{\alpha\beta} = C_\alpha \cap
C_\beta$ is the shared subset. Each cluster carries a log-potential
function $\theta_{C_\alpha}$ that aggregates the original log-potentials
assigned to it during partitioning.  MBE then uses the join graph clustering to produce an upper bound on the MAP value as
\begin{equation}
D_0 = \sum_{\alpha=1}^{|\mathcal{C}|} \max_{\mathbf{x}_{C_\alpha}}
\theta_{C_\alpha}(\mathbf{x}_{C_\alpha}) \;\geq\; \mathrm{MAP}^*,
\label{eq:mbe_bound}
\end{equation}
where the inequality comes from the observation that a maximum of a sum is always less than or equal to the sum of maximums, i.e., $\max_x (f(x) + g(x)) \leq \max_x f(x) + \max_x g(x)$.

The bound~\eqref{eq:mbe_bound} is loose because adjacent clusters
maximize independently and may pick conflicting values on their
shared separator. \emph{Reparameterization} tightens the bound by
shifting cost between adjacent clusters via edge messages
$\lambda_{\alpha\beta}(\mathbf{x}_S) \in \mathbb{R}$:
\begin{equation}
\theta_{C_\alpha}(\mathbf{x}_{C_\alpha},\lambda) =
\theta_{C_\alpha}(\mathbf{x}_{C_\alpha}) + \sum_{\beta \in N(\alpha)}
\lambda_{\alpha\beta}(\mathbf{x}_{S_{\alpha\beta}}),
\label{eq:reparam}
\end{equation}
subject to an \emph{antisymmetry constraint}
\begin{equation}
\lambda_{\alpha\beta}(\mathbf{x}_S) +
\lambda_{\beta\alpha}(\mathbf{x}_S) = 0,\qquad
\forall (\alpha, \beta) \in \mathcal{E},\;\forall \mathbf{x}_S.
\label{eq:antisymmetry}
\end{equation}
Antisymmetry preserves the underlying joint: for any complete
assignment $\mathbf{x}^*$, the message added to one cluster on a
shared separator is exactly the message subtracted from its
neighbor, so $\sum_\alpha \theta_{C_\alpha}(\mathbf{x}^*_{C_\alpha},\lambda)
= \sum_\alpha \theta_{C_\alpha}(\mathbf{x}^*_{C_\alpha})$. Replacing
the assignment-specific sum by per-cluster maxes can only increase
the value, so the reparameterized bound
\begin{equation}
D(\boldsymbol\theta, \boldsymbol\lambda) =
\sum_{\alpha=1}^{|\mathcal{C}|} \max_{\mathbf{x}_{C_\alpha}}
\theta_{C_\alpha}(\mathbf{x}_{C_\alpha},\lambda)
\;\geq\; \mathrm{MAP}^*
\label{eq:dual}
\end{equation}
holds for every $\boldsymbol\lambda$ in the antisymmetric subspace.

As  $D(\boldsymbol\theta, \boldsymbol\lambda)$ is convex in
$\boldsymbol\lambda$ (a sum of pointwise maxima of affine functions), a variety of techniques can now be applied to find the minimum of this upper bound.  The JGLP~\citep{ihler2012join} method minimizes it by coordinate descent.
Each iteration of JGLP corresponds to one pass over the separator edges: for
$(\alpha, \beta) \in \mathcal{E}$, the algorithm computes the
max-marginals of $C_\alpha$ and $C_\beta$ onto $S_{\alpha\beta}$,
equalizes them to their geometric mean (the arithmetic mean in log
domain), and applies the corresponding antisymmetric correction to
both cluster tables (\citet{ihler2012join} Algorithm~4, lines~5--8).
The fixed point is the optimum $\min_{\boldsymbol\lambda} D(\boldsymbol\theta, \boldsymbol\lambda)$, which corresponds to the optimum of a specific LP relaxation of the MAP problem on the chosen join graph. 

JGLP starts from a \emph{moment-matched} mini-bucket join graph
rather than raw MBE: at construction, one round of cost-shifting on
the mini-buckets within each elimination bucket equalizes their
max-marginals on the shared scope (\citet{ihler2012join}
Algorithm~5, ``MBE-MM''). This produces a slightly tighter starting
point than~\eqref{eq:mbe_bound} before any JGLP iterations have run.
Throughout this paper, ``$\boldsymbol\lambda = \mathbf{0}$''
denotes the JGLP coordinate-descent state at iteration zero
\emph{after} MBE-MM construction, matching the convention of all
JGLP implementations we benchmark against.

\section{Method}
\label{sec:method}

Reparameterizations of a join-graph
decomposition~\citep{ihler2012join,wainwright2005map} are constrained
to lie in the antisymmetric subspace defined
by~\eqref{eq:antisymmetry}; any element of this subspace yields a
valid bound. Existing methods (JGLP, MPLP, TRW-BP) navigate this
subspace iteratively. However, convergence of JGLP often
requires hundreds of iterations on the benchmark models we consider, with
each iteration costing $\mathcal{O}(|\mathcal{E}| \cdot d^{\,i+1})$, where $d$ is the maximum domain size.  As a result, a poor initialization can be far from the tightest upper bound.  This motivates our approach in which we train a neural network to directly predict the $\lambda$s from a given partial assignment to the random variables.  This approach is particularly applicable in typical scenarios in which a PGM has already been fit to data and we are simply using the PGM to make (repeated) MAP inference queries.

A few remarks. First, computing the MBE bound scales exponentially in the $i$-bound: polynomial-time for fixed $i$, but far from practical on large networks. Our aim is to move that cost from test time to training time, so that a network of sufficient capacity, given enough training configurations, approximates what the iterative methods compute and the test-time cost of a ``good'' bound reduces to a single forward pass. To efficiently encode the upper bound prediction task in a neural network architecture, we make a few small changes to the representation of the MAP prediction task.

First, we will train our neural network by directly minimizing the dual objective.  So, no labeled training examples are required.  This makes the approach flexible across many different models and does not require us to compute expensive $i$-bounds during training.  To simplify training, we replace the hard $\max$ in the upper bound with a smooth log-sum-exp surrogate:
\begin{equation}
    D_\tau(\boldsymbol{\theta}, \boldsymbol{\lambda}) = \sum_{\alpha} \tau \log \sum_{\mathbf{x}_{C_\alpha}} %
    \exp\left(\frac{\theta_{C_\alpha}(\mathbf{x}_{C_\alpha},\lambda)}{\tau}\right),
    \label{eq:smooth_dual}
\end{equation}
where $\tau > 0$ is a temperature. As $\tau \to 0$, $D_\tau \to D$. 

\begin{proposition}\label{prop:validity}
$D_\tau(\boldsymbol{\theta}, \boldsymbol{\lambda}) \geq D(\boldsymbol{\theta}, \boldsymbol{\lambda}) \geq \mathrm{MAP}^*$ for all $\tau > 0$ and all $\boldsymbol{\lambda}$.
\end{proposition}
\begin{proof}
\emph{Step 1 ($D \ge \mathrm{MAP}^*$).} Fix any complete assignment $\mathbf{x}$. Each edge
contributes $\lambda_{\alpha\beta}$ to one cluster and $\lambda_{\beta\alpha}$ to the other,
both at the same separator sub-assignment since $\mathbf{x}$ is one global assignment, so
by~\eqref{eq:antisymmetry} the pair cancels and
\[
\sum_\alpha \theta_{C_\alpha}(\mathbf{x}_{C_\alpha}, \lambda)
\;=\; \sum_{\gamma \in \mathcal{F}} \theta_\gamma(\mathbf{x}_\gamma),
\]
each original log-potential having been assigned to exactly one cluster during partitioning
(Sec.~\ref{sec:background}). Replacing each cluster term by its maximum only increases the sum, so $D$ bounds the
score of every $\mathbf{x}$, hence $\mathrm{MAP}^*$.

\emph{Step 2 ($D_\tau \ge D$).} Let $M_\alpha$ be the maximum reparameterized entry of cluster
$\alpha$ and $n_\alpha$ its number of configurations. The sum
$\sum_{y} \exp(\theta_{C_\alpha}(y,\lambda)/\tau)$ contains $\exp(M_\alpha/\tau)$ and all terms
are positive, so $\tau\log$ of it is at least $M_\alpha$; summing over clusters gives the claim.
That same sum has $n_\alpha$ terms, none larger than $\exp(M_\alpha/\tau)$, which gives
\[
0 \;\le\; D_\tau - D \;\le\; \tau \sum_\alpha \log n_\alpha .
\]
The surrogate gap is $O(\tau)$, the quantitative form of $D_\tau \to D$, and why annealing
$\tau$ is safe. Neither step uses any property of the network: Step 1 needs only that $\lambda$
is antisymmetric, which $A$ enforces by construction, and Step 2 only $\tau > 0$. Validity
belongs to the parameterization; only tightness depends on training.
\end{proof}

$D(\boldsymbol{\theta}, \boldsymbol{\lambda})$ is convex in $\boldsymbol{\lambda}$ (sum of pointwise maxima of affine functions), so subgradient methods could be used directly, but we use the smooth surrogate $D_\tau$ for compatibility with standard gradient-based optimizers.

Prior amortized-inference
work either supervises against a slower exact
solver~\citep{nair2020solving} or unrolls iterations of the target
algorithm and backpropagates through them. We do neither: $D_\tau$ is
a closed-form differentiable function, so a single
backward pass through the smooth dual yields the gradient. The
training distribution is evidence configurations sampled at varying
observation ratios, with no inference labels of any kind. This makes
training substantially cheaper than supervised alternatives and removes LP-solver availability as a deployment requirement.

Second, we represent $\lambda$ more compactly. The antisymmetry
constraint~\eqref{eq:antisymmetry} couples the two directions of every edge, so
it suffices to store one real number per (undirected-edge,
separator-assignment) pair. Stack these into a single edge-message vector
$m \in \mathbb{R}^{\mathrm{edim}}$, where
\[
\mathrm{edim} \;=\; \sum_{(\alpha,\beta) \in \mathcal{E}} \prod_{i \in S_{\alpha\beta}} d_i
\]
counts those pairs and $d_i$ is the domain size of variable $i$. Let
$\lambda_{\dim} = \sum_{\alpha} \prod_{i \in C_\alpha} d_i$ be the total number
of cluster-table entries, and let $\theta_{\mathrm{flat}} \in
\mathbb{R}^{\lambda_{\dim}}$ be the concatenation of all cluster tables,
ordered so that the $\alpha$-th block of size $\prod_{i \in C_\alpha} d_i$
holds the entries $\{\theta_{C_\alpha}(x_{C_\alpha})\}_{x_{C_\alpha}}$ of
cluster~$\alpha$. The reparameterization is then a single linear map
\[
\theta'_{\mathrm{flat}} \;=\; \theta_{\mathrm{flat}} + A\,m,
\]
where $A \in \{-1, 0, +1\}^{\lambda_{\dim} \times \mathrm{edim}}$ is a fixed
sparse matrix determined entirely by the join-graph structure. Each column of
$A$ corresponds to one (edge, separator-assignment) pair
$\bigl((\alpha,\beta), x_S\bigr)$ and contains $+1$ in every row of
$\theta_{\mathrm{flat}}$ that lies in cluster~$\alpha$ and is consistent with
$x_S$, $-1$ in every row that lies in cluster~$\beta$ and is consistent with
$x_S$, and $0$ elsewhere. The matched $\pm 1$ pattern is exactly the antisymmetry of the reparameterization~\eqref{eq:antisymmetry}: a complete assignment selects one entry per cluster, so the $+1$ and $-1$ contributions of
each column cancel. (The column itself sums to zero only when the two clusters on an edge have equally many entries consistent with $\mathbf{x}_S$; validity does not depend on this.)

As the neural network essentially produces $\lambda$s that satisfy the antisymmetry constraint, its output can always be turned into a valid upper bound on the MAP problem.
Consequently, training never has to enforce feasibility as a
constraint or check it post-hoc; the architecture restricts
predictions to the valid set. This is structurally different from
calibration-based methods, which produce predictions in an
unrestricted space and use a held-out calibration set to estimate
the residual violation probability.

Finally, note that the upper bounds produced by our methods and the iterative methods in the best case yield the tightest upper bound on $\text{MAP}^*$ attainable by the given join graph decomposition, but this bound is tight only in special cases, e.g., tree-structured models or log-supermodular models \citep{ruozzi2012bethe}.

\section{Experiments}
\label{sec:exp-map}

We evaluate on 15 PGMs from the UAI 2022 inference competition,
covering Bayesian networks, image segmentation and grid Markov random fields. The headline results are Table~\ref{tab:map} (MAP) and Tables~\ref{tab:t3_main_singlepass}--\ref{tab:t3_main_crossmu} ($K{=}2$ CMPE); full results for $K{=}1$ CMPE are in App.~\ref{app:meanstd}. All experiments
use $i$-bound $4$, with $100$ evidence seeds per PGM for MAP and $50$ for CMPE (the $37$-PGM suite of App.~\ref{app:meanstd} remains at $10$). Our reference $D^{\mathrm{LP}}$ is the exact LP optimum solved with
CVXPY/CLARABEL (Appendix~\ref{app:lp}); The JGLP coordinate
descent occasionally terminates at non-globally-optimal fixed points,
so we use the LP optimum rather than JGLP's converged value as the
ground-truth lower bound on the dual.

\subsection{Neural Network Training vs. JGLP}
\label{sec:standalone}

The primary evaluation of our neural bounds is how much iterative inference they save. A single neural forward pass ($0.23$--$0.39$\,ms on CPU, App.~\ref{app:standalone_v2}) produces a valid upper bound on MAP$^*$; this bound is typically tighter than the MBE starting point and can serve as a warm-start for JGLP. We measure the effective speedup by counting JGLP iterations needed to reach a target relative error from the converged dual $D^*$, comparing zero-init vs neural-init. %

We use a 3-layer MLP with LayerNorm and GELU activations: $512 \to 512 \to 256 \to e_{\dim}$. The output layer has no activation function (edge messages are unconstrained). The output is initialized with small random weights (std $= 0.001$) so that the initial bound $D(f_\phi(\mathbf{e})) \approx D_0$ (the MBE bound).

We anneal $\tau$, the temperature parameter of the smooth surrogate, during training: $\tau = 0.01$ for the first third, then $\tau = 0.001$ for the remaining epochs.  We use AdamW optimization with cosine annealing (lr $= 10^{-3}$, 500 gradient steps at batch size 16, drawing a fresh evidence
configuration per sample). Training data consists of evidence configurations sampled at various evidence ratios (10--90\% of variables observed). No JGLP or other iterative inference is required during training.

\begin{figure}[t]
\centering
\includegraphics[width=\textwidth]{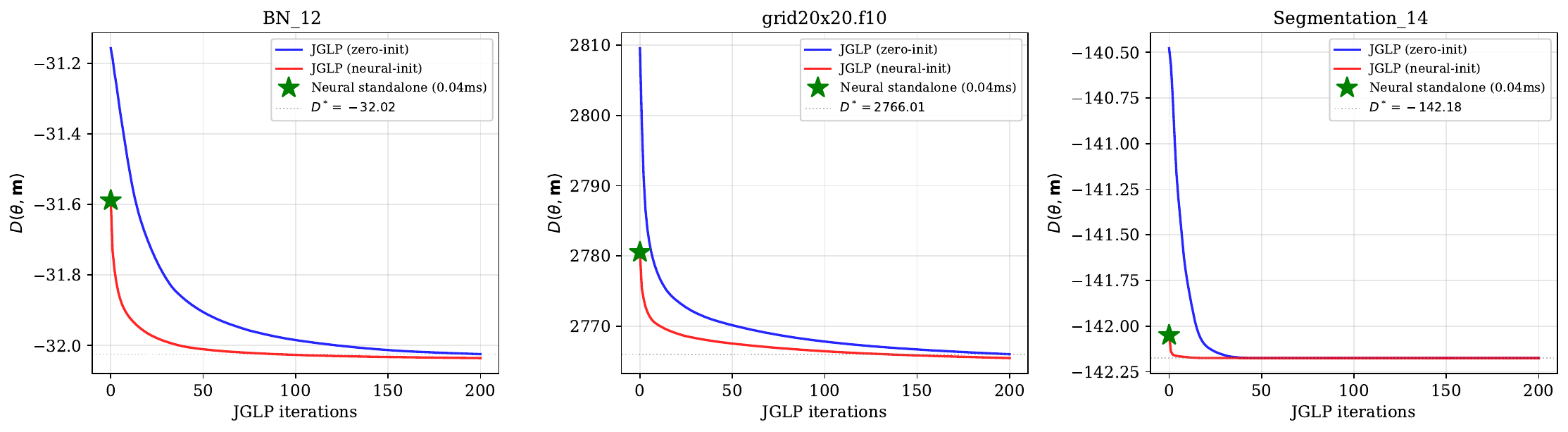}
\caption{\textbf{JGLP convergence with and without neural warm-start} on three
representative PGMs (BN\_12, grid20$\times$20.f10, Segmentation\_14). Both
curves provably converge to the same optimum $D^*$ since $D(\boldsymbol{\theta},
\mathbf{m})$ is convex in $\mathbf{m}$. The green star marks the standalone
neural bound from a single forward pass ($\sim\!0.04$~ms on GPU) and is used as the
warm-start for JGLP at iteration~0; it starts strictly below zero-init and
reaches a small neighborhood of $D^*$ in fewer JGLP iterations. The three
panels span the BN, Grid, and Segmentation families; full results across all
37~PGMs are reported in Appendix~\ref{app:table1_full} and
Table~\ref{tab:map}.}
\label{fig:convergence}
\end{figure}

Since $D(\boldsymbol{\theta}, \mathbf{m})$ is a convex function of $m$, we can compute the exact optimum via CVXPY/CLARABEL (formulation in Appendix~\ref{app:lp}).  %
Figure~\ref{fig:convergence} shows one PGM per family; Table~\ref{tab:t1_main} reports the absolute 
gap $D^{(t)} - D^{\mathrm{LP}}$ at fixed JGLP iteration checkpoints
$t \in \{10, 50, 100\}$, mean $\pm$ std over 100 bin-0 test seeds per
PGM. Lower is better. %

Across the 15 PGMs, neural warm-start matches or beats zero-init
at iter=10 on $100\%$ of seeds for $11$ of them. The gap then closes
rapidly as JGLP iterates: on the Segmentation family it drops by $4$--$6$
orders of magnitude by iter=100---for example, Seg\_12's neural-init gap
falls from $0.701$ to $2{\times}10^{-7}$ in App.~\ref{app:meanstd}---while BN models show smaller
but consistent gains ($\sim$1--2 orders). Three regimes resist this
pattern in the full $37$-PGM suite (App.~\ref{app:meanstd}). (i) \emph{Degenerate grids} (grid10$\times$10, grid40$\times$40.f2)
already sit within $10^{-8}$ of $D^{\mathrm{LP}}$ at MBE construction with
$i$-bound~$4$; neural cannot improve, and the small overshoot at iter=10
is floating-point round-off, not a training failure.
(ii) \emph{Promedas networks} have near-zero dual values, so absolute
gaps look small even when relative gains are substantial.
(iii) \emph{Seg\_14} converges fast enough on its own that zero-init
reaches $2{\times}10^{-6}$ by iter=50, below the neural arm; both are then at
the LP solver's working accuracy, so that ordering is not a result.

\begin{table*}[t]\centering\scriptsize
\caption{\textbf{Standalone bound quality}: dual gap $|D - D^{\mathrm{LP}}|$ (mean $\pm$ std) at fixed JGLP iterations. J = zero-init, N = neural-init; lower is closer to the LP optimum, so \textbf{bold} marks the neural arm where it ties or beats zero-init within one standard deviation. Mean and population standard deviation over $100$ bin-0 test seeds per PGM (Segmentation\_18 and \_19 have $99$: one seed each fails deterministically in the LP solver and is reported rather than dropped). All five checkpoints $\{0,10,50,100,500\}$ are in App.~\ref{app:meanstd}; the paired per-seed comparison, which removes the shared instance difficulty these standard deviations are dominated by, is Table~\ref{tab:paired_map}. $^{\dagger}$Same UAI model as Grids\_29 under a second name; $^{\ddagger}$a distinct $400$-variable grid, itself shipped as Grids\_30. Both rows of a duplicated pair are kept because each carries an independently trained network, but every count is over the $15$ distinct models rather than the $16$ rows (App.~\ref{app:table1_full}).}
\label{tab:map}\label{tab:t1_main}\label{tab:iters_eps_v2}
\resizebox{\textwidth}{!}{%
\begin{tabular}{@{}l|rr|rr|rr@{}}
\toprule
  & \multicolumn{2}{c|}{iter=10} & \multicolumn{2}{c|}{iter=50} & \multicolumn{2}{c}{iter=100} \\
\textbf{PGM} & J & N & J & N & J & N \\
\midrule
BN\_12 & 0.525~$\pm$~0.248 & \textbf{0.304~$\pm$~0.178} & 0.150~$\pm$~0.131 & \textbf{0.095~$\pm$~0.093} & 0.065~$\pm$~0.078 & \textbf{0.042~$\pm$~0.056} \\
BN\_13 & 0.469~$\pm$~0.129 & \textbf{0.267~$\pm$~0.103} & 0.113~$\pm$~0.071 & \textbf{0.089~$\pm$~0.070} & 0.055~$\pm$~0.050 & \textbf{0.047~$\pm$~0.048} \\
BN\_14 & 0.305~$\pm$~0.142 & \textbf{0.191~$\pm$~0.113} & 0.058~$\pm$~0.060 & \textbf{0.045~$\pm$~0.051} & 0.015~$\pm$~0.028 & \textbf{0.013~$\pm$~0.024} \\
BN\_9 & 0.047~$\pm$~0.061 & \textbf{0.039~$\pm$~0.055} & 2.96e-03~$\pm$~9.89e-03 & \textbf{3.09e-03~$\pm$~0.011} & 6.27e-04~$\pm$~2.63e-03 & \textbf{6.97e-04~$\pm$~3.07e-03} \\
Grids\_26 & 2.70~$\pm$~2.11 & \textbf{2.71~$\pm$~2.10} & 0.237~$\pm$~0.500 & \textbf{0.237~$\pm$~0.500} & 0.036~$\pm$~0.159 & \textbf{0.037~$\pm$~0.160} \\
Grids\_28 & 4.18~$\pm$~3.41 & \textbf{4.19~$\pm$~3.40} & 0.388~$\pm$~0.813 & \textbf{0.388~$\pm$~0.814} & 0.068~$\pm$~0.295 & \textbf{0.069~$\pm$~0.297} \\
Grids\_29 & 1.26~$\pm$~0.917 & \textbf{1.24~$\pm$~0.912} & 0.123~$\pm$~0.223 & \textbf{0.123~$\pm$~0.222} & 0.021~$\pm$~0.091 & \textbf{0.021~$\pm$~0.092} \\
Seg\_12 & 1.06~$\pm$~0.615 & \textbf{0.793~$\pm$~0.523} & 0.135~$\pm$~0.220 & \textbf{0.142~$\pm$~0.223} & 0.032~$\pm$~0.093 & \textbf{0.034~$\pm$~0.095} \\
Seg\_14 & 0.345~$\pm$~0.265 & \textbf{0.123~$\pm$~0.165} & 1.54e-03~$\pm$~9.44e-03 & \textbf{1.04e-03~$\pm$~5.33e-03} & 1.58e-07~$\pm$~2.87e-07 & \textbf{5.63e-06~$\pm$~7.88e-06} \\
Seg\_16 & 1.88~$\pm$~0.514 & \textbf{0.381~$\pm$~0.379} & 0.253~$\pm$~0.213 & \textbf{0.025~$\pm$~0.071} & 0.017~$\pm$~0.045 & \textbf{3.06e-04~$\pm$~1.87e-03} \\
Seg\_18 & 0.873~$\pm$~0.632 & \textbf{0.576~$\pm$~0.491} & 0.065~$\pm$~0.120 & \textbf{0.058~$\pm$~0.111} & 5.23e-03~$\pm$~0.034 & \textbf{5.68e-03~$\pm$~0.033} \\
Seg\_19 & 0.152~$\pm$~0.207 & \textbf{0.110~$\pm$~0.147} & 0.015~$\pm$~0.050 & \textbf{0.014~$\pm$~0.044} & 4.43e-03~$\pm$~0.022 & \textbf{4.15e-03~$\pm$~0.019} \\
grid10$\times$10.f5 & 0.311~$\pm$~0.343 & \textbf{0.302~$\pm$~0.337} & 6.65e-03~$\pm$~0.042 & \textbf{6.48e-03~$\pm$~0.042} & 4.61e-04~$\pm$~3.55e-03 & \textbf{4.40e-04~$\pm$~3.43e-03} \\
grid10$\times$10.f10 & 0.536~$\pm$~0.799 & \textbf{0.537~$\pm$~0.792} & 0.079~$\pm$~0.214 & \textbf{0.079~$\pm$~0.212} & 0.015~$\pm$~0.090 & \textbf{0.015~$\pm$~0.089} \\
grid20$\times$20.f5$^{\dagger}$ & 1.26~$\pm$~0.917 & \textbf{1.24~$\pm$~0.912} & 0.123~$\pm$~0.223 & \textbf{0.123~$\pm$~0.222} & 0.021~$\pm$~0.091 & \textbf{0.021~$\pm$~0.092} \\
grid20$\times$20.f10$^{\ddagger}$ & 14.39~$\pm$~3.37 & \textbf{12.43~$\pm$~3.03} & 5.08~$\pm$~1.92 & \textbf{4.81~$\pm$~1.87} & 2.77~$\pm$~1.41 & \textbf{2.71~$\pm$~1.43} \\
\bottomrule\end{tabular}}\end{table*}

\paragraph{Amortized cost analysis.} A single-query forward pass costs 0.23--0.39 ms on CPU
(the $\sim$0.04 ms figure quoted for batched-GPU throughput is not the per-query latency),
against 1.6--131 ms for one JGLP iteration depending on PGM complexity, so the prediction is a
fraction of an iteration and is recovered whenever it saves one. Per-PGM wall-clock for
constrained MAP is reported in App.~\ref{app:wallclock}. The defensible claim is bound quality
at a fixed \emph{iteration} budget --- what a branch-and-bound node expansion actually spends ---
rather than wall-clock dominance at every scale.

\paragraph{Edge vs cluster parameterization.}
The edge-message parameterization ($e_{\dim}$) is 2$\times$ smaller than the cluster-table parameterization ($\lambda_{\dim}$) while capturing 98.6\% of JGLP's corrections (lstsq residual 1.4\%). The reduced output dimension improves generalization: self-supervised training with edge messages achieves 94\% neural-better vs.\ 0\% with direct cluster-table prediction (which violates the reparameterization constraint).

\paragraph{Self-supervised vs supervised training.}
We compare pure self-supervised training (directly minimizing $D_\tau$) against supervised training with JGLP edge-message targets. On BN\_12 bin 0: self-supervised achieves 94\% neural-better vs 72\% for MSE-supervised. The MSE model produces larger magnitude edge shifts ($2.8\times$ scale ratio) but sometimes overshoots the optimal reparameterization, resulting in 28\% of instances worse than zero-init. The self-supervised loss naturally avoids this by directly optimizing the bound quality. (App.~\ref{app:ablations})

\subsection{Constrained Most Probable Explanation}
\label{sec:exp-cmpe}
\citet{rahman2021novel} consider a constrained version of the most probable explanation task that allows constraints that are given by PGMs that, for simplicity, are over the same graph and factorization.  Formally, let $(G,\mathcal{F}, \theta)$ be a PGM.  Consider $K$ constraint PGMs $(G,\mathcal{F}, \theta^k)$ for each $k\in\{1,\ldots,K\}$ over the same graph and factorization  with possibly different potential functions.  Each constraint will be of the form $\sum_{\alpha \in \mathcal{F}} \theta_\alpha^k(\mathbf{x}_\alpha) \leq q_k$ for some $q_k\in\mathbb{R}$.  This yields the optimization problem
\begin{align*}
    \max_\mathbf{x} \sum_{\alpha \in \mathcal{F}} \theta_\alpha(\mathbf{x}_\alpha) \text{ subject to}\\
    \sum_{\alpha \in \mathcal{F}} \theta_\alpha^k(\mathbf{x}_\alpha) \leq q_k \text{ for each $k\in\{1,\ldots,K\}$.}
\end{align*}
Note that we can apply the same argument to the case where each one of the PGMs is replaced by the same join graph (again with possibly different log-factors).  To handle constrained optimization problems of this form, \citet{rahman2021novel} consider a Lagrangian relaxation
\begin{align*}
  \mathcal{L}(\boldsymbol\mu) &= \max_\mathbf{x} \left[\sum_{\alpha \in \mathcal{F}} \theta_\alpha(\mathbf{x}_\alpha) - \sum_k \mu_k\left(\sum_{\alpha \in \mathcal{F}} \theta_\alpha^k(\mathbf{x}_\alpha) - q_k\right)\right]\\
  & \leq \sum_{\alpha \in \mathcal{F}} \max_{\mathbf{x}_\alpha} \left[\theta_\alpha(\mathbf{x}_\alpha) - \sum_k \mu_k \theta_\alpha^k(\mathbf{x}_\alpha)\right] - \sum_k \mu_k q_k.
\end{align*}

Since the upper bound on $\mathcal{L}$ is convex in $\mu \geq 0$, we can minimize it via projected (sub)gradient descent. %
To compute the subgradients, we must compute the $\max$ in the lower bound, which can be done (approximately) with JGLP.  As a result, we can train a $\mu$-conditioned neural model for this task: the input is $(\text{evidence\_encoding}, \mu)$ and the output is edge messages $\mathbf{m}$. Training data consists of $100$ evidence configurations $\times$ $50$ values of $\mu \in [0, 3]$, yielding $5{,}000$ samples per PGM. 

Following \cite{rahman2021novel}, we construct each CMPE instance by
perturbing the log-potentials of the original model: for every clique
$\alpha$ and assignment $x_\alpha$, we draw $\epsilon \sim
\mathcal{N}(0, \sigma^2)$ i.i.d.\ per table entry and set
$\theta^k_\alpha(x_\alpha) = \theta_\alpha(x_\alpha) + \epsilon$. The
constraint threshold $q_k$ is the $80$th percentile of $\sum_\alpha
\theta^k_\alpha(x_\alpha)$ on the test sample. We use $\sigma = 0.5$ to keep the constraint active at the
optimal multiplier ($\mu^* > 0$); at $\sigma = 0.1$ the combined Lagrangian
collapses to noise near $\mu^*$ (App.~\ref{app:cmpe}). The two-constraint case
draws independent noise and its own threshold for each $k$.

We compare four different methods to estimate the subgradients: (1) \textbf{JGLP}--- initialized at $\mathbf{m} = 0$; (2) \textbf{NN}---inject NN-predicted $\mathbf{m}$ then run JGLP; (3) \textbf{JGLP+WS}---warm-start JGLP from $\mathbf{m}$ given by the most recent run of JGLP; (4) \textbf{NN+TTA}---test-time adaptation, at each gradient step $k$, fine-tune the NN on all previous $({\mu_j}, \boldsymbol{\theta}^{(\mu_j)})$ pairs via the self-supervised $D_\tau$ loss (replay buffer of up to 20 entries, 10 inner gradient steps). The validity guarantee $L(\mu) \geq \text{CMPE}^*$ holds for any $\mathbf{m}$.

The buffer is FIFO and the inner loss is identical to the offline objective, so no separate test-time validation is required and the buffer accumulates no out-of-distribution examples by construction.

\paragraph{Results.} We run 100 gradient steps on $\mu$ and at each step run JGLP from each initialization for up to 500 iterations (or until $|D(t) - D(t{-}1)| < 10^{-6}$). Tables~\ref{tab:t3_main_singlepass}--\ref{tab:t3_main_crossmu} report the dual gap $(L(\mu) - L^{\mathrm{LP}}(\mu))$ at fixed inner JGLP iteration checkpoints, as mean $\pm$ std over $50$ evidence seeds $\times$ $100$ $\mu$-grad steps; the $K{=}1$ counterpart is Table~\ref{tab:cmpe_eps_v2} in App.~\ref{app:meanstd}.

Test-time replay adaptation drives the dual gap $(L(\mu) - L^{\mathrm{LP}}(\mu))$ closer to zero than zero-init on every non-degenerate PGM: by roughly one order of magnitude at iter=50 on the BN and Segmentation families ($4.7$--$18\times$ at $K{=}1$, $3.2$--$14\times$ at $K{=}2$), and by up to four orders on grid10$\times$10.f5, where zero-initialized JGLP stalls. Per-seed win rates at iter=10 ($n = 50$ seeds) are $94$--$100\%$ on the BN and Segmentation models and $84$--$100\%$ on the non-degenerate grids; zero-init wins only on the two degenerate grid40$\times$40 rows ($4\%$ and $14\%$), where MBE already sits within $10^{-8}$ of $L^{\mathrm{LP}}$ (App.~\ref{app:cmpe_paired}). Even offline NN warm-start (M2) helps on most PGMs, with M3 additionally specializing to the local $\mu$ trajectory via the replay buffer.

JGLP+WS (no learning, just carrying converged
$\Delta\lambda$ across $\mu$-steps) is the strongest non-neural
baseline. On Bayesian networks (BN\_9, BN\_12, BN\_13, BN\_14),
$\mu$-trajectories converge quickly and successive optimal $\lambda$
values are nearby, so $\lambda$-continuity alone closes most of the
gap (e.g., BN\_12 iter=10: $0.082 \pm 0.095$ vs NN+TTA $0.097 \pm 0.126$). On grids
and segmentation models, $\mu$ keeps moving across the trajectory and
stale $\Delta\lambda$ pulls JGLP toward the previous optimum;
NN+TTA's learned structural prior dominates by one to two orders of magnitude (Seg\_14 iter=10: JGLP+WS $9.49 \pm 5.87$ vs NN+TTA $0.083 \pm 0.314$; Grids\_28: $1019.27 \pm 884.50$ vs $52.97 \pm 90.61$). The two adaptation channels are complementary: $\lambda$-continuity exploits local smoothness in the optimization landscape, while the learned prior captures global structure that holds across the trajectory. We leave a hybrid (carrying $\Delta\lambda$ \emph{and} TTA-fine-tuning the network) to future work.

\begin{table*}[t]\centering\scriptsize
\caption{\textbf{CMPE bound quality} (two constraints, single-pass methods): absolute dual gap $|L(\mu) - L^{\mathrm{LP}}(\mu)|$ (mean $\pm$ std) at three fixed inner JGLP iteration budgets on the 15 PGM rows (14 distinct models) common to MAP, CMPE $K=1$ and CMPE $K=2$. Methods: \textbf{JGLP} = zero-init JGLP (no learning, no continuity); \textbf{NN} = single neural forward pass producing edge messages, frozen weights. Statistics aggregate over (seed $\times$ $\mu$-gradient step) trials (50 seeds $\times$ 100 $\mu$-steps each, i.e.\ 5{,}000 trials per cell).}
\label{tab:t3_main_singlepass}
\resizebox{\textwidth}{!}{%
\begin{tabular}{@{}l|rr|rr|rr@{}}
\toprule
  & \multicolumn{2}{c|}{iter=10} & \multicolumn{2}{c|}{iter=50} & \multicolumn{2}{c}{iter=100} \\
\textbf{PGM} & JGLP & NN & JGLP & NN & JGLP & NN \\
\midrule
BN\_12 & 0.622~$\pm$~0.152 & 0.523~$\pm$~0.136 & 0.207~$\pm$~0.068 & 0.164~$\pm$~0.065 & 0.107~$\pm$~0.052 & 0.087~$\pm$~0.047 \\
BN\_13 & 0.461~$\pm$~0.127 & 0.351~$\pm$~0.123 & 0.121~$\pm$~0.109 & 0.063~$\pm$~0.051 & 0.055~$\pm$~0.107 & 0.025~$\pm$~0.028 \\
BN\_14 & 0.305~$\pm$~0.126 & 0.240~$\pm$~0.121 & 0.050~$\pm$~0.082 & 0.024~$\pm$~0.033 & 0.020~$\pm$~0.076 & 6.60e-03~$\pm$~0.014 \\
BN\_9 & 0.179~$\pm$~0.090 & 0.167~$\pm$~0.074 & 0.053~$\pm$~0.062 & 0.027~$\pm$~0.035 & 0.026~$\pm$~0.055 & 9.18e-03~$\pm$~0.019 \\
Grids\_26 & 36.07~$\pm$~52.99 & 37.03~$\pm$~52.82 & 4.42~$\pm$~11.52 & 4.43~$\pm$~11.51 & 0.664~$\pm$~2.63 & 0.663~$\pm$~2.63 \\
Grids\_28 & 82.68~$\pm$~120.14 & 84.23~$\pm$~120.35 & 9.15~$\pm$~24.42 & 9.18~$\pm$~24.46 & 1.05~$\pm$~4.68 & 1.06~$\pm$~4.70 \\
Grids\_29 & 7.71~$\pm$~11.94 & 8.30~$\pm$~11.90 & 0.775~$\pm$~2.48 & 0.781~$\pm$~2.43 & 0.122~$\pm$~0.588 & 0.121~$\pm$~0.572 \\
Seg\_12 & 0.709~$\pm$~0.721 & 1.05~$\pm$~0.969 & 0.094~$\pm$~0.179 & 0.168~$\pm$~0.333 & 0.022~$\pm$~0.076 & 0.055~$\pm$~0.160 \\
Seg\_14 & 0.828~$\pm$~1.02 & 0.832~$\pm$~0.879 & 0.023~$\pm$~0.063 & 0.017~$\pm$~0.051 & 2.83e-03~$\pm$~0.018 & 1.91e-03~$\pm$~0.015 \\
Seg\_16 & 2.37~$\pm$~2.08 & 1.51~$\pm$~1.58 & 0.185~$\pm$~0.263 & 0.087~$\pm$~0.176 & 0.014~$\pm$~0.059 & 8.58e-03~$\pm$~0.048 \\
Seg\_18 & 0.883~$\pm$~0.830 & 0.628~$\pm$~0.586 & 0.057~$\pm$~0.116 & 0.031~$\pm$~0.062 & 3.39e-03~$\pm$~0.027 & 2.33e-03~$\pm$~7.20e-03 \\
Seg\_19 & 2.54~$\pm$~3.53 & 2.06~$\pm$~2.59 & 0.600~$\pm$~1.10 & 0.467~$\pm$~0.864 & 0.305~$\pm$~0.691 & 0.235~$\pm$~0.558 \\
grid10$\times$10.f5 & 0.172~$\pm$~0.280 & 0.244~$\pm$~0.244 & 0.044~$\pm$~0.183 & 4.58e-03~$\pm$~0.037 & 0.041~$\pm$~0.180 & 1.42e-04~$\pm$~1.33e-03 \\
grid20$\times$20.f5 & 7.71~$\pm$~11.94 & 8.33~$\pm$~11.87 & 0.775~$\pm$~2.48 & 0.774~$\pm$~2.45 & 0.122~$\pm$~0.588 & 0.121~$\pm$~0.579 \\
grid20$\times$20.f10 & 207.15~$\pm$~202.17 & 204.66~$\pm$~201.72 & 64.52~$\pm$~68.23 & 63.58~$\pm$~68.39 & 33.86~$\pm$~39.42 & 33.48~$\pm$~39.47 \\
\bottomrule
\end{tabular}}
\end{table*}

\begin{table*}[t]\centering\scriptsize
\caption{\textbf{CMPE bound quality} (two constraints, cross-$\mu$-step methods): absolute dual gap $|L(\mu) - L^{\mathrm{LP}}(\mu)|$ (mean $\pm$ std) at three fixed inner JGLP iteration budgets on the same 15 rows. Methods: \textbf{JGLP+WS} = JGLP warm-started by injecting the converged cluster shift from the previous $\mu$-step (no neural network; tests whether free $\lambda$-continuity alone suffices); \textbf{NN+TTA} = neural warm-start with replay-buffer test-time adaptation of the NN weights across $\mu$-steps (ours). Statistics aggregate over (seed $\times$ $\mu$-gradient step) trials (50 seeds $\times$ 100 $\mu$-steps each, i.e.\ 5{,}000 trials per cell).}
\label{tab:t3_main_crossmu}
\resizebox{\textwidth}{!}{%
\begin{tabular}{@{}l|rr|rr|rr@{}}
\toprule
  & \multicolumn{2}{c|}{iter=10} & \multicolumn{2}{c|}{iter=50} & \multicolumn{2}{c}{iter=100} \\
\textbf{PGM} & JGLP+WS & NN+TTA & JGLP+WS & NN+TTA & JGLP+WS & NN+TTA \\
\midrule
BN\_12 & 0.082~$\pm$~0.095 & 0.097~$\pm$~0.126 & 0.030~$\pm$~0.038 & 0.046~$\pm$~0.054 & 0.019~$\pm$~0.026 & 0.029~$\pm$~0.034 \\
BN\_13 & 0.065~$\pm$~0.096 & 0.044~$\pm$~0.082 & 0.015~$\pm$~0.030 & 0.014~$\pm$~0.027 & 7.20e-03~$\pm$~0.015 & 6.93e-03~$\pm$~0.015 \\
BN\_14 & 0.072~$\pm$~0.110 & 0.027~$\pm$~0.071 & 7.09e-03~$\pm$~0.022 & 4.80e-03~$\pm$~0.018 & 1.76e-03~$\pm$~9.29e-03 & 1.61e-03~$\pm$~8.25e-03 \\
BN\_9 & 0.028~$\pm$~0.046 & 0.021~$\pm$~0.046 & 2.47e-03~$\pm$~7.30e-03 & 4.63e-03~$\pm$~0.016 & 5.32e-04~$\pm$~2.85e-03 & 1.72e-03~$\pm$~8.12e-03 \\
Grids\_26 & 428.95~$\pm$~370.99 & 16.02~$\pm$~34.12 & 13.75~$\pm$~13.29 & 2.13~$\pm$~7.35 & 1.30~$\pm$~2.84 & 0.337~$\pm$~1.86 \\
Grids\_28 & 1019.27~$\pm$~884.50 & 52.97~$\pm$~90.61 & 30.34~$\pm$~28.43 & 6.23~$\pm$~18.70 & 2.46~$\pm$~5.10 & 0.717~$\pm$~3.79 \\
Grids\_29 & 92.55~$\pm$~80.50 & 1.60~$\pm$~5.65 & 2.84~$\pm$~3.41 & 0.163~$\pm$~1.07 & 0.301~$\pm$~0.730 & 0.024~$\pm$~0.236 \\
Seg\_12 & 6.32~$\pm$~3.47 & 0.129~$\pm$~0.362 & 0.537~$\pm$~0.598 & 0.022~$\pm$~0.088 & 0.124~$\pm$~0.226 & 6.18e-03~$\pm$~0.038 \\
Seg\_14 & 9.49~$\pm$~5.87 & 0.083~$\pm$~0.314 & 0.389~$\pm$~0.446 & 3.00e-03~$\pm$~0.022 & 0.030~$\pm$~0.084 & 8.46e-04~$\pm$~9.08e-03 \\
Seg\_16 & 9.02~$\pm$~5.04 & 0.220~$\pm$~0.676 & 0.418~$\pm$~0.346 & 0.017~$\pm$~0.076 & 0.019~$\pm$~0.066 & 1.24e-03~$\pm$~0.015 \\
Seg\_18 & 6.19~$\pm$~3.12 & 0.085~$\pm$~0.262 & 0.219~$\pm$~0.274 & 4.17e-03~$\pm$~0.026 & 0.018~$\pm$~0.041 & 3.57e-04~$\pm$~4.78e-03 \\
Seg\_19 & 12.03~$\pm$~8.07 & 0.436~$\pm$~1.25 & 1.04~$\pm$~1.29 & 0.188~$\pm$~0.534 & 0.371~$\pm$~0.766 & 0.125~$\pm$~0.382 \\
grid10$\times$10.f5 & 2.30~$\pm$~2.11 & 0.015~$\pm$~0.072 & 0.022~$\pm$~0.076 & 3.89e-04~$\pm$~8.73e-03 & 7.00e-04~$\pm$~5.47e-03 & 5.96e-06~$\pm$~2.58e-04 \\
grid20$\times$20.f5 & 92.55~$\pm$~80.50 & 1.60~$\pm$~5.65 & 2.84~$\pm$~3.41 & 0.162~$\pm$~1.07 & 0.301~$\pm$~0.730 & 0.024~$\pm$~0.237 \\
grid20$\times$20.f10 & 913.11~$\pm$~525.72 & 101.83~$\pm$~134.02 & 150.23~$\pm$~49.77 & 41.94~$\pm$~52.83 & 48.58~$\pm$~29.88 & 24.49~$\pm$~32.52 \\
\bottomrule
\end{tabular}}
\end{table*}

\paragraph{Test-Time Adaptation}
\label{sec:tta}
The offline training distribution for the neural network sampled $\mu \sim \text{unif}(0, 1)$, but the
projected gradient descent on $\mu$ at test time is unconstrained
above and routinely drives $\mu$ outside this range: across all 15
PGMs, $36\%$ of $K{=}1$ queries and $25\%$ of $K{=}2$ queries occur at
$\mu > 1$, with maxima of $36.1$ and $19.4$ respectively. A frozen
offline-trained network on these tails would predict edge messages
fitted to a different Lagrangian; deterministic validity (Prop.~\ref{prop:validity})
still holds, but tightness degrades. Replay-buffer TTA continually
re-fits the network to the local $\mu$-region the trajectory has
actually visited, recovering tightness within $\sim 50$ inner JGLP
iterations even on tails $20$--$36\times$ the training upper bound
(Tables~\ref{tab:t3_main_singlepass}--\ref{tab:t3_main_crossmu}). This explains the
large NN+TTA-vs-NN gap on grids and segmentations, where $\mu$
keeps moving across the trajectory: NN+TTA's primary job is online
distribution adaptation, not just specialization.

\section{Related Work}
\label{sec:related}

\paragraph{Learning to branch / Learning for combinatorial optimization.}
A growing literature uses neural networks to guide search in combinatorial optimization \cite{nair2020solving, gasse2019exact, khalil2017learning}. These methods typically learn heuristics (branching decisions, node selection) rather than bounds, and provide no formal guarantees on bound validity. Concurrently, \cite{huang2022learning} learn cutting sets for integer programming, but require problem-specific cut generation. Our approach is orthogonal: we learn \emph{dual bounds} rather than search heuristics, with deterministic validity guarantees.

\paragraph{Amortized optimization.}
Amortized optimization \cite{amos2023tutorial} trains networks to predict solutions to optimization problems, trading per-instance computation for upfront training cost. Our work falls in this category but with a crucial difference: the ``feasibility'' of our predictions (validity of the bound) is guaranteed by the architecture, not learned. This contrasts with approaches that learn feasible solutions and must verify post-hoc \cite{donti2021dc3}.

\paragraph{Differentiating from RL for MILP.}
The RL-for-MILP literature \cite{nair2020solving, khalil2017learning} learns search heuristics (branching, node selection) via reinforcement learning, providing no formal bound guarantees. Our approach differs in three key ways: (1) we learn \emph{bounds} rather than heuristics, via self-supervised optimization (not RL); (2) we exploit PGM-specific reparameterization structure (antisymmetric messages) that has no analog in generic MILP; and (3) our bounds are \emph{deterministically} valid by construction, not statistically calibrated. The deterministic guarantee is the core contribution: a learned bound that occasionally violates $D \geq \text{MAP}^*$ is useless for pruning, while ours is safe by design.

\section{Conclusion}
\label{sec:conclusion}

On UAI benchmark PGMs spanning Bayesian networks, image
segmentation, and grid Markov random fields, neural warm-starts match or beat
zero-initialized JGLP everywhere except the degenerate grids, where the mini-bucket
bound already sits at the LP optimum, and are up to six orders of magnitude tighter in
the best case, with the gains concentrated in the early-iteration regime
(Table~\ref{tab:map}). On constrained MAP with one or two constraints, neural warm-start
with replay-buffer test-time adaptation gives the tightest dual gaps on the grid and
segmentation families by one to two orders of magnitude, while $\lambda$-continuity is
comparable on Bayesian networks
(Tables~\ref{tab:t3_main_singlepass}--\ref{tab:t3_main_crossmu}; full $K{=}1$ in
Appendix~\ref{app:meanstd}).

None of this is conditional on the
network being right. The guarantee $D(\theta - \mu g, m) \geq \mathrm{CMPE}^*$ holds
throughout, unconditionally on the network output, with no calibration set, conformal
quantile, or coverage budget. This is the line between a usable bound and an unusable
one: a learned bound that violates $D \geq \mathrm{MAP}^*$ even rarely cannot be used for
pruning in branch-and-bound or for certification, because one violation discards the
optimal solution or invalidates the certificate. Here validity is carried by the
antisymmetric parameterization rather than by training accuracy, so it survives untrained
and adversarial weights alike, at every point along the $\mu$-trajectory. Training buys
tightness; the architecture buys correctness.

\paragraph{Beyond JGLP.} The framework extends to any Lagrangian dual whose decomposition
admits antisymmetric messages---dual decomposition, TRW-BP, weighted CSPs,
minimum-cost flow---and to non-PGM problems with the same structure. (i) \emph{SVM/SMO}:
the SVM dual is concave in the multipliers $\boldsymbol{\alpha}$ and linear in the data,
structurally identical to $D(\boldsymbol{\theta}, \lambda)$, so a network predicting
warm-start $\boldsymbol{\alpha}$ would yield valid dual bounds by construction and
accelerate sequential minimal optimization where many SVMs are trained on similar data.
(ii) \emph{Network flow}: the LP dual of minimum-cost flow carries antisymmetric dual
variables on edges, directly analogous to our edge-message parameterization, so the same
scatter matrix $A$ applies. (iii) \emph{Weighted CSPs}: soft-constraint tables are
log-potentials on the same join-graph structure, so the construction applies unchanged,
though our suite (App.~\ref{app:benchmarks}) does not yet include them. In each case the
architecture and the self-supervised loss are identical; only $A$ changes.

\paragraph{Limitations.} We state the scope of our claims directly. The network is trained
per join graph; a new topology, or the same graph at a different $i$-bound, requires
retraining, so the amortization we demonstrate is over repeated queries on a fixed topology
rather than across topologies. The benefit is confined to the early-iteration regime: both
initializations descend the same convex objective to the same optimum, so the warm start
buys the shape of the first few tens of iterations, not the converged bound. Where the
moment-matched mini-bucket bound already sits at the LP optimum (grid10$\times$10,
grid40$\times$40.f2), no reparameterization can improve it and the neural start cannot help.
And on Bayesian networks, where successive $\mu$-step optima are close, $\lambda$-continuity
is the stronger strategy; the learned prior dominates where the trajectory drifts.

\paragraph{Outlook.} Two directions follow. First, JGLP+WS ($\lambda$-continuity, no
learning) and NN+TTA (learning, no $\lambda$-continuity) dominate on disjoint families in
Tables~\ref{tab:t3_main_singlepass}--\ref{tab:t3_main_crossmu}, which suggests they are
complementary rather than competing: carrying $\Delta\lambda$ across $\mu$-steps \emph{and}
fine-tuning the network should dominate either alone, particularly when test-time
adaptation must absorb out-of-distribution $\mu$. Second, and more consequential, our MLP
is tied to a single join graph. Since $A$ is determined entirely by the join-graph
structure, a graph neural network with edge-level readouts transfers the parameterization
unchanged and is the natural route to amortizing across topologies. Whether that closes the
gap, or whether cross-topology transfer is harder than the construction makes it look, is
the open problem we leave.

\begin{ack}
This work was funded in part by the DARPA ANSR, CODORD, and CLARA programs under Awards FA8750-23-2-0004, HR00112590089, and HR0011269E224, respectively and by the NSF grant 2327245.
\end{ack}

\bibliographystyle{plainnat}
\bibliography{references}
\clearpage
\appendix

\paragraph{Appendix roadmap} The appendix covers the benchmark suite and per-model sizes
(App.~\ref{app:benchmarks}) with contingency and timing breakdowns (App.~\ref{app:contingency});
the architecture, notation and a worked example (Apps.~\ref{app:arch}, \ref{app:notation}, \ref{app:worked}) and the
LP formulation of the dual (App.~\ref{app:lp}); CMPE method descriptions
(App.~\ref{app:cmpe_methods}), Lagrangian bound quality (App.~\ref{app:cmpe_quality}),
wall-clock timing (App.~\ref{app:wallclock}) and the method-family split
(App.~\ref{app:t3_split}); standalone bound quality across all evidence bins
(Apps.~\ref{app:standalone_allbins}, \ref{app:standalone_pct}, \ref{app:standalone_v2}), the v1
metric (Apps.~\ref{app:v1_metric}, \ref{app:cmpe_v1}), AOBB root bounds (App.~\ref{app:aobb}),
monotonicity (App.~\ref{app:monotonicity}), the validity check (App.~\ref{app:validity}) and the
numerical floor (App.~\ref{app:numerics}); paired per-seed statistics (Apps.~\ref{app:paired},
\ref{app:cmpe_paired}); TTA on unconstrained MAP (App.~\ref{app:tta_map}), an LP-free CMPE pilot
(App.~\ref{app:nolp}) and the LP-supervised alternative (App.~\ref{app:lp_supervised}); an
$11$-variant ablation suite (App.~\ref{app:ablations}) with the evidence-encoding
(App.~\ref{app:encoding}), gated-initialization (App.~\ref{app:gated}), architecture and
training-set-size (App.~\ref{app:p7}) and seed-subset (App.~\ref{app:seed_sensitivity}) studies;
two negative results (Apps.~\ref{app:m5}, \ref{app:gnn}); and an adversarial CMPE pilot on image
classifiers (App.~\ref{app:mnist_cmpe}).

\section{Edge-message Architecture}
\label{app:arch}

\begin{figure}[ht!]
\centering
\begin{tikzpicture}[
    >=Latex,
    box/.style={draw, rounded corners=3pt, minimum height=0.7cm, minimum width=1.8cm, align=center, font=\small},
    learned/.style={box, fill=blue!12, draw=blue!50},
    computed/.style={box, fill=orange!12, draw=orange!50},
    input/.style={box, fill=green!10, draw=green!40},
    arr/.style={->, thick},
    brace/.style={decorate, decoration={brace, amplitude=6pt, mirror}},
]
\node[input] (evid) at (0,0) {Evidence $\mathbf{e}$};
\node[input] (tables) at (3.2,0) {$\theta_{\alpha_1}, \ldots, \theta_{\alpha_F}$};
\node[learned] (mlp) at (1.6, 1.3) {MLP $f_\phi$};
\node[learned] (msgs) at (1.6, 2.5) {$m_1, \ldots, m_{\mathrm{edim}}$};
\node[computed] (scatter) at (1.6, 3.8) {$\boldsymbol{\theta}' \!=\! \boldsymbol{\theta} + A\, \mathbf{m}$};
\node[font=\scriptsize\itshape, text=orange!70!black, right=0.2cm of scatter] {$\lambda_{\alpha\!\to\!\beta}\!=\!+m$, $\lambda_{\beta\!\to\!\alpha}\!=\!-m$};
\node[computed] (dual) at (1.6, 5.1) {$D(\boldsymbol{\theta}, \mathbf{m}) = \sum_\alpha \max_{x_{C_\alpha}}\, [\boldsymbol{\theta} + A\mathbf{m}]_\alpha$};
\node[box, fill=red!15, draw=red!40, font=\small\bfseries] (ub) at (1.6, 6.3) {Upper Bound $\geq \mathrm{MAP}^*$};
\draw[arr] (evid) -- (mlp);
\draw[arr] (tables) -- (mlp);
\draw[arr, blue!60] (mlp) -- node[right, font=\scriptsize, blue!70] {edge msgs} (msgs);
\draw[arr, orange!60] (msgs) -- (scatter);
\draw[arr, orange!60] (scatter) -- (dual);
\draw[arr, red!50] (dual) -- (ub);
\draw[brace, blue!60, thick] (-1.8, -0.3) -- (-1.8, 2.8) node[midway, left=8pt, font=\small\bfseries, blue!70, rotate=90, anchor=south] {Learned};
\draw[brace, orange!60, thick] (-1.8, 3.1) -- (-1.8, 6.6) node[midway, left=8pt, font=\small\bfseries, orange!70!black, rotate=90, anchor=south] {Computed (not learned)};
\end{tikzpicture}
\caption{\textbf{Edge-message architecture.} A learned MLP $f_\phi$ (blue) maps evidence $\mathbf{e}$ and factor tables $\{\theta_\alpha\}$ to edge messages $\mathbf{m} \in \mathbb{R}^{\mathrm{edim}}$, one per (edge, separator-assignment) pair on the join graph. The fixed antisymmetric scatter matrix $A$ (orange) maps $\mathbf{m}$ into cluster-table modifications: each column of $A$ adds $+m$ to one cluster and $-m$ to the neighboring cluster on a shared separator, so the two contributions cancel for every complete assignment. This restricts predictions to the antisymmetric subspace where reparameterization is valid; consequently $D(\boldsymbol{\theta}, \mathbf{m}) \geq \mathrm{MAP}^*$ holds for \emph{any} network weights (Proposition~\ref{prop:validity}), and training never has to enforce feasibility.}
\label{fig:architecture}
\end{figure}
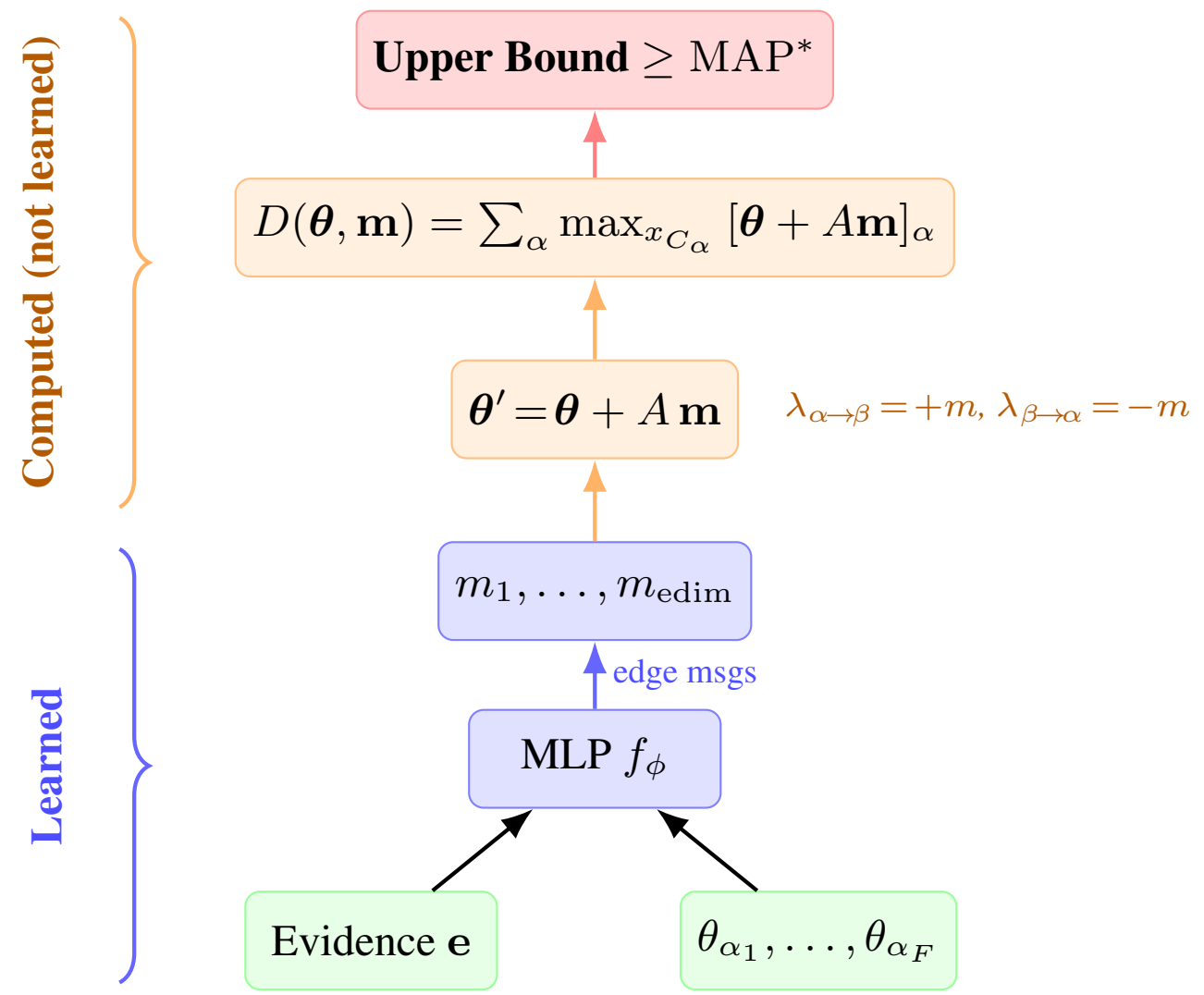

Figure~\ref{fig:architecture} sketches the parameterization.

\subsection{Notation}
\label{app:notation}
Table~\ref{tab:notation} collects the symbols used in the main body.

\begin{table}[t]\centering\small
\caption{Notation used throughout. Dimensions are for the join graph induced by
mini-bucket elimination at the chosen $i$-bound.}
\label{tab:notation}
\begin{tabular}{@{}llp{0.52\textwidth}@{}}
\toprule
\textbf{Symbol} & \textbf{Dimension} & \textbf{Meaning} \\
\midrule
$\mathcal{C}, \mathcal{E}$ & --- & clusters and edges of the join graph \\
$S_{\alpha\beta}$ & --- & separator $C_\alpha \cap C_\beta$ of edge $(\alpha,\beta)$ \\
$\theta_{C_\alpha}$ & $\prod_{i \in C_\alpha} d_i$ & log-potential table of cluster $\alpha$ \\
$\theta_{\mathrm{flat}}$ & $\lambda_{\dim}$ & all cluster tables concatenated into one vector \\
$\lambda_{\dim}$ & scalar & $\sum_\alpha \prod_{i \in C_\alpha} d_i$, total number of cluster-table entries \\
$\lambda_{\alpha\beta}$ & $\prod_{i \in S_{\alpha\beta}} d_i$ & directed edge message, constrained by $\lambda_{\alpha\beta} + \lambda_{\beta\alpha} = 0$ \\
$\mathbf{m}$ & $\mathrm{edim}$ & the network's output: one real per (undirected edge, separator assignment) \\
$\mathrm{edim}$ & scalar & $\sum_{(\alpha,\beta) \in \mathcal{E}} \prod_{i \in S_{\alpha\beta}} d_i$; typically $\mathrm{edim} \ll \lambda_{\dim}$ \\
$A$ & $\lambda_{\dim} \times \mathrm{edim}$ & fixed $\{-1,0,+1\}$ scatter map; $\theta_{\mathrm{flat}} + A\mathbf{m}$ is a reparameterization \\
$D_0$ & scalar & $\sum_\alpha \max_{x} \theta_{C_\alpha}$, the MBE bound at $\mathbf{m} = \mathbf{0}$ (after MBE-MM) \\
$D(\boldsymbol\theta, \mathbf{m})$ & scalar & $\sum_\alpha \max_{x} [\theta_{\mathrm{flat}} + A\mathbf{m}]_\alpha \;\ge\; \mathrm{MAP}^*$ \\
$D_\tau$ & scalar & log-sum-exp surrogate for $D$ at temperature $\tau$; $D_\tau \ge D$ \\
$D^{\mathrm{LP}}$ & scalar & $\min_{\mathbf{m}} D(\boldsymbol\theta, \mathbf{m})$, the LP optimum for this join graph \\
$\boldsymbol\mu$ & $K$ & Lagrange multipliers for the $K$ CMPE constraints \\
\bottomrule
\end{tabular}
\end{table}

\subsection{A Worked Three-Cluster Example}
\label{app:worked}
Table~\ref{tab:worked_example} works the example of Section~\ref{sec:background}
end to end, with the per-cluster maxima before and after the shift. Every entry
is machine-computed by the implementation we release. In code the shift is the
single product $\boldsymbol\theta' = \boldsymbol\theta + A\mathbf{m}$ with
$A \in \{-1,0,+1\}^{12\times 4}$: each column carries $-1$ on one cluster's
entries consistent with a separator value and $+1$ on its neighbour's, and a
complete assignment selects exactly one entry per cluster, so the pair cancels.
Sampling $\mathbf{m} \sim \mathcal{N}(0, 10^2)$ gives $0$ violations in $2{,}000$
draws, none of them tighter than the exact $17$ above.

\begin{table}[t]\centering\small
\caption{\textbf{A worked three-cluster example.} Four binary variables
$x_0 \ldots x_3$; clusters $C_1(x_0,x_1)$, $C_2(x_1,x_2)$, $C_3(x_2,x_3)$
chained by separators $\{x_1\}$ and $\{x_2\}$. Here $\lambda_{\dim} = 12$,
$\mathrm{edim} = 4$, and $A$ is $12 \times 4$. The true optimum is
$\mathrm{MAP}^* = 17$ at $\mathbf{x} = (1,1,1,1)$, while the unshifted bound is
$D_0 = 7 + 8 + 4 = 19$: the clusters disagree about $x_1$ and $x_2$, and that
disagreement is exactly the gap of $2$. With edge messages
$\mathbf{m} = (-3, -2, -2, -3)$ --- one value per (edge, separator assignment)
--- every cluster maximum moves and the bound becomes exact. The reparameterized
tables are $\theta'_{C_1}(x_0,x_1) = \theta_{C_1} - m_{01}(x_1)$,
$\theta'_{C_2}(x_1,x_2) = \theta_{C_2} + m_{01}(x_1) - m_{12}(x_2)$, and
$\theta'_{C_3}(x_2,x_3) = \theta_{C_3} + m_{12}(x_2)$: what one cluster gives up
on a separator, its neighbour gains, so the score of \emph{every} complete
assignment is unchanged (verified for all $16$) while the sum of maxima drops
from $19$ to $17$.}
\label{tab:worked_example}
\begin{tabular}{@{}lrrrr|lrrrr@{}}
\toprule
\multicolumn{5}{c|}{\textbf{original} $\theta$} & \multicolumn{5}{c}{\textbf{reparameterized} $\theta' = \theta + A\mathbf{m}$} \\
\midrule
$C_1(x_0,x_1)$ & $x_1{=}0$ & $x_1{=}1$ & & max & $C_1'$ & $x_1{=}0$ & $x_1{=}1$ & & max \\
$x_0{=}0$ & 2 & 5 & & & $x_0{=}0$ & 5 & 7 & & \\
$x_0{=}1$ & 1 & \textbf{7} & & 7 & $x_0{=}1$ & 4 & \textbf{9} & & 9 \\
\midrule
$C_2(x_1,x_2)$ & $x_2{=}0$ & $x_2{=}1$ & & max & $C_2'$ & $x_2{=}0$ & $x_2{=}1$ & & max \\
$x_1{=}0$ & \textbf{8} & 1 & & 8 & $x_1{=}0$ & \textbf{7} & 1 & & \\
$x_1{=}1$ & 4 & 6 & & & $x_1{=}1$ & 4 & \textbf{7} & & 7 \\
\midrule
$C_3(x_2,x_3)$ & $x_3{=}0$ & $x_3{=}1$ & & max & $C_3'$ & $x_3{=}0$ & $x_3{=}1$ & & max \\
$x_2{=}0$ & 3 & 0 & & & $x_2{=}0$ & 1 & $-2$ & & \\
$x_2{=}1$ & 2 & \textbf{4} & & 4 & $x_2{=}1$ & $-1$ & \textbf{1} & & 1 \\
\midrule
\multicolumn{4}{@{}l}{$D_0 = 7 + 8 + 4 = 19$} & & \multicolumn{4}{l}{$D(\boldsymbol\theta,\mathbf{m}) = 9 + 7 + 1 = 17 = \mathrm{MAP}^*$} & \\
\bottomrule
\end{tabular}
\end{table}

\section{CMPE Method Descriptions}
\label{app:cmpe_methods}

All four methods share the same outer loop: a $\mu$-gradient descent trajectory 
$\mu_0, \mu_1, \ldots, \mu_T$ that optimizes the Lagrangian $L(\mu)$ for the 
constrained MAP (CMPE) problem. They differ only in how JGLP is initialized 
at each $\mu$-step.

\begin{algorithm}[!htbp]
\caption{\textbf{JGLP} (zero-init baseline)}
\label{alg:m1}
\begin{algorithmic}[1]
\REQUIRE $\mu$-trajectory $\{\mu_k\}_{k=0}^{T}$, iteration budget $I$, join-graph parameters
\FOR{$k = 0, 1, \ldots, T$}
    \STATE Build join graph $\mathcal{J}_k$ from combined potentials at $\mu_k$
    \STATE $\lambda^{(0)}_k \leftarrow \mathbf{0}$ \COMMENT{zero initialization}
    \STATE Run JGLP for $I$ iterations from $\lambda^{(0)}_k$ $\rightarrow$ $\lambda^{*}_k$
    \STATE Record dual bound $D(\lambda^{*}_k)$ at checkpoints
    \STATE \textit{Discard} $\lambda^{*}_k$
\ENDFOR
\end{algorithmic}
\end{algorithm}

\begin{algorithm}[!htbp]
\caption{\textbf{NN} (neural warm-start, frozen)}
\label{alg:m2}
\begin{algorithmic}[1]
\REQUIRE $\mu$-trajectory $\{\mu_k\}_{k=0}^{T}$, iteration budget $I$, pre-trained network $f_\theta$
\FOR{$k = 0, 1, \ldots, T$}
    \STATE Build join graph $\mathcal{J}_k$ from combined potentials at $\mu_k$
    \STATE Extract base cluster tables $\lambda^{\mathrm{base}}_k$ from $\mathcal{J}_k$
    \STATE $\mathbf{m}_k \leftarrow f_\theta(\mathbf{x}_k, \mu_k)$ \COMMENT{NN predicts edge messages; weights $\theta$ frozen}
    \STATE $\lambda^{(0)}_k \leftarrow \lambda^{\mathrm{base}}_k + A\mathbf{m}_k$ \COMMENT{inject via antisymmetric reparam matrix $A$}
    \STATE Run JGLP for $I$ iterations from $\lambda^{(0)}_k$ $\rightarrow$ $\lambda^{*}_k$
    \STATE Record dual bound $D(\lambda^{*}_k)$ at checkpoints
    \STATE \textit{Discard} $\lambda^{*}_k$; \textit{do not update} $\theta$
\ENDFOR
\end{algorithmic}
\end{algorithm}

\begin{algorithm}[!htbp]
\caption{\textbf{NN+TTA} (neural warm-start with test-time adaptation; \textit{ours})}
\label{alg:m3}
\begin{algorithmic}[1]
\REQUIRE $\mu$-trajectory $\{\mu_k\}_{k=0}^{T}$, iteration budget $I$, pre-trained network $f_\theta$, TTA learning rate $\eta$, replay buffer size $B$, inner TTA steps $S$
\STATE $\theta' \leftarrow \theta$ \COMMENT{copy weights for adaptation}
\STATE $\mathcal{R} \leftarrow \emptyset$ \COMMENT{replay buffer}
\FOR{$k = 0, 1, \ldots, T$}
    \STATE Build join graph $\mathcal{J}_k$ from combined potentials at $\mu_k$
    \STATE Extract base cluster tables $\lambda^{\mathrm{base}}_k$ and $\mathrm{logz}^{\mathrm{base}}_k$ from $\mathcal{J}_k$
    \STATE $\mathbf{m}_k \leftarrow f_{\theta'}(\mathbf{x}_k, \mu_k)$ \COMMENT{NN predicts edge messages with adapted weights}
    \STATE $\lambda^{(0)}_k \leftarrow \lambda^{\mathrm{base}}_k + A\mathbf{m}_k$ \COMMENT{inject NN prediction}
    \STATE Run JGLP for $I$ iterations from $\lambda^{(0)}_k$ $\rightarrow$ $\lambda^{*}_k$
    \STATE Record dual bound $D(\lambda^{*}_k)$ at checkpoints
    \STATE $\mathcal{R} \leftarrow \mathcal{R} \cup \{(\mathbf{x}_k, \mu_k, \lambda^{\mathrm{base}}_k, \mathrm{logz}^{\mathrm{base}}_k)\}$; trim to most recent $B$ entries
    \FOR{$s = 1, \ldots, S$} \STATE \COMMENT{TTA: fine-tune NN on recent $\mu$-region}
        \STATE $\mathcal{L} \leftarrow \frac{1}{|\mathcal{R}|} \sum_{(\mathbf{x},\mu,\lambda,z) \in \mathcal{R}} D_\tau\!\bigl(f_{\theta'}(\mathbf{x}, \mu),\; \lambda,\; z\bigr)$
        \STATE $\theta' \leftarrow \theta' - \eta \nabla_{\theta'} \mathcal{L}$ \COMMENT{only NN weights update}
    \ENDFOR
    \STATE \textit{Discard} $\lambda^{*}_k$ \COMMENT{JGLP-converged $\lambda$ not carried to step $k{+}1$}
\ENDFOR
\end{algorithmic}
\end{algorithm}

\begin{algorithm}[!htbp]
\caption{\textbf{JGLP+WS} (warm-start with previous $\mu$-step's converged shift; fair-compute baseline)}
\label{alg:m4}
\begin{algorithmic}[1]
\REQUIRE $\mu$-trajectory $\{\mu_k\}_{k=0}^{T}$, iteration budget $I$
\STATE $\Delta\lambda \leftarrow \mathbf{0}$, \quad $\Delta z \leftarrow 0$ \COMMENT{no previous shift at step 0}
\FOR{$k = 0, 1, \ldots, T$}
    \STATE Build join graph $\mathcal{J}_k$ from combined potentials at $\mu_k$
    \STATE Extract initial tables $\lambda^{\mathrm{init}}_k$ and $z^{\mathrm{init}}_k$ from $\mathcal{J}_k$
    \IF{$k > 0$}
        \STATE $\lambda^{(0)}_k \leftarrow \lambda^{\mathrm{init}}_k + \Delta\lambda$ \COMMENT{apply previous JGLP shift}
        \STATE $z^{(0)}_k \leftarrow z^{\mathrm{init}}_k + \Delta z$
    \ELSE
        \STATE $\lambda^{(0)}_k \leftarrow \lambda^{\mathrm{init}}_k$ \COMMENT{step 0: equivalent to JGLP}
    \ENDIF
    \STATE Run JGLP for $I$ iterations from $\lambda^{(0)}_k$ $\rightarrow$ $\lambda^{*}_k$, \quad $z^{*}_k$
    \STATE Record dual bound $D(\lambda^{*}_k)$ at checkpoints
    \STATE $\Delta\lambda \leftarrow \lambda^{*}_k - \lambda^{\mathrm{init}}_k$ \COMMENT{save what JGLP learned this step}
    \STATE $\Delta z \leftarrow z^{*}_k - z^{\mathrm{init}}_k$
\ENDFOR
\end{algorithmic}
\end{algorithm}

\paragraph{Key differences.}
JGLP and NN are \emph{memoryless}: each $\mu$-step starts fresh (from zero or from a single NN forward pass). 
NN+TTA carries state via the NN weights $\theta'$ and the replay buffer $\mathcal{R}$, but discards JGLP-converged $\lambda$ after each step.
JGLP+WS carries state via the converged cluster shift $(\Delta\lambda, \Delta z)$, but uses no neural network.
These represent two orthogonal adaptation channels: learned prediction (NN+TTA) versus direct $\lambda$-continuity (JGLP+WS).
On Bayesian networks where $\mu$ converges quickly, $\lambda$-continuity dominates (JGLP+WS $<$ NN+TTA).
On grids and segmentation models where $\mu$ keeps moving, stale shifts destabilize JGLP and the neural structural prior proves essential (NN+TTA $\ll$ JGLP+WS).

\subsection{LP Formulation of the JGLP Dual}
\label{app:lp}

Let $\boldsymbol{\theta}$ be the flat concatenation of cluster log-tables after mini-bucket decomposition (length $\lambda_{\dim}$) and let $A \in \mathbb{R}^{\lambda_{\dim} \times e_{\dim}}$ be the antisymmetric scatter matrix mapping edge messages $\mathbf{m} \in \mathbb{R}^{e_{\dim}}$ onto cluster entries. The JGLP dual objective is
$D(\boldsymbol{\theta}, \mathbf{m}) = \sum_c \max_{x_c} \left[\boldsymbol{\theta}_{c,x_c} + (A\mathbf{m})_{c,x_c}\right]$.
The LP relaxation (epigraph form) is
\[
\min_{\mathbf{m}, \mathbf{t}}\ \sum_c t_c \quad \text{s.t.}\ t_c \geq \boldsymbol{\theta}_{c,x_c} + (A\mathbf{m})_{c,x_c}\quad \forall c, x_c,
\]
with $D_{\text{LP}} = \min \sum_c t_c$. We solve this with CVXPY's CLARABEL backend (fallback: ECOS, SCS). On grid10$\times$10 at $i$-bound 4 ($\lambda_{\dim} = 3{,}854$, $e_{\dim} = 1{,}904$), CLARABEL returns the LP optimum in 2.9s; JGLP at 500 iterations converges to a value $8.6 \times 10^{-4}$ above the LP optimum, demonstrating that JGLP's coordinate-descent can stop short of global optimality.

\subsection{Full MAP Results (All 37 PGMs)}
\label{app:table1_full}
Figure~\ref{fig:convergence} repeats the main-body convergence plot for reference.
Table~\ref{tab:table1_full} lists all 37 evaluated PGM \emph{rows} in the new
$\log_{10}|D - D_{\text{LP}}|$ format, sorted by neural win rate at iter=10.
Promedas models exhibit small dynamic range (log-error $\sim 0$) because their
dual objectives are near-zero, which compresses the relative scale.

\paragraph{Duplicate UAI files across names.} The UAI collection ships four of
these graphs twice, under both a \texttt{Grids\_NN} name and a
\texttt{grid}$R{\times}C$\texttt{.fK} name. The file pairs are byte- or
token-identical, so \textbf{these 37 rows are 33 distinct models}:
\texttt{Grids\_29} $=$ \texttt{grid20$\times$20.f5.wrap},
\texttt{Grids\_30} $=$ \texttt{grid20$\times$20.f10},
\texttt{Grids\_18} $=$ \texttt{grid40$\times$40.f10.wrap}, and
\texttt{Grids\_25} $=$ \texttt{grid40$\times$40.f2}. We keep both rows of each
pair because each carries an independently trained network, but no count in this
paper treats them as separate models. The same four collisions make the
$40$-row benchmark suite of Table~\ref{tab:benchmark_suite} $36$ distinct
models.

\paragraph{Per-row evaluation setting.} The main suite uses $i$-bound~4 with a
$20\%$ evidence fraction throughout. Fourteen of the appendix-only rows do not,
and we list them rather than let the table imply a single protocol: the ten
Promedas rows use a $40\%$ evidence fraction (and \texttt{Promedas\_51}
$i$-bound~3), and the four largest grids use a reduced $i$-bound with $50\%$
evidence --- $i$-bound~3 for \texttt{grid40$\times$40.f2} and $i$-bound~2 for
\texttt{grid40$\times$40.f10.wrap}, \texttt{grid80$\times$80.f10} and
\texttt{grid80$\times$80.f5.wrap} --- because the join graph at $i$-bound~4 is
not tractable for them at $20\%$ evidence. This matters for reading the table:
a weaker relaxation on a more heavily conditioned problem is easier for MBE to
be LP-optimal on, which is why those four rows show near-zero gaps while their
standard-protocol twins (\texttt{Grids\_18}, \texttt{Grids\_25}) show gaps of
$14.3$ and $4.94$ nats on the identical files. Rows evaluated under different
settings should not be compared with each other.

\begin{table}[ht]
\caption{Full MAP bound results: $\log_{10}|D - D_{\text{LP}}|$ at fixed JGLP iterations on 37 PGMs. BN\_126 column blanked due to LP ill-conditioning (CLARABEL returned status `optimal\_inaccurate').}
\label{tab:table1_full}
\centering
\scriptsize
\begin{tabular}{@{}l|r|cc|cc|cc|cc|cc|r@{}}
\toprule
& & \multicolumn{2}{c|}{iter=0} & \multicolumn{2}{c|}{iter=10} & \multicolumn{2}{c|}{iter=50} & \multicolumn{2}{c|}{iter=100} & \multicolumn{2}{c|}{conv} & \\
\textbf{PGM} & $D_{\text{LP}}$ & J & N & J & N & J & N & J & N & J & N & win@10 \\
\midrule
BN\_12 & $-31.64$ & -0.02 & -0.17 & -0.22 & -0.48 & -0.74 & -1.05 & -1.65 & -2.04 & -6.79 & -6.77 & 100\% \\
BN\_13 & $-44.93$ & -0.04 & -0.19 & -0.29 & -0.60 & -0.97 & -1.15 & -1.34 & -1.55 & -7.17 & -6.98 & 100\% \\
BN\_14 & $-38.83$ & -0.34 & -0.39 & -0.66 & -0.83 & -1.23 & -1.33 & -4.22 & -4.52 & -6.49 & -6.42 & 100\% \\
Seg\_12 & $-114.24$ & 0.21 & -0.05 & -0.01 & -1.01 & -4.13 & -4.94 & -7.10 & -6.69 & -7.10 & -6.69 & 100\% \\
Seg\_14 & $-136.64$ & 0.18 & -0.42 & -0.73 & -2.00 & -6.34 & -4.62 & -6.34 & -6.13 & -6.34 & -6.19 & 100\% \\
Seg\_16 & $-132.72$ & 0.58 & 0.13 & 0.29 & -0.49 & -0.67 & -3.88 & -6.07 & -5.71 & -6.23 & -6.10 & 100\% \\
PF\_11 & $4145$ & 1.90 & 1.70 & 1.34 & 1.29 & 0.85 & 0.84 & 0.55 & 0.51 & -0.40 & -0.40 & 100\% \\
Grids\_18 & $11284$ & 1.84 & 1.84 & 1.14 & 1.13 & 0.35 & 0.33 & 0.05 & 0.04 & -2.04 & -2.01 & 100\% \\
Grids\_19 & $2327$ & 1.02 & 1.01 & 0.23 & 0.13 & -1.87 & -1.86 & -6.19 & -4.34 & -6.41 & -6.23 & 100\% \\
Grids\_21 & $16907$ & 2.02 & 2.00 & 1.31 & 1.30 & 0.51 & 0.50 & 0.23 & 0.23 & -1.73 & -1.74 & 100\% \\
Grids\_23 & $5618$ & 1.51 & 1.52 & 1.16 & 1.14 & 0.37 & 0.38 & -0.06 & -0.04 & -2.11 & -2.15 & 100\% \\
Grids\_24 & $16740$ & 1.99 & 1.99 & 1.68 & 1.67 & 1.00 & 1.00 & 0.56 & 0.56 & -0.89 & -0.93 & 100\% \\
Grids\_25 & $2332$ & 1.15 & 1.10 & 0.70 & 0.53 & -0.26 & -0.46 & -0.99 & -1.17 & -4.90 & -4.91 & 100\% \\
Grids\_30 & $2767$ & 1.71 & 1.50 & 1.17 & 1.11 & 0.66 & 0.62 & 0.42 & 0.38 & -0.41 & -0.40 & 100\% \\
grid20$\times$20.f10 & $2767$ & 1.71 & 1.51 & 1.17 & 1.11 & 0.66 & 0.62 & 0.42 & 0.38 & -0.41 & -0.40 & 100\% \\
Grids\_22 & $5635$ & 1.48 & 1.49 & 0.86 & 0.85 & -0.14 & -0.17 & -0.88 & -0.96 & -3.92 & -4.00 & 90\% \\
Grids\_27 & $11174$ & 1.82 & 1.81 & 1.50 & 1.49 & 0.83 & 0.81 & 0.35 & 0.33 & -0.98 & -1.06 & 90\% \\
grid20$\times$20.f5 & $1322$ & 0.89 & 0.87 & 0.07 & 0.03 & -1.05 & -1.05 & -8.79 & -6.00 & -9.24 & -6.82 & 90\% \\
Grids\_29 & $1322$ & 0.89 & 0.87 & 0.07 & 0.03 & -1.05 & -1.05 & -8.79 & -5.60 & -9.24 & -6.78 & 80\% \\
Promedas\_51 & $-8.29$ & 0.08 & -0.22 & 0.08 & -1.64 & 0.08 & -2.80 & 0.08 & -3.71 & 0.08 & -6.19 & 80\% \\
Seg\_18 & $-111.99$ & -0.24 & -0.06 & -0.66 & -0.87 & -3.35 & -3.79 & -6.02 & -6.15 & -6.02 & -6.15 & 80\% \\
BN\_9 & $-39.98$ & -0.59 & -0.60 & -0.97 & -1.53 & -2.71 & -5.77 & -6.68 & -6.94 & -6.68 & -6.98 & 70\% \\
Grids\_26 & $2634$ & 1.23 & 1.18 & 0.38 & 0.39 & -1.12 & -1.12 & -8.46 & -6.25 & -8.46 & -6.74 & 60\% \\
Promedas\_32 & $-5.61$ & 0.54 & 0.54 & 0.54 & -0.56 & 0.54 & -2.53 & 0.54 & -3.20 & 0.54 & -6.01 & 60\% \\
Promedas\_33 & $-4.14$ & 0.37 & 0.28 & 0.37 & -0.72 & 0.37 & -1.47 & 0.37 & -1.85 & 0.37 & -4.05 & 60\% \\
Grids\_28 & $3954$ & 1.36 & 1.34 & 0.72 & 0.73 & -0.90 & -0.90 & -8.56 & -5.62 & -8.56 & -6.76 & 50\% \\
Promedas\_44 & $-3.82$ & -2.79 & -0.20 & -3.35 & -1.59 & -6.27 & -2.64 & -7.72 & -3.23 & -7.72 & -6.23 & 50\% \\
Promedas\_40 & $-3.77$ & -5.69 & -0.25 & -5.69 & -1.69 & -7.62 & -2.96 & -7.62 & -3.86 & -7.62 & -6.34 & 40\% \\
Seg\_19 & $-93.60$ & -0.41 & -0.24 & -3.42 & -1.16 & -5.56 & -3.97 & -5.71 & -5.62 & -6.25 & -6.60 & 33\% \\
Promedas\_21 & $-7.69$ & -6.14 & 0.40 & -6.14 & -1.13 & -6.14 & -2.44 & -6.14 & -3.40 & -6.14 & -6.15 & 30\% \\
grid80$\times$80.f10 & $28287$ & 0.65 & 0.73 & -0.52 & -0.64 & -6.38 & -4.60 & -7.85 & -6.99 & -7.85 & -7.01 & 10\% \\
Promedas\_38 & $-1.51$ & -7.54 & -0.30 & -7.54 & -1.26 & -7.54 & -2.02 & -7.54 & -2.54 & -7.54 & -5.55 & 0\% \\
grid10x10 & $335.89$ & -8.78 & -0.83 & -8.78 & -3.04 & -8.78 & -7.35 & -8.78 & -7.35 & -8.78 & -7.35 & 0\% \\
grid40$\times$40.f10 & $7220$ & -4.68 & -0.46 & -9.03 & -3.45 & -9.03 & -7.40 & -9.03 & -7.40 & -9.03 & -7.40 & 0\% \\
grid40$\times$40.f2 & $1478$ & -8.88 & -0.44 & -9.34 & -3.33 & -9.34 & -7.06 & -9.34 & -7.36 & -9.34 & -7.36 & 0\% \\
grid80$\times$80.f5 & $14246$ & -2.78 & 0.05 & -6.83 & -3.00 & -8.17 & -4.98 & -8.17 & -6.26 & -8.17 & -6.77 & 0\% \\
\bottomrule
\end{tabular}
\end{table}

\subsection{CMPE Lagrangian Bound Quality}
\label{app:cmpe_quality}
Table~\ref{tab:cmpe_lagrangian} reports the Lagrangian bound at $\mu^*$ against the best MAP bound, and Table~\ref{tab:cmpe_ub} compares it with CMPE\_UB~\citep{rahman2021novel}.

\begin{table}[ht]
\caption{CMPE bounds: MAP bounds ($\mu = 0$) vs Lagrangian relaxation at optimal $\mu^*$. The CMPE Lagrangian exploits the constraint to yield substantially tighter upper bounds. Lower = tighter.}
\label{tab:cmpe_lagrangian}
\centering
\small
\begin{tabular}{@{}lcccc@{}}
\toprule
\textbf{PGM} & \textbf{MBE} & \textbf{JGLP} & \textbf{CMPE LR} & \textbf{Improvement} \\
\midrule
BN\_12 (90v) & $-30.74$ & $-31.45$ & $-64.80$ & $33.4$ \\
BN\_9 (105v) & $-39.20$ & $-39.34$ & $-79.82$ & $40.5$ \\
BN\_14 (115v) & $-38.67$ & $-39.11$ & $-82.60$ & $43.5$ \\
Seg\_12 & $-117.40$ & $-118.75$ & $-434.34$ & $316.2$ \\
Seg\_14 & $-129.12$ & $-130.76$ & $-478.06$ & $347.4$ \\
grid10$\times$10 & $348.07$ & $347.76$ & $43.03$ & $305.2$ \\
\bottomrule
\end{tabular}
\end{table}

\begin{table}[ht]
\caption{Our Lagrangian relaxation (JGLP, $i$-bound 4) vs CMPE\_UB \citep{rahman2021novel} ($i$-bound 10). Our method produces tighter bounds in less time.}
\label{tab:cmpe_ub}
\centering
\small
\begin{tabular}{@{}lcccc@{}}
\toprule
\textbf{PGM} & \textbf{Our LR} & \textbf{CMPE\_UB (JG)} & \textbf{Our time} & \textbf{UB time} \\
\midrule
BN\_12 & $-64.48$ & $-41.43$ & 48.8s & 112.6s \\
BN\_9 & $-79.15$ & $-54.59$ & 30.1s & 54.7s \\
BN\_14 & $-80.83$ & $-54.44$ & 41.7s & 88.4s \\
\bottomrule
\end{tabular}
\end{table}

\subsection{Full Standalone Bound Results (All Bins)}
\label{app:standalone_allbins}
\begin{table}[ht]
\caption{Full standalone bound results across all PGMs and evidence bins. ``\% Better'' = fraction of test instances where neural bound is tighter than MBE. Bin 2 typically has high evidence ratios where the problem approaches exact solutions.}
\centering
\small
\begin{tabular}{@{}lccc@{}}
\toprule
\textbf{PGM} & \textbf{Bin} & \textbf{$e_{\dim}$} & \textbf{\% Better} \\
\midrule
\multirow{3}{*}{BN\_12} & 0 & 2,130 & 94\% \\
& 1 & 1,076 & 84\% \\
& 2 & 298 & 0\% \\
\midrule
\multirow{3}{*}{BN\_9} & 0 & 1,354 & 80\% \\
& 1 & 828 & 72\% \\
& 2 & 220 & 0\% \\
\midrule
\multirow{3}{*}{BN\_14} & 0 & 1,812 & 92\% \\
& 1 & 1,086 & 92\% \\
& 2 & 298 & 8\% \\
\midrule
\multirow{3}{*}{grid10$\times$10} & 0 & 808 & 68\% \\
& 1 & 102 & 0\% \\
& 2 & 20 & 46\% \\
\midrule
\multirow{3}{*}{Seg\_12} & 0 & 2,594 & 94\% \\
& 1 & 1,558 & 0\% \\
& 2 & 72 & 4\% \\
\midrule
\multirow{3}{*}{Seg\_14} & 0 & 3,046 & 100\% \\
& 1 & 1,896 & 12\% \\
& 2 & 88 & 0\% \\
\midrule
\multirow{3}{*}{Seg\_18} & 0 & 3,414 & 74\% \\
& 1 & 1,550 & 50\% \\
& 2 & 92 & 2\% \\
\midrule
\multirow{3}{*}{grid20$\times$20.f5} & 0 & 2,988 & 20\% \\
& 1 & 442 & 32\% \\
& 2 & 76 & 32\% \\
\midrule
\multirow{3}{*}{grid20$\times$20.f10} & 0 & 5,002 & 98\% \\
& 1 & 678 & 0\% \\
& 2 & 98 & 16\% \\
\midrule
grid40$\times$40.f10 & 0 & 1,400 & 50\% \\
grid40$\times$40.f2 & 0 & 1,664 & 0\% \\
grid80$\times$80.f10 & 0 & 5,726 & 82\% \\
grid80$\times$80.f5 & 0 & 5,262 & 36\% \\
\bottomrule
\end{tabular}
\end{table}

\subsection{CMPE Application Details}
\label{app:cmpe}

\paragraph{Scope of this subsection: the $\sigma = 0.1$ diagnostic regime.}
The results in this subsection were obtained with constraint noise
$\sigma = 0.1$, an early diagnostic setting that we \emph{do not} use for any
reported result. At $\sigma = 0.1$ we have $\theta^c \approx \theta$, so the
combined Lagrangian $\theta - \mu^\ast \theta^c$ at $\mu^\ast \approx 1$
collapses to pure noise: the constraint is effectively inactive and the
combined model is out of distribution for any network trained on $\theta$.
Every headline CMPE result uses $\sigma = 0.5$, which keeps the constraint
active at the optimum ($\mu^\ast > 0$). Conclusions drawn here therefore
characterize the degenerate $\sigma = 0.1$ regime only and do not transfer to
the $\sigma = 0.5$ setting used in the main body.

\paragraph{CMPE formulation.} Given two log-linear models $f$ and $g$ over the same graph, the constrained MPE (CMPE) problem is $\max_{\mathbf{x}} f(\mathbf{x}) \text{ s.t. } g(\mathbf{x}) \leq q$. The Lagrangian relaxation yields:
\begin{equation}
L(\mu) = \max_{\mathbf{x}} [f(\mathbf{x}) - \mu(g(\mathbf{x}) - q)] = D_\mu(\lambda) + \mu q, \quad \mu \geq 0
\end{equation}
where $D_\mu(\lambda)$ is the MAP dual bound for the combined model $M_\mu \equiv f - \mu g$. Since any feasible $\mathbf{x}^*$ satisfies $g(\mathbf{x}^*) \leq q$, we have $-\mu(g(\mathbf{x}^*) - q) \geq 0$ for $\mu \geq 0$, so $L(\mu) \geq f(\mathbf{x}^*) \geq \text{CMPE}^*$. Our framework computes $D_\mu(\lambda)$ for any $\mu$ by building the join graph for $M_\mu$ and applying JGLP.

\paragraph{$\mu$ search.} We find $\mu^* = \arg\min_{\mu \geq 0} L(\mu)$ via golden-section search over $[0, 5]$ with 15 function evaluations. For our test instances ($g = f + \epsilon$, $\sigma = 0.1$), the optimal multiplier concentrates at $\mu^* \approx 0.97$ because the combined model $f - \mu g$ at $\mu = 1$ reduces to $-\epsilon$ (pure noise), making the bound very tight.

\paragraph{Gradient vs JGLP for inner solver.} Table~\ref{tab:cmpe_inner} shows that JGLP-200 consistently outperforms gradient descent (Adam, 200 iterations on $D_\tau$ with $\tau = 0.001$) for the inner MAP problem at each $\mu$. JGLP's combinatorial message-passing directly tightens $D$ while gradient descent optimizes the smooth surrogate $D_\tau$, which can temporarily worsen $D$.

\begin{table}[ht]
\caption{Inner solver comparison for CMPE: JGLP-200 vs gradient descent (200 iterations) on the combined model $M_{\mu^*}$. All methods use $i$-bound 4. Lower = tighter CMPE bound.}
\label{tab:cmpe_inner}
\centering
\small
\begin{tabular}{@{}lccccc@{}}
\toprule
\textbf{PGM} & \textbf{MBE} & \textbf{JGLP-200} & \textbf{Grad-200} & \textbf{N+Grad} & \textbf{CMPE LR} \\
\midrule
BN\_12 & $-64.19$ & $-64.46$ & $-64.30$ & $-64.30$ & $-64.63$ \\
BN\_9 & $-80.05$ & $-80.21$ & $-80.11$ & $-80.11$ & $-80.32$ \\
BN\_14 & $-81.18$ & $-81.51$ & $-81.37$ & $-81.28$ & $-81.66$ \\
\bottomrule
\end{tabular}
\end{table}

\paragraph{Neural $\mu$ prediction.} We train an MLP to predict $\mu^*$ directly from the evidence encoding, but find that for $\sigma = 0.1$ instances, the combined model at $\mu^* \approx 1$ is near-zero-mean noise, which is out-of-distribution for edge-message models trained on the original $f$. Consequently, in this degenerate regime the neural warm-start of edge messages provides no advantage over zero-initialization. This is a statement about $\sigma = 0.1$ specifically, where the constraint is inactive and the combined model carries no signal; it is \emph{not} a general finding about CMPE, and it does not hold at the $\sigma = 0.5$ setting used throughout the main body, where the neural warm-start does help.

\paragraph{Gap analysis.} Table~\ref{tab:cmpe_gaps} shows the gap between upper bounds and the best feasible solution (lower bound) found by random search. The CMPE Lagrangian reduces the gap by 5--7$\times$ compared to MAP bounds and 4--5$\times$ compared to CMPE\_UB, demonstrating that the Lagrangian relaxation effectively exploits the constraint structure.

\begin{table}[ht]
\caption{Gap from best feasible solution (lower bound). Smaller gap = tighter bound. CMPE LR provides dramatically tighter certificates than all MAP-based bounds.}
\label{tab:cmpe_gaps}
\centering
\small
\begin{tabular}{@{}lccccc@{}}
\toprule
\textbf{PGM} & \textbf{MBE} & \textbf{JGLP} & \textbf{Neural} & \textbf{CMPE LR} & \textbf{CMPE\_UB} \\
\midrule
BN\_12 & 41.30 & 40.60 & 40.71 & \textbf{6.64} & 29.70 \\
BN\_9 & 46.84 & 46.54 & 46.62 & \textbf{6.96} & 32.80 \\
BN\_14 & 52.72 & 52.23 & 52.35 & \textbf{8.02} & 34.89 \\
\bottomrule
\end{tabular}
\end{table}

\subsection{CMPE Wall-Clock Timing}
\label{app:wallclock}

\begin{table}[ht]
\caption{CMPE gradient-descent wall-clock time (single constraint, 50 $\mu$
steps). JGLP runs 50 iterations per inner solve; NN uses a single forward pass.
$L^*$ is an \emph{upper} bound on $\mathrm{CMPE}^*$, so \textbf{lower is
tighter}; $\Delta L^*$ is NN minus JGLP, and a positive value means the NN
bound is that many nats \emph{looser}. The speedup is therefore not a free
win: on BN\_14 and Seg\_14 the single forward pass is $12.6\times$ and
$8.8\times$ faster but returns a bound $42.0$ and $317$ nats looser than
JGLP's, which is not a usable trade for those two models. This table reports
throughput at whatever bound quality the forward pass happens to give; the
bound-quality comparison at matched budgets is
Tables~\ref{tab:t3_main_singlepass}--\ref{tab:t3_main_crossmu}. These runs use
the $\sigma = 0.1$ diagnostic regime of App.~\ref{app:cmpe}, not the
$\sigma = 0.5$ setting of the main body.}
\label{tab:cmpe_gradient}
\centering
\small
\begin{tabular}{@{}lcc|cc|cc@{}}
\toprule
& \multicolumn{2}{c|}{\textbf{JGLP}} & \multicolumn{2}{c|}{\textbf{NN}} & & \\
\textbf{PGM} & $L^*$ & time & $L^*$ & time & $\Delta L^*$ & \textbf{Speedup} \\
\midrule
BN\_12         & $-62.0$ & 22.8s & $-54.2$ & 1.8s & $+7.8$  & 12.9$\times$ \\
BN\_9          & $-80.6$ & 17.3s & $-78.3$ & 1.5s & $+2.3$  & 11.5$\times$ \\
BN\_14         & $-80.0$ & 21.7s & $-38.0$ & 1.7s & $+42.0$ & 12.6$\times$ \\
Seg\_12        & $-261$  & 31.2s & $-224$  & 4.1s & $+37$   & $\phantom{0}$7.7$\times$ \\
Seg\_14        & $-458$  & 37.3s & $-141$  & 4.3s & $+317$  & $\phantom{0}$8.8$\times$ \\
grid10$\times$10 & $339$ & $\phantom{0}$9.6s & $341$   & 1.4s & $+2$ & $\phantom{0}$6.7$\times$ \\
\bottomrule
\end{tabular}
\end{table}

\begin{table}[ht]
\caption{Two-constraint CMPE wall-clock time (50 gradient steps on
$(\mu_1, \mu_2)$). Conventions as in Table~\ref{tab:cmpe_gradient}: $L^*$ is an
upper bound on $\mathrm{CMPE}^*$ (lower is tighter) and $\Delta L^*$ is NN
minus JGLP, so positive means the NN bound is looser. Here the two arms agree
to within a few nats on five of six models, and on BN\_14 the NN bound is
$19.1$ nats \emph{tighter}. Note that the $L^*$ values are not comparable
across Tables~\ref{tab:cmpe_gradient} and~\ref{tab:cmpe_two}: adding a second
constraint changes the constrained problem, so BN\_14's JGLP optimum moves
from $-80.0$ ($K=1$) to $-59.1$ ($K=2$). The $K=2$ Lagrangian bound being
\emph{looser} than the $K=1$ bound on BN\_14, despite a strictly smaller
feasible set, indicates that the $(\mu_1,\mu_2)$ search has not reached the
optimal multiplier pair on that model rather than any property of the
relaxation. These runs use the $\sigma = 0.1$ diagnostic regime of
App.~\ref{app:cmpe}.}
\label{tab:cmpe_two}
\centering
\small
\begin{tabular}{@{}lcc|cc|cc@{}}
\toprule
& \multicolumn{2}{c|}{\textbf{JGLP}} & \multicolumn{2}{c|}{\textbf{NN}} & & \\
\textbf{PGM} & $L^*$ & time & $L^*$ & time & $\Delta L^*$ & \textbf{Speedup} \\
\midrule
BN\_12         & $-60.9$ & 22.7s & $-58.3$ & 1.9s & $+2.6$  & 11.9$\times$ \\
BN\_9          & $-79.5$ & 17.3s & $-78.8$ & 1.5s & $+0.7$  & 11.6$\times$ \\
BN\_14         & $-59.1$ & 21.9s & $-78.2$ & 1.8s & $-19.1$ & 12.4$\times$ \\
Seg\_12        & $-122$  & 31.7s & $-121$  & 4.1s & $+1$    & $\phantom{0}$7.7$\times$ \\
Seg\_14        & $-144$  & 37.1s & $-143$  & 4.2s & $+1$    & $\phantom{0}$8.8$\times$ \\
grid10$\times$10 & $339$ & $\phantom{0}$9.7s & $341$   & 1.4s & $+2$ & $\phantom{0}$6.8$\times$ \\
\bottomrule
\end{tabular}
\end{table}

Table~\ref{tab:breakeven_resolved} gives the full accounting behind the
amortized-cost discussion in Section~\ref{sec:exp-map}. Each cost component
(join-graph build, one JGLP iteration, one forward pass, one replay-buffer TTA
inner loop, one-off training) is measured directly on an unloaded machine, and
combined with the iteration at which each individual run first reaches the
target gap.

Three things about how this is measured are worth stating, because an earlier
version of this analysis got all three wrong and reported a much more
pessimistic break-even as a result. First, the crossing iteration is recorded
\emph{exactly}, by instrumenting JGLP at every iteration, rather than being
rounded up to the next checkpoint in $\{0,10,50,100,500\}$ --- at checkpoint
granularity both arms frequently land in the same bucket, the iteration term
cancels, and the comparison silently degenerates to a difference of fixed
overheads. Second, every comparison is \emph{paired}: all three arms solve the
identical query, so the saving is computed per query and then aggregated,
never as a difference of two marginal medians. Third, runs that never reach
the target are censored and counted rather than dropped from the median; this
matters because the arms censor at very different rates (on BN\_12, $49.8\%$
for zero-init against $30.9\%$ for NN+TTA), so a median over survivors
compares the baseline on a strictly easier subsample of queries.

\begin{table*}[t]\centering\scriptsize
\caption{\textbf{Does the amortization pay for itself? ($K{=}1$, target $10^{-2}$.)} Iteration counts are the \emph{exact} iteration at which the gap to the LP optimum first falls below the target, not a checkpoint bucket, and every comparison is paired: all three arms solve the identical query (same seed, same $\mu$-step, same $D^{\mathrm{LP}}$). Runs that never reach the target are counted in the censoring columns rather than dropped from the medians, which matters because the arms censor at very different rates. ``win\%'' is the fraction of paired queries on which the arm reaches the target in strictly fewer iterations than zero-init. Break-even is the one-off training cost divided by the paired per-query saving; $\infty$ means the arm has no per-query saving at this target, so training is never recovered. Rows above the rule are the main suite. Only $5$ of these $30$ (model, arm) cells have a finite break-even, and the four Bayesian networks are the only ones that reach it at a practical query count; we report this rather than claim wall-clock dominance. The per-iteration re-run was done for $K{=}1$; $K{=}2$ is covered by the bound-quality tables.}
\label{tab:breakeven_resolved}
\resizebox{\textwidth}{!}{%
\begin{tabular}{@{}lr|r|rr|rr|rr|rr@{}}
\toprule
 & & M1 & \multicolumn{2}{c|}{median iters} & \multicolumn{2}{c|}{paired win\%} & \multicolumn{2}{c|}{censored (\%)} & \multicolumn{2}{c@{}}{break-even (queries)} \\
\textbf{PGM} & $n$ & iters & M2 & M3 & M2 & M3 & M1 & M3 & M2 & M3 \\
\midrule
BN\_12 & 120 & 201 & 206 & 136 & 66 & 31 & 60 & 43 & 15{,}278 & $\infty$ \\
BN\_13 & 120 & 69 & 69 & 18 & 66 & 1 & 2 & 2 & 26{,}968 & $\infty$ \\
BN\_14 & 120 & 109 & 41 & 50 & 98 & 43 & 1 & 0 & 3{,}215 & $\infty$ \\
BN\_9 & 120 & 49 & 26 & 8 & 58 & 0 & 0 & 0 & 19{,}805 & $\infty$ \\
Grids\_26 & 120 & 52 & 76 & 40 & 0 & 10 & 0 & 0 & $\infty$ & $\infty$ \\
Grids\_28 & 120 & 60 & 76 & 45 & 0 & 0 & 0 & 0 & $\infty$ & $\infty$ \\
Grids\_29 & 120 & 57 & 66 & 39 & 50 & 0 & 0 & 0 & $\infty$ & $\infty$ \\
Seg\_12 & 120 & 33 & 28 & 17 & 40 & 0 & 0 & 0 & $\infty$ & $\infty$ \\
Seg\_14 & 120 & 20 & 39 & 24 & 28 & 0 & 0 & 0 & $\infty$ & $\infty$ \\
Seg\_16 & 120 & 36 & 41 & 27 & 28 & 0 & 0 & 0 & $\infty$ & $\infty$ \\
Seg\_18 & 120 & 77 & 64 & 65 & 48 & 2 & 0 & 0 & $1.5{\times}10^{6}$ & $\infty$ \\
Seg\_19 & 120 & 42 & 52 & 33 & 27 & 0 & 0 & 0 & $\infty$ & $\infty$ \\
grid10$\times$10.f10 & 120 & 0 & 44 & 16 & 17 & 5 & 0 & 0 & $\infty$ & $\infty$ \\
grid10$\times$10.f5 & 120 & 10 & 22 & 8 & 33 & 0 & 0 & 0 & $\infty$ & $\infty$ \\
grid20$\times$20.f5 & 120 & 57 & 62 & 40 & 50 & 0 & 0 & 0 & $\infty$ & $\infty$ \\
\midrule
Promedas\_38 & 120 & 0 & 16 & 8 & 0 & 0 & 0 & 0 & $\infty$ & $\infty$ \\
Promedas\_44 & 120 & 0 & 16 & 9 & 0 & 0 & 0 & 0 & $\infty$ & $\infty$ \\
Promedas\_51 & 120 & 0 & 8 & 6 & 0 & 0 & 0 & 0 & $\infty$ & $\infty$ \\
grid40$\times$40.f10 & 120 & 0 & 24 & 10 & 0 & 0 & 0 & 0 & $\infty$ & $\infty$ \\
grid40$\times$40.f2 & 120 & 0 & 24 & 4 & 0 & 0 & 0 & 0 & $\infty$ & $\infty$ \\
\bottomrule\end{tabular}}
\end{table*}

\paragraph{Cost of adaptation versus replay batch size.}
Table~\ref{tab:tta_batchsize} isolates how the cost of test-time adaptation
scales with the replay batch size $B$, on one small and one large PGM. The
cost is strongly sublinear in $B$: entering the inner loop at all is what is
expensive, while quadrupling $B$ from $5$ to $20$ adds only $50\%$ on the small
PGM and $19\%$ on the large one. A practitioner choosing $B$ is therefore
trading very little wall-clock for a larger replay buffer. Whether the cost is
repaid at all is a separate question, and the lower panel answers it
negatively in most cells: adaptation buys back its cost only on the large PGM
at the loosest target ($10^{-1}$), and at tight targets the adapted run reaches
the target in \emph{more} iterations than zero-init, so there is no budget to
spend. This is the same boundary the rest of the paper reports --- the benefit
of the learned initialization is concentrated early, and both arms descend the
same convex objective to the same optimum.

\begin{table}[t]\centering\scriptsize
\caption{\textbf{Cost of replay-buffer test-time adaptation as a function of the replay batch size $B$.} \emph{Top}: median wall-clock of one TTA step (inner loop of $10$ steps), interquartile range in brackets, over $15$ interleaved rounds --- one timing per $B$ per round, so machine-load drift cannot order the arms. $B{=}0$ is the no-adaptation baseline (forward pass only), which is why it matches the forward-pass latency rather than scaling with $B$. Cost is strongly sublinear in $B$: almost all of it is the fixed cost of entering the inner loop, and quadrupling $B$ from $5$ to $20$ adds $50\%$ on the small PGM and $19\%$ on the large one. \emph{Bottom}: whether that cost is repaid. ``Iterations saved'' is $\mathrm{M1}$ minus $\mathrm{M3}$ iterations to first reach the target; the budget is those iterations priced at the measured per-iteration cost; ``largest affordable $B$'' is the largest tested $B$ fitting in it, and the break-even $B$ extrapolates where cost meets budget. Adaptation repays its cost only on the large PGM at the loosest target; at tight targets the adapted run needs \emph{more} iterations than zero-init (negative savings), consistent with both arms descending the same objective to the same optimum.}
\label{tab:tta_batchsize}
\begin{tabular}{@{}ll|rrrr@{}}
\toprule
\multicolumn{6}{@{}l@{}}{\emph{(a) Cost of one TTA step}} \\
\midrule
\textbf{PGM} & size & $B{=}0$ & $B{=}5$ & $B{=}10$ & $B{=}20$ \\
 & & (ms) & (ms) & (ms) & (ms) \\
\midrule
grid10$\times$10.f5 & small & 0.29 & 63 {\tiny[29]} & 81 {\tiny[15]} & 95 {\tiny[28]} \\
Seg\_18 & large & 0.35 & 232 {\tiny[49]} & 267 {\tiny[52]} & 276 {\tiny[50]} \\
\bottomrule\end{tabular}
\\[6pt]
\begin{tabular}{@{}ll|rrrrr@{}}
\toprule
\multicolumn{7}{@{}l@{}}{\emph{(b) Is that cost repaid?}} \\
\midrule
\textbf{PGM} & target & iters M1 & iters M3 & saved & budget (ms) & largest $B$ \\
\midrule
grid10$\times$10.f5 & $10^{-1}$ & 6.5 & 2.0 & +4.5 & +20 & $0^{\ast}$ \\
grid10$\times$10.f5 & $10^{-2}$ & 10.0 & 8.0 & +2.0 & +9 & $0^{\ast}$ \\
grid10$\times$10.f5 & $10^{-3}$ & 11.0 & 20.0 & -9.0 & -40 & -- \\
grid10$\times$10.f5 & $10^{-4}$ & 11.0 & 33.0 & -22.0 & -97 & -- \\
grid10$\times$10.f5 & $10^{-6}$ & 11.0 & 56.5 & -45.5 & -201 & -- \\
Seg\_18 & $10^{-1}$ & 54.0 & 30.5 & +23.5 & +413 & 20 \\
Seg\_18 & $10^{-2}$ & 77.0 & 65.0 & +12.0 & +211 & $0^{\ast}$ \\
Seg\_18 & $10^{-3}$ & 86.0 & 80.0 & +6.0 & +106 & $0^{\ast}$ \\
Seg\_18 & $10^{-4}$ & 76.0 & 88.0 & -12.0 & -211 & -- \\
Seg\_18 & $10^{-6}$ & 73.0 & 111.0 & -38.0 & -668 & -- \\
\bottomrule\end{tabular}
\\[2pt]{\scriptsize $^{\ast}$budget is positive but smaller than the cheapest adapting configuration ($B{=}5$), so only the no-adaptation baseline fits; ``--'' marks a negative budget.}
\end{table}

\subsection{Standalone Bound Quality}
\label{app:standalone_pct}

\begin{figure}[ht]
\centering
\includegraphics[width=0.95\textwidth]{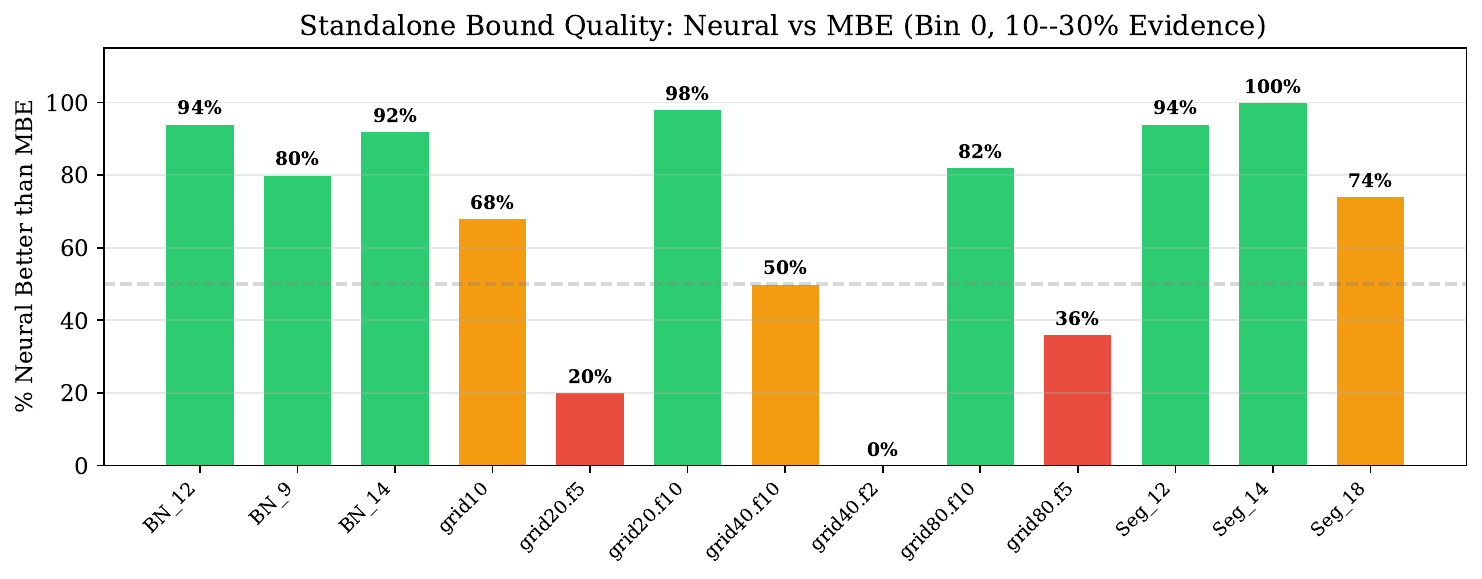}
\caption{\% neural-better than MBE across 13 PGMs at bin~0. Green: $\geq$80\%, orange: 50--80\%, red: $<$50\%.}
\label{fig:pct_better}
\end{figure}

For completeness, we also measure the fraction of test samples where the neural standalone bound is strictly tighter than the MBE (zero-init) bound. Table~\ref{tab:standalone} reports this on the original 13 PGMs at bin~0 (10--30\% evidence); Figure~\ref{fig:pct_better} visualizes the same data.

\begin{table}[ht]
\caption{Standalone bound quality at bin~0. ``\% Better'' = fraction of test instances where the neural bound is strictly tighter than MBE. All bounds are valid by construction (Proposition~\ref{prop:validity}). ``Fwd Time'' is the median wall-clock of a \emph{single-query} forward pass on CPU (200 timed calls after 20 warm-up), which is the quantity a caller actually pays; it ranges from $0.23$ to $0.70$~ms and grows with the input and edge-message dimensions. Batched GPU throughput is $0.03$--$0.05$~ms per query, roughly an order of magnitude below this single-query latency. This view complements the iterations-to-$\epsilon$ analysis in Table~\ref{tab:iters_eps_v2}.}
\label{tab:standalone}
\centering
\small
\begin{tabular}{@{}lccccc@{}}
\toprule
\textbf{PGM} & \textbf{Vars} & \textbf{$e_{\dim}$} & \textbf{\% Neural Better} & \textbf{Fwd Time} \\
\midrule
BN\_12          & 90  & 2,130   & 94\%  & 0.26ms \\
BN\_9           & 105 & 1,354   & 80\%  & 0.24ms \\
BN\_14          & 68  & 1,812   & 92\%  & 0.25ms \\
\midrule
grid10$\times$10  & 100 & 808    & 68\%  & 0.23ms \\
grid20$\times$20.f5 & 400 & 2,988  & 20\%  & 0.28ms \\
grid20$\times$20.f10 & 400 & 5,002  & 98\%  & 0.32ms \\
grid40$\times$40.f10 & 1,600 & 1,400 & 50\%  & 0.48ms \\
grid40$\times$40.f2 & 1,600 & 1,664  & 0\%   & 0.34ms \\
grid80$\times$80.f10 & 6,400 & 5,726 & 82\%  & 0.70ms \\
grid80$\times$80.f5 & 6,400 & 5,262  & 36\%  & 0.68ms \\
\midrule
Seg\_12         & 154 & 2,594   & 94\%  & 0.37ms \\
Seg\_14         & 230 & 3,046   & 100\% & 0.39ms \\
Seg\_18         & 232 & 3,414   & 74\%  & 0.34ms \\
\bottomrule
\end{tabular}
\end{table}

\subsection{AOBB Root Bound Evaluation}
\label{app:aobb}

As an additional application, we evaluate neural bounds as root-node upper bounds for AND/OR Branch-and-Bound (AOBB) search. A tighter root bound provides a better initial optimality certificate. Table~\ref{tab:aobb_root} shows neural root bounds are tighter than MBE in 55--100\% of queries across 11 PGMs. Full search integration (per-node heuristics throughout AOBB) requires per-subproblem neural bounds, which we leave to future work.

\begin{table}[ht]
\caption{Root bound comparison: Neural vs MBE upper bounds at the AOBB root node.}
\label{tab:aobb_root}
\centering
\small
\begin{tabular}{@{}lccc@{}}
\toprule
\textbf{PGM} & \textbf{\% Neural Tighter} & \textbf{Mean $\Delta$} & \textbf{Neural Time} \\
\midrule
BN\_12 & 100\% & 0.66 & 0.35ms \\
BN\_9 & 85\% & 0.18 & 0.37ms \\
BN\_14 & 100\% & 0.46 & 0.41ms \\
Seg\_12 & 100\% & 2.24 & 0.47ms \\
Seg\_14 & 100\% & 1.62 & 0.48ms \\
Seg\_18 & 70\% & 0.60 & 0.49ms \\
grid20$\times$20.f10 & 100\% & 34.05 & 0.51ms \\
\bottomrule
\end{tabular}
\end{table}

\subsection{Empirical Monotonicity}
\label{app:monotonicity}

Since the neural network predicts edge \emph{messages} $\mathbf{m}$ (not bound values directly), there is no architectural reason to expect monotonicity in observed evidence. We verify this empirically: adding one observed variable at a time while tracking the predicted bound, the standard MLP exhibits a $54\%$ average violation rate across 13 PGMs---indistinguishable from random. Enforcing monotonicity architecturally (e.g., via Smooth Min-Max Networks~\citep{igel2023smooth}) is a promising direction for future work when predicting bound \emph{values} directly rather than messages.

\subsection{Full Benchmark Suite}
\label{app:benchmarks}

Table~\ref{tab:benchmark_suite} lists all 40 PGMs in our benchmark suite with their structural properties.

\begin{table}[ht!]
\caption{Full benchmark suite: 40 PGMs spanning 7 categories. $n$ = variables, $e_{\dim}$ = edge-message dimension, $\lambda_{\dim}$ = cluster-table dimension, $i$-bound = 4 except for the four largest grids and \texttt{Promedas\_51}, whose
per-row settings are listed in App.~\ref{app:table1_full}. $\dagger$ = new PGMs
beyond the original 13. Four graphs appear twice under two UAI names
(\texttt{Grids\_29}/\texttt{grid20$\times$20.f5.wrap},
\texttt{Grids\_30}/\texttt{grid20$\times$20.f10},
\texttt{Grids\_18}/\texttt{grid40$\times$40.f10.wrap},
\texttt{Grids\_25}/\texttt{grid40$\times$40.f2}), so these 40 rows are 36
distinct models.}
\label{tab:benchmark_suite}
\centering
\small
\begin{tabular}{@{}llrrrr@{}}
\toprule
\textbf{PGM} & \textbf{Category} & $n$ & $e_{\dim}$ & $\lambda_{\dim}$ & \textbf{MBE--JGLP gap} \\
\midrule
BN\_9 & BN & 105 & 1,354 & 3,122 & 0.26 \\
BN\_12 & BN & 90 & 2,130 & 4,346 & 0.71 \\
BN\_13$^\dagger$ & BN & 125 & 2,130 & 4,574 & 0.56 \\
BN\_14 & BN & 115 & 1,812 & 3,974 & 0.78 \\
BN\_126$^\dagger$ & BN & 512 & 5,268 & 11,414 & 0.23 \\
\midrule
Seg\_12 & Segmentation & 231 & 3,998 & 8,460 & 0.40 \\
Seg\_14 & Segmentation & 231 & 3,958 & 8,388 & 2.30 \\
Seg\_16$^\dagger$ & Segmentation & 228 & 4,236 & 8,764 & 4.00 \\
Seg\_18 & Segmentation & 235 & 3,902 & 8,606 & 1.50 \\
Seg\_19$^\dagger$ & Segmentation & 228 & 4,132 & 8,712 & 0.90 \\
\midrule
PF\_11$^\dagger$ & ProteinFolding & 400 & 6,366 & 13,798 & 73.4 \\
\midrule
grid10$\times$10.f5 & Grid & 100 & 808 & 1,982 & 0.83 \\
grid20$\times$20.f5 & Grid & 400 & 2,988 & 7,312 & 11.7 \\
grid20$\times$20.f10 & Grid & 400 & 5,002 & 10,958 & 44.4 \\
Grids\_25$^\dagger$ & Grid & 1,600 & 28,422 & 59,368 & 63.8 \\
Grids\_26$^\dagger$ & Grid & 400 & 7,138 & 14,858 & 86.8 \\
Grids\_28$^\dagger$ & Grid & 400 & 7,138 & 14,858 & 132.6 \\
Grids\_29$^\dagger$ & Grid & 400 & 7,138 & 14,858 & 41.4 \\
Grids\_30$^\dagger$ & Grid & 400 & 6,366 & 13,798 & 49.3 \\
Grids\_18$^\dagger$ & Grid & 1,600 & 27,964 & 58,454 & 240.6 \\
Grids\_19$^\dagger$ & Grid & 1,600 & 27,964 & 58,454 & 51.0 \\
Grids\_21$^\dagger$ & Grid & 1,600 & 27,964 & 58,454 & 369.5 \\
Grids\_22$^\dagger$ & Grid & 1,600 & 27,964 & 58,454 & 136.2 \\
Grids\_23$^\dagger$ & Grid & 1,600 & 28,422 & 59,368 & 160.0 \\
Grids\_24$^\dagger$ & Grid & 1,600 & 28,422 & 59,368 & 479.1 \\
Grids\_27$^\dagger$ & Grid & 1,600 & 28,422 & 59,368 & 317.6 \\
grid40$\times$40.f10 & Grid & 1,600 & 12,568 & 30,556 & 64.2 \\
grid40$\times$40.f2 & Grid & 1,600 & 14,996 & 35,382 & 12.5 \\
grid80$\times$80.f10 & Grid & 6,400 & 54,558 & 130,372 & 318.1 \\
grid80$\times$80.f5 & Grid & 6,400 & 50,280 & 122,728 & 108.4 \\
\midrule
Promedas\_40$^\dagger$ & Promedas & 1,203 & 8,234 & 20,952 & 1.08 \\
Promedas\_32$^\dagger$ & Promedas & 1,274 & 11,098 & 26,920 & 4.02 \\
Promedas\_33$^\dagger$ & Promedas & 1,174 & 9,736 & 23,654 & 3.36 \\
Promedas\_44$^\dagger$ & Promedas & 999 & 7,990 & 19,788 & 0.73 \\
Promedas\_51$^\dagger$ & Promedas & 859 & 6,118 & 15,368 & 1.19 \\
Promedas\_38$^\dagger$ & Promedas & 1,086 & 9,224 & 22,378 & 0.49 \\
Promedas\_21$^\dagger$ & Promedas & 1,702 & 14,142 & 34,166 & 3.16 \\
Promedas\_22$^\dagger$ & Promedas & 1,664 & 15,424 & 36,954 & 2.70 \\
Promedas\_19$^\dagger$ & Promedas & 1,876 & 16,394 & 39,560 & 3.17 \\
Promedas\_13$^\dagger$ & Promedas & 2,023 & 19,752 & 46,564 & 1.83 \\
\bottomrule
\end{tabular}
\end{table}

\subsection{Contingency and Timing}
\label{app:contingency}

\begin{figure}[ht]
\centering
\includegraphics[width=\textwidth]{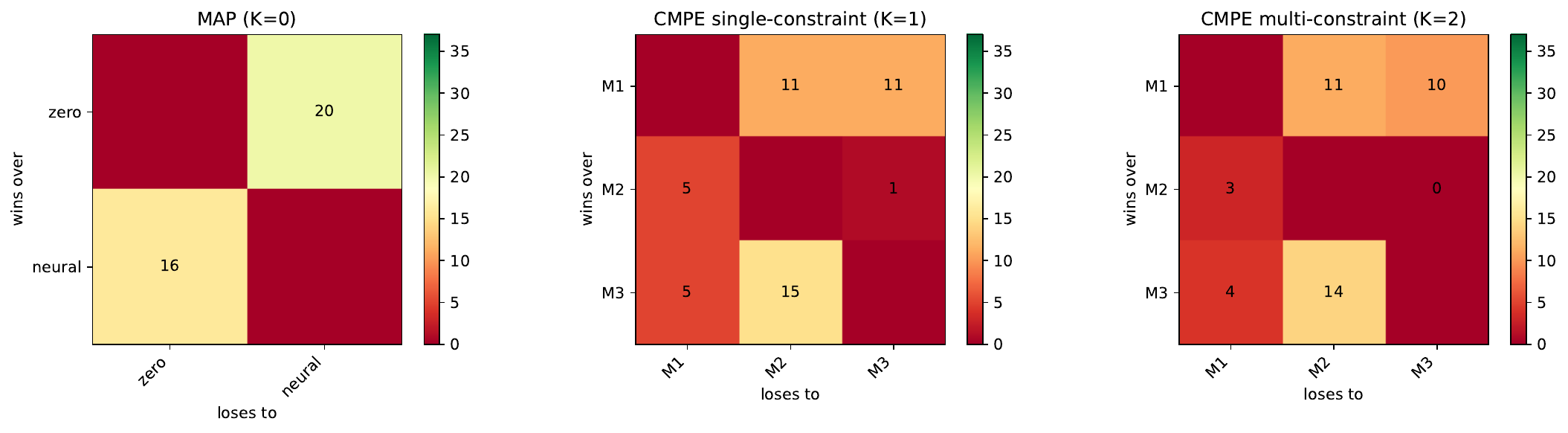}
\caption{Pairwise contingency across 37 MAP / 16 CMPE K=1 / 14 CMPE K=2 PGMs at iter=100. Cell $(i, j)$ is the count of PGMs where method $i$'s log-error is strictly lower (closer to $D_{\text{LP}}$) than method $j$'s. Diagonals are zero by construction.}
\label{fig:contingency}
\end{figure}

Figure~\ref{fig:contingency} counts PGMs where method $i$ achieves strictly lower log-error than method $j$ at iter=100 (ties within $10^{-3}$ excluded). On MAP (37 PGMs), neural-init wins on the majority; on CMPE K=1 (16 PGMs), M3 (NN + replay) wins over M1 (zero-init) and M2 (NN offline) on most PGMs; K=2 shows a similar pattern.

\begin{figure}[ht]
\centering
\includegraphics[width=\textwidth]{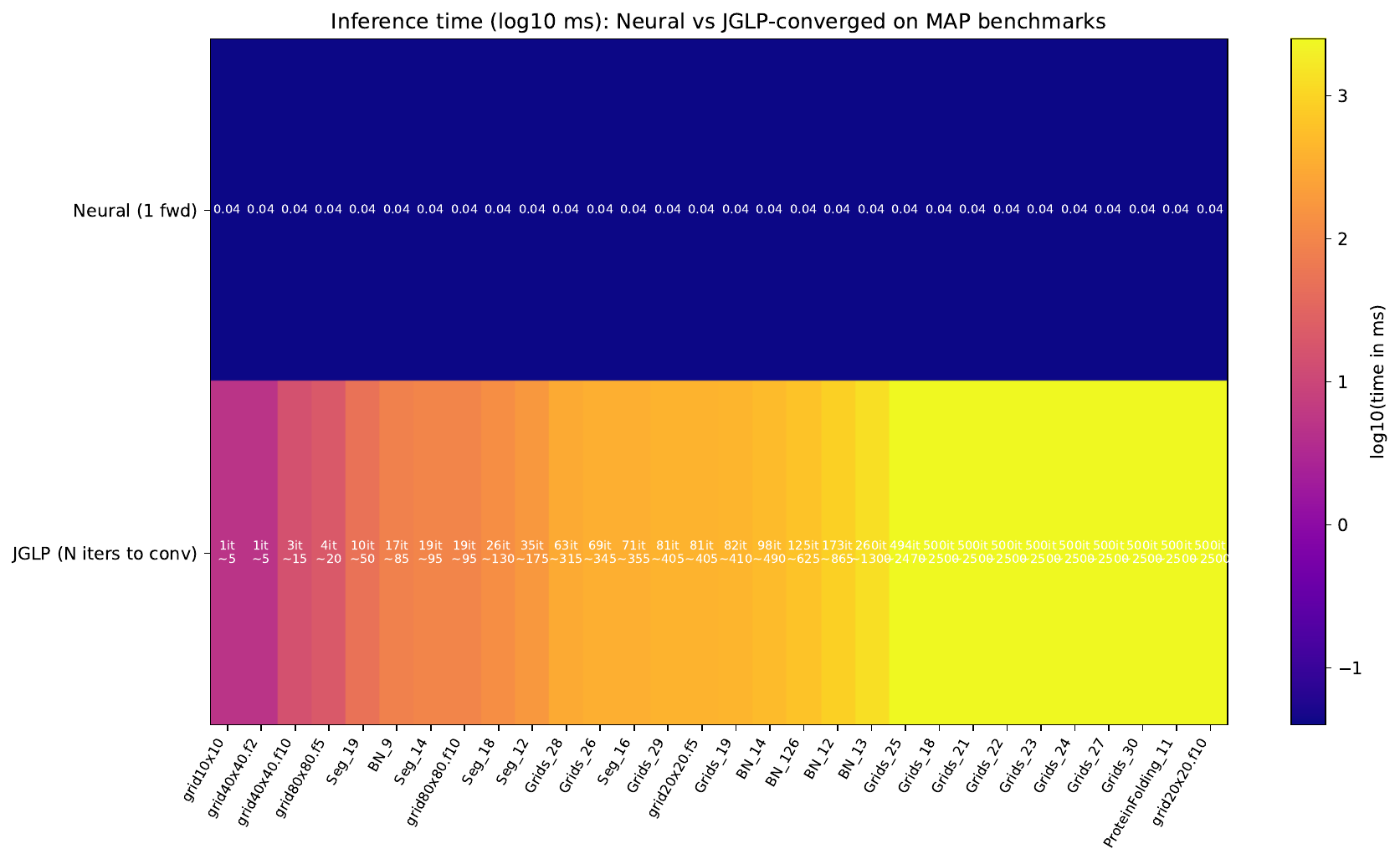}
\caption{Log-scale inference times (ms): a single neural forward pass ($\sim 0.04$ms) vs JGLP iterating to convergence. Time axis is $\log_{10}$(ms); number under JGLP cell shows median iterations to convergence.}
\label{fig:time_heatmap}
\end{figure}

Figure~\ref{fig:time_heatmap} shows the per-PGM inference-time disparity. Neural forward pass is $\sim 0.04$ms independent of PGM size; JGLP time scales with $\lambda_{\dim}$ and convergence-iteration count.

\subsection{Adversarial CMPE on Image Classifiers (FashionMNIST pilot)}
\label{app:mnist_cmpe}

We test whether the framework extends to adversarial CMPE on image classifiers. Following~\cite{rahman2021novel}, we binarize FashionMNIST at threshold $0.5$, crop to a centered $12 \times 12$ patch (144 binary variables), and train a linear SVM (one-vs-rest for class 0) to 92\% test accuracy. The factor graph $f$ encodes the (negated) SVM score as 144 unary factors and a 4-neighbor pairwise smoothness prior; the constraint graph $g$ encodes Hamming distance from the original image.

On this PGM: join-graph $\lambda_{\dim} = 4{,}750$, edge dim $= 2{,}118$, induced width $> 4$. At $i$-bound $4$, the MBE bound (iter 0) equals the LP optimum $D_{\text{LP}} = 79.2$ to within $1.6 \times 10^{-8}$, so JGLP has no room to tighten. This is the same degenerate pattern seen on grid10$\times$10 and grid40$\times$40.f2: when the graph-theoretic MBE at chosen $i$-bound is already at LP, learned warm-starts cannot help. Richer factorizations (deeper classifier, structured priors) remain an avenue for future work; the framework \emph{applies} to MNIST-class problems in that the LP solver and JGLP both run cleanly on the 4,750-dim dual, but demonstrating a genuine speedup requires a non-degenerate instance.

\subsection{Comprehensive Ablation Suite}
\label{app:ablations}

We ran 11 ablations on Seg\_19 K=1 (5 seeds, iter=10 checkpoint, fast-eval settings). Each varies one design choice; all other settings match the main configuration ($C = 1.0$ pure MSE, on-the-fly LP target, $\mu \sim U(0,1)$, hidden = (512, 512, 256), replay = 20, $n_{\text{grad}} = 100$). Lower M3 log-error and higher win@10 are better.

\begin{table}[ht]
\caption{Ablation suite on Seg\_19 K=1 ($\log_{10}|D - D_{\text{LP}}|$ at iter=10). Main config gets win@10 = 72\%. Bold = better than main (lower log-error or higher win\%).}
\label{tab:ablation_suite}
\centering
\small
\begin{tabular}{@{}p{0.45\linewidth}|cccc@{}}
\toprule
\textbf{Variant} & M1 (zero) & M2 (NN off) & M3 (NN+adapt) & win@10 \\
\midrule
\multicolumn{5}{@{}l}{\emph{Loss weight (main is C=1.0 pure MSE):}} \\
loss C=0.0 (pure $D_\tau$) & -0.60 & \textbf{-0.73} & \textbf{-0.93} & \textbf{81\%} \\
loss C=0.5 (combined)      & -0.60 & \textbf{-0.75} & \textbf{-0.91} & \textbf{80\%} \\
\midrule
\multicolumn{5}{@{}l}{\emph{Gradient steps on $\mu$ (main is 100):}} \\
$n_{\text{grad}}$ = 30  & -0.52 & -0.20 & -0.58 & 61\% \\
$n_{\text{grad}}$ = 200 & -0.62 & -0.21 & \textbf{-1.49} & \textbf{86\%} \\
\midrule
\multicolumn{5}{@{}l}{\emph{Replay buffer size (main is 20):}} \\
replay = 5  & -0.60 & -0.16 & -0.63 & 68\% \\
replay = 50 & -0.60 & -0.16 & \textbf{-0.77} & 72\% \\
\midrule
\multicolumn{5}{@{}l}{\emph{Architecture (main is 512/512/256):}} \\
arch = 256/256/128  & -0.60 & -0.24 & \textbf{-0.78} & 70\% \\
arch = 768/768/512  & -0.60 & -0.30 & -0.70 & 71\% \\
\midrule
\multicolumn{5}{@{}l}{\emph{Other (main is on-the-fly + LP target + $\mu$ in [0,1])}} \\
fixed dataset (8000 samples) & -0.60 & -0.23 & -0.69 & 72\% \\
JGLP-converged target & -0.60 & -0.31 & -0.73 & 70\% \\
$\mu \sim U(0, 3)$ training & -0.60 & -0.16 & \textbf{-0.99} & \textbf{75\%} \\
\midrule
\textbf{main (default)} & -0.60 & -0.16 & -0.73 & 72\% \\
\bottomrule
\end{tabular}
\end{table}

Key findings:
\begin{itemize}
\item \textbf{Gradient steps matter most} (61--86\% across $n_{\text{grad}} \in \{30, 100, 200\}$). $n_{\text{grad}} = 100$ is reasonable but 200 gives a substantial $+14$pp improvement; we recommend tuning this per application.
\item \textbf{Loss weight: pure $D_\tau$ (C=0) marginally beats pure MSE on this PGM} (81\% vs 72\%), opposite of the recommendation that motivated our main configuration. This is one PGM; the broader story (the CMPE tables, App.~\ref{app:meanstd}) used pure MSE, which performed strongly across 16/14 PGMs. The combined loss (C=0.5) is comparable.
\item \textbf{LP target vs JGLP target}: 72\% vs 70\% — within noise. Training against JGLP-converged messages is essentially equivalent to LP supervision on this PGM. The LP solver remains valuable as an evaluation reference (Tables~\ref{tab:cmpe_eps_v2}--\ref{tab:cmpe_eps_two_v2}).
\item \textbf{On-the-fly vs fixed dataset}: 72\% vs 72\% — no significant difference, so the on-the-fly cost can be replaced by a one-shot dataset generation if compute is constrained. Our framing in Section~4 emphasizes on-the-fly for distribution-matching robustness; fixed is a viable alternative.
\item \textbf{Architecture (256/256/128 vs 768/768/512)}: 70--71\% — depth/width within ±2pp. The default (512/512/256) is well-chosen.
\item \textbf{Replay buffer size}: 5 hurts ($-4$pp); 20 (default) and 50 are equivalent; 20 is the right balance between memory and signal.
\item \textbf{$\mu$ range U(0,3) vs U(0,1)}: $+3$pp improvement, suggesting wider $\mu$ training gives the NN more coverage of the gradient-descent trajectory.
\end{itemize}

\subsection{Evidence Encoding: Scalar vs One-Hot}
\label{app:encoding}

The input encoding $\mathbf{e}_i = v_i / (d_i - 1)$ for observed variables 
maps each observed value into $[0, 1]$ via scalar division, with $-1$ as a sentinel for unobserved variables. A reasonable alternative is one-hot encoding: a $(\sum_i d_i)$-dimensional sparse vector with a $1$ at position $(i, v_i)$ for each observed $(i, v_i)$ pair and zeros elsewhere, plus a separate binary mask channel marking observed variables. We chose scalar normalization for three reasons.

\textbf{Input dimensionality.} For our benchmarks, scalar encoding gives input dimension $n$ (one entry per variable), while one-hot gives $\sum_i d_i$. On Promedas networks ($n \approx 1{,}500$--$2{,}000$ variables, $d_i \in \{2, 3\}$ typical), one-hot expands the input to $\sim 3{,}000$--$6{,}000$ dimensions, $2$--$3\times$ larger. On grid models with binary variables ($d_i = 2$), the two encodings produce identically-sized inputs, but on multi-valued domains the gap widens. Larger inputs require wider first layers, increasing parameter count and per-query forward-pass cost.

\textbf{Empirical performance is already strong.} Table~\ref{tab:standalone} shows scalar encoding achieves 94--100\% neural-better on BN and Segmentation PGMs at bin~0; the encoding is not the empirical bottleneck. The dominant variation across PGMs (Table~\ref{tab:ablation_suite}) comes from gradient-step count, replay-buffer size, and $\mu$ training range, not from input representation.

\textbf{Limitation: ordinal bias on categorical variables.} Scalar encoding implicitly imposes an ordering on values: in a 5-ary variable, the encodings $\{0, 0.25, 0.5, 0.75, 1.0\}$ suggest value $1$ is ``between'' values $0$ and $2$. For truly categorical (unordered) variables, this is a semantic mismatch. Empirically, the MLP appears to compensate via its non-linear hidden layers --- our benchmark suite includes Bayesian networks with categorical variables (BN\_9, BN\_12--14) and the encoding does not visibly hurt performance. For PGMs where categorical structure dominates (e.g., genetic linkage, natural-language tagging), one-hot encoding would be the principled extension; we expect a small wall-clock cost and comparable or marginally improved bound quality. We leave a systematic ablation comparing scalar vs.\ one-hot encoding to future work.

\subsection{Seed Sensitivity}
\label{app:seed_sensitivity}

Recomputing the M3 win\% on Seg\_19 K=1 using subsets of size $n \in \{1, 2, 3, 5\}$ of the 5 available seeds yields 76.0, 75.0, 74.7, 72.0\%. The 5-seed summary is stable to within 4 percentage points against subsetting, confirming that the 5-seed median in Tables~\ref{tab:cmpe_eps_v2}--\ref{tab:cmpe_eps_two_v2} is not dominated by a single-seed outlier.

\subsection{Full Standalone Bound Results (v2 metric)}
\label{app:standalone_v2}

The compact main-body Table~\ref{tab:iters_eps_v2} shows only the top 15 PGMs. The full 37-PGM v2 table follows; same metric and win convention.

\begin{table*}[t]\centering\small
\caption{Standalone bound quality $(D - D^{\mathrm{LP}})$ at fixed iterations. Method N uses a neural warm-start; J uses zero-init. Win\% = fraction of test seeds where N\,$\le$\,J$+\varepsilon$ ($\varepsilon=10^{-7}$). $\star$ marks 0\% cells where both methods are within numerical tolerance of $D^{\mathrm{LP}}$.}
\label{tab:standalone_v2}
\resizebox{\textwidth}{!}{%
\begin{tabular}{lrrrrrrrrrrrrrrr}\toprule
PGM & J (iter=0~$(\times 10^{-1})$) & N & win\% & J (iter=10~$(\times 10^{-2})$) & N & win\% & J (iter=50~$(\times 10^{-4})$) & N & win\% & J (iter=100~$(\times 10^{-6})$) & N & win\% & J (iter=500~$(\times 10^{-7})$) & N & win\% \\
\midrule
BN\_12 & 9.66 & 6.79 & \textbf{100} & 60.9 & 32.9 & \textbf{100} & 1820 & 964 & 90 & 50375 & 20465 & \textbf{100} & 1.80 & 1.90 & \textbf{100} \\
BN\_126 & 2.94 & 5.29 & 0 & 9.33 & 8.69 & 89 & 83.9 & 75.1 & 56 & 2.88 & 2.92 & 67 & 4.96 & 6.05 & 67 \\
BN\_13 & 9.08 & 6.54 & \textbf{100} & 50.8 & 25.2 & \textbf{100} & 1078 & 733 & \textbf{100} & 45571 & 33649 & 60 & 0.70 & 1.05 & 80 \\
BN\_14 & 4.63 & 4.12 & 90 & 22.0 & 15.0 & \textbf{100} & 592 & 478 & 80 & 1241 & 33.0 & 80 & 3.27 & 3.87 & 90 \\
BN\_9 & 2.58 & 2.50 & 30 & 12.2 & 2.93 & 70 & 34.8 & 0.02 & 70 & 0.21 & 0.12 & \textbf{100} & 2.08 & 1.05 & \textbf{100} \\
Grids\_18 & 696 & 685 & 90 & 1382 & 1350 & \textbf{100} & 22276 & 21575 & 90 & 1121910 & 1094359 & 80 & 92624 & 98491 & 60 \\
Grids\_19 & 104 & 102 & 80 & 171 & 136 & \textbf{100} & 142 & 146 & 60 & 0.68 & 47.3 & 20 & 3.99 & 5.97 & 10 \\
Grids\_21 & 1044 & 1008 & \textbf{100} & 2041 & 1989 & \textbf{100} & 32011 & 31541 & 80 & 1720365 & 1699160 & 70 & 184739 & 181540 & 70 \\
Grids\_22 & 304 & 307 & 40 & 723 & 707 & 90 & 7345 & 6861 & 50 & 130873 & 110242 & 50 & 2140 & 1555 & 40 \\
Grids\_23 & 322 & 329 & 70 & 1464 & 1376 & \textbf{100} & 23271 & 23909 & 80 & 870712 & 918743 & 60 & 87875 & 84131 & 60 \\
Grids\_24 & 977 & 976 & 70 & 4768 & 4684 & \textbf{100} & 100092 & 99704 & 80 & 3611215 & 3621534 & 40 & 1292926 & 1211802 & 40 \\
Grids\_25 & 143 & 126 & \textbf{100} & 496 & 340 & \textbf{100} & 5512 & 3485 & \textbf{100} & 104972 & 67560 & 80 & 166 & 152 & 20 \\
Grids\_26 & 172 & 152 & 60 & 241 & 244 & 60 & 782 & 784 & 40 & 0.00 & 0.58 & 10 & 0.04 & 1.83 & 10 \\
Grids\_27 & 653 & 652 & 60 & 3147 & 3073 & 90 & 67563 & 65280 & 80 & 2224740 & 2130994 & 60 & 1237842 & 1089313 & 50 \\
Grids\_28 & 231 & 221 & 50 & 529 & 542 & 50 & 1279 & 1304 & 30 & 0.00 & 2.50 & 0 & 0.03 & 1.76 & 30 \\
Grids\_29 & 78.5 & 73.8 & 30 & 118 & 107 & 80 & 1041 & 1040 & 50 & 0.00 & 2.63 & 30 & 0.01 & 1.67 & 30 \\
Grids\_30 & 512 & 318 & \textbf{100} & 1488 & 1277 & \textbf{100} & 45258 & 41701 & 80 & 2653707 & 2427422 & 70 & 4052408 & 4193701 & 60 \\
Promedas\_21 & -0.00 & 25.2 & 0 & -0.00 & 7.43 & 30 & -0.00 & 36.5 & 30 & -0.14 & 398 & 30 & -1.36 & 5.23 & 30 \\
Promedas\_32 & 35.1 & 35.0 & 30 & 351 & 27.8 & 60 & 35066 & 29.4 & 60 & 3506558 & 649 & 60 & 35065577 & 11.2 & 60 \\
Promedas\_33 & 25.0 & 20.4 & 60 & 250 & 21.4 & 60 & 25048 & 335 & 60 & 2504801 & 14214 & 60 & 25048011 & 890 & 60 \\
Promedas\_38 & -0.00 & 4.99 & 0 & -0.00 & 5.52 & 0 & -0.00 & 96.2 & 0 & -0.04 & 2895 & 0 & -0.38 & 28.0 & 0 \\
Promedas\_40 & -0.00 & 5.60 & 40 & -0.00 & 2.03 & 40 & -0.00 & 10.9 & 0 & -0.02 & 137 & 0 & -0.24 & 4.27 & 0 \\
Promedas\_44 & 3.64 & 6.39 & 10 & 2.85 & 2.56 & 50 & 0.00 & 22.8 & 0 & -0.02 & 582 & 0 & -0.18 & 5.61 & 0 \\
Promedas\_51 & 11.9 & 6.01 & 80 & 119 & 2.31 & 80 & 11923 & 15.7 & 80 & 1192349 & 196 & 80 & 11923488 & 6.47 & 80 \\
ProteinFolding\_11 & 793 & 507 & \textbf{100} & 2189 & 1938 & \textbf{100} & 71543 & 68609 & 90 & 3597120 & 3296646 & 90 & 4348965 & 4429694 & 50 \\
Segmentation\_12 & 20.0 & 12.6 & \textbf{100} & 98.7 & 37.0 & \textbf{100} & 2.83 & 0.12 & 20 & 0.08 & 0.20 & 50 & 0.80 & 2.04 & 50 \\
Segmentation\_14 & 15.3 & 3.78 & \textbf{100} & 18.7 & 1.03 & \textbf{100} & 0.00 & 0.26 & 0 & 0.48 & 0.82 & 0 & 4.76 & 6.67 & 0 \\
Segmentation\_16 & 38.5 & 13.5 & \textbf{100} & 194 & 32.2 & \textbf{100} & 2272 & 1.34 & 70 & 0.91 & 1.98 & 30 & 5.88 & 8.01 & 20 \\
Segmentation\_18 & 6.94 & 9.43 & 40 & 29.8 & 19.1 & 80 & 138 & 2.44 & 40 & 0.99 & 0.73 & 10 & 9.87 & 7.35 & 10 \\
Segmentation\_19 & 3.93 & 5.74 & 22 & 0.04 & 6.91 & 33 & 0.03 & 1.08 & 11 & 1.96 & 2.37 & 11 & 5.66 & 2.51 & 22 \\
grid10x10.f5 & 0.00 & 1.48 & 0 & 0.00 & 0.09 & 0 & 0.00 & 0.00 & $67$\textsuperscript{$\star$} & 0.00 & 0.04 & $100$\textsuperscript{$\star$} & 0.00 & 0.45 & $100$\textsuperscript{$\star$} \\
grid20x20.f10 & 512 & 322 & \textbf{100} & 1488 & 1279 & \textbf{100} & 45258 & 41900 & 80 & 2653707 & 2427322 & 70 & 4052408 & 4178662 & 60 \\
grid20x20.f5 & 78.5 & 73.6 & 30 & 118 & 107 & 90 & 1041 & 1038 & 50 & 0.00 & 1.02 & 20 & 0.01 & 1.52 & 30 \\
grid40x40.f10 & 0.20 & 3.44 & 0 & 0.00 & 0.04 & 0 & 0.00 & 0.00 & $80$\textsuperscript{$\star$} & 0.00 & 0.04 & $100$\textsuperscript{$\star$} & 0.01 & 0.39 & $100$\textsuperscript{$\star$} \\
grid40x40.f2 & 0.00 & 3.59 & 0 & 0.00 & 0.05 & 0 & 0.00 & 0.00 & $60$\textsuperscript{$\star$} & 0.00 & 0.04 & $100$\textsuperscript{$\star$} & 0.00 & 0.44 & $100$\textsuperscript{$\star$} \\
grid80x80.f10 & 45.3 & 53.3 & 0 & 31.2 & 23.2 & 10 & 0.21 & 0.26 & 20 & 0.03 & 0.10 & 60 & 0.25 & 0.97 & $80$\textsuperscript{$\star$} \\
grid80x80.f5 & 4.06 & 11.8 & 0 & 0.00 & 0.11 & 0 & 0.00 & 0.10 & 0 & 0.01 & 0.54 & 0 & 0.08 & 1.69 & 0 \\
\bottomrule\end{tabular}}\end{table*}

\subsection{Full mean \texorpdfstring{$\pm$}{\textpm} std tables (companion to the main body)}
\label{app:meanstd}

The main body reports a subset of iteration checkpoints for MAP and, for CMPE, the two-constraint setting only. The tables below give the same quantities over all five checkpoints $\{0,10,50,100,500\}$ and for both $K{=}1$ and $K{=}2$, in the same mean $\pm$ std format, together with the $1{,}000$-seed stability check. All of them report the same underlying trials as the main-body tables; Table~\ref{tab:t1_meanstd_100} is the $100$-seed companion to Table~\ref{tab:map}.

\begin{table*}[t]\centering
\caption{\textbf{Standalone bound quality (mean $\pm$ std)}: dual gap $|D - D^{\mathrm{LP}}|$ at fixed JGLP iterations across 37 PGMs. J = zero-init, N = neural-init. Mean and std (population, ddof=0) computed over 10 bin-0 test seeds per PGM. \textbf{Bold} = neural is tie-or-better than J (winning, or within $\max(\sigma_N, \sigma_J)$).}
\label{tab:t1_meanstd}
\resizebox{\textwidth}{!}{%
\begin{tabular}{@{}l|rr|rr|rr|rr@{}}
\toprule
  & \multicolumn{2}{c|}{iter=10} & \multicolumn{2}{c|}{iter=50} & \multicolumn{2}{c|}{iter=100} & \multicolumn{2}{c|}{iter=500} \\
\textbf{PGM} & J & N & J & N & J & N & J & N \\
\midrule
BN\_12 & 0.638~$\pm$~0.238 & \textbf{0.373~$\pm$~0.179} & 0.184~$\pm$~0.147 & \textbf{0.116~$\pm$~0.105} & 0.0878~$\pm$~0.0971 & \textbf{0.0595~$\pm$~0.0723} & 0.00828~$\pm$~0.0170 & \textbf{0.00699~$\pm$~0.0145} \\
BN\_126 & 0.248~$\pm$~0.342 & \textbf{0.239~$\pm$~0.338} & 0.0647~$\pm$~0.147 & \textbf{0.0639~$\pm$~0.146} & 0.0150~$\pm$~0.0389 & \textbf{0.0149~$\pm$~0.0388} & 1.01e-06~$\pm$~1.16e-06 & \textbf{1.08e-06~$\pm$~1.14e-06} \\
BN\_13 & 0.474~$\pm$~0.115 & \textbf{0.255~$\pm$~0.0878} & 0.116~$\pm$~0.0731 & \textbf{0.0884~$\pm$~0.0715} & 0.0584~$\pm$~0.0525 & \textbf{0.0503~$\pm$~0.0538} & 0.00197~$\pm$~0.00502 & \textbf{0.00263~$\pm$~0.00556} \\
BN\_14 & 0.278~$\pm$~0.173 & \textbf{0.174~$\pm$~0.128} & 0.0937~$\pm$~0.137 & \textbf{0.0420~$\pm$~0.0368} & 0.0546~$\pm$~0.143 & \textbf{0.00575~$\pm$~0.00800} & 0.0483~$\pm$~0.145 & \textbf{8.25e-06~$\pm$~1.59e-05} \\
BN\_9 & 0.162~$\pm$~0.156 & \textbf{0.0607~$\pm$~0.0701} & 0.126~$\pm$~0.169 & \textbf{0.00680~$\pm$~0.0188} & 0.125~$\pm$~0.170 & \textbf{0.00147~$\pm$~0.00442} & 0.125~$\pm$~0.170 & \textbf{1.35e-07~$\pm$~6.90e-08} \\
Grids\_18 & 14.3~$\pm$~5.05 & \textbf{13.7~$\pm$~4.89} & 2.01~$\pm$~0.870 & \textbf{1.96~$\pm$~0.864} & 1.07~$\pm$~0.561 & \textbf{1.06~$\pm$~0.553} & 0.0228~$\pm$~0.0303 & \textbf{0.0228~$\pm$~0.0303} \\
Grids\_19 & 1.68~$\pm$~0.984 & \textbf{1.38~$\pm$~0.814} & 0.0356~$\pm$~0.0413 & \textbf{0.0276~$\pm$~0.0291} & 0.00756~$\pm$~0.0148 & \textbf{0.00627~$\pm$~0.0116} & 1.60e-06~$\pm$~2.42e-06 & \textbf{1.48e-06~$\pm$~1.62e-06} \\
Grids\_21 & 20.8~$\pm$~6.78 & \textbf{20.0~$\pm$~6.67} & 2.88~$\pm$~1.37 & \textbf{2.83~$\pm$~1.33} & 1.60~$\pm$~0.823 & \textbf{1.59~$\pm$~0.818} & 0.0603~$\pm$~0.117 & \textbf{0.0599~$\pm$~0.117} \\
Grids\_22 & 6.92~$\pm$~1.21 & \textbf{6.69~$\pm$~1.24} & 0.810~$\pm$~0.564 & \textbf{0.810~$\pm$~0.572} & 0.284~$\pm$~0.330 & \textbf{0.282~$\pm$~0.347} & 0.0147~$\pm$~0.0299 & \textbf{0.0145~$\pm$~0.0269} \\
Grids\_23 & 14.6~$\pm$~2.33 & \textbf{13.7~$\pm$~2.22} & 2.56~$\pm$~0.917 & \textbf{2.48~$\pm$~0.943} & 0.904~$\pm$~0.571 & \textbf{0.901~$\pm$~0.581} & 0.0616~$\pm$~0.120 & \textbf{0.0568~$\pm$~0.114} \\
Grids\_24 & 47.7~$\pm$~6.89 & \textbf{46.8~$\pm$~6.79} & 9.80~$\pm$~3.14 & \textbf{9.76~$\pm$~3.16} & 4.11~$\pm$~1.56 & \textbf{4.12~$\pm$~1.60} & 0.183~$\pm$~0.164 & \textbf{0.181~$\pm$~0.162} \\
Grids\_25 & 4.94~$\pm$~0.820 & \textbf{3.35~$\pm$~0.604} & 0.544~$\pm$~0.301 & \textbf{0.397~$\pm$~0.283} & 0.105~$\pm$~0.0682 & \textbf{0.0752~$\pm$~0.0596} & 7.14e-05~$\pm$~0.000161 & \textbf{7.14e-05~$\pm$~0.000162} \\
Grids\_26 & 3.13~$\pm$~2.19 & \textbf{3.12~$\pm$~2.17} & 0.138~$\pm$~0.147 & \textbf{0.138~$\pm$~0.147} & 2.50e-08~$\pm$~4.90e-08 & \textbf{8.62e-07~$\pm$~9.57e-07} & 2.51e-08~$\pm$~4.86e-08 & 1.83e-07~$\pm$~6.33e-08 \\
Grids\_27 & 31.4~$\pm$~4.33 & \textbf{30.6~$\pm$~4.22} & 6.42~$\pm$~1.84 & \textbf{6.34~$\pm$~1.84} & 2.51~$\pm$~1.06 & \textbf{2.49~$\pm$~1.08} & 0.133~$\pm$~0.134 & \textbf{0.131~$\pm$~0.131} \\
Grids\_28 & 5.21~$\pm$~3.22 & \textbf{5.22~$\pm$~3.19} & 0.279~$\pm$~0.383 & \textbf{0.279~$\pm$~0.380} & 9.18e-09~$\pm$~1.30e-08 & 2.65e-06~$\pm$~1.92e-06 & 9.20e-09~$\pm$~1.32e-08 & 1.90e-07~$\pm$~1.35e-07 \\
Grids\_29 & 1.33~$\pm$~0.849 & \textbf{1.30~$\pm$~0.851} & 0.117~$\pm$~0.115 & \textbf{0.115~$\pm$~0.114} & 0.00990~$\pm$~0.0226 & \textbf{0.00923~$\pm$~0.0207} & 2.77e-08~$\pm$~5.44e-08 & 1.56e-07~$\pm$~7.86e-08 \\
Grids\_30 & 14.8~$\pm$~3.16 & \textbf{12.9~$\pm$~2.76} & 5.11~$\pm$~2.08 & \textbf{4.78~$\pm$~1.98} & 2.57~$\pm$~1.55 & \textbf{2.49~$\pm$~1.51} & 0.576~$\pm$~0.588 & \textbf{0.569~$\pm$~0.594} \\
Promedas\_21 & 0.948~$\pm$~1.45 & \textbf{0.266~$\pm$~0.296} & 0.948~$\pm$~1.45 & \textbf{0.00372~$\pm$~0.000281} & 0.948~$\pm$~1.45 & \textbf{0.000406~$\pm$~3.82e-05} & 0.948~$\pm$~1.45 & \textbf{1.45e-06~$\pm$~2.65e-06} \\
Promedas\_32 & 2.18~$\pm$~1.79 & \textbf{0.215~$\pm$~0.129} & 2.18~$\pm$~1.79 & \textbf{0.00316~$\pm$~0.000636} & 2.18~$\pm$~1.79 & \textbf{0.000683~$\pm$~0.000285} & 2.18~$\pm$~1.79 & \textbf{1.82e-06~$\pm$~1.43e-06} \\
Promedas\_33 & 1.85~$\pm$~1.59 & \textbf{0.343~$\pm$~0.276} & 1.85~$\pm$~1.59 & \textbf{0.0326~$\pm$~0.00332} & 1.85~$\pm$~1.59 & \textbf{0.0138~$\pm$~0.00147} & 1.85~$\pm$~1.59 & \textbf{8.92e-05~$\pm$~3.12e-06} \\
Promedas\_38 & 3.79e-08~$\pm$~3.50e-08 & 0.0555~$\pm$~0.00182 & 3.79e-08~$\pm$~3.50e-08 & 0.00962~$\pm$~0.000214 & 3.80e-08~$\pm$~3.50e-08 & 0.00285~$\pm$~7.78e-05 & 3.79e-08~$\pm$~3.50e-08 & 2.78e-06~$\pm$~1.19e-07 \\
Promedas\_40 & 0.0391~$\pm$~0.0479 & \textbf{0.0204~$\pm$~0.000670} & 5.64e-07~$\pm$~8.04e-07 & 0.00104~$\pm$~0.000120 & 5.64e-07~$\pm$~8.04e-07 & 0.000133~$\pm$~2.02e-05 & 5.64e-07~$\pm$~8.04e-07 & \textbf{6.50e-07~$\pm$~4.99e-07} \\
Promedas\_44 & 0.0285~$\pm$~0.0285 & \textbf{0.0272~$\pm$~0.00309} & 6.43e-07~$\pm$~1.00e-06 & 0.00239~$\pm$~0.000430 & 3.77e-07~$\pm$~1.07e-06 & 0.000586~$\pm$~0.000107 & 3.77e-07~$\pm$~1.07e-06 & \textbf{8.07e-07~$\pm$~7.50e-07} \\
Promedas\_51 & 0.954~$\pm$~0.477 & \textbf{0.0212~$\pm$~0.00364} & 0.954~$\pm$~0.477 & \textbf{0.00141~$\pm$~0.000552} & 0.954~$\pm$~0.477 & \textbf{0.000204~$\pm$~0.000106} & 0.954~$\pm$~0.477 & \textbf{6.49e-07~$\pm$~3.41e-08} \\
PF\_11 & 22.2~$\pm$~4.05 & \textbf{19.4~$\pm$~3.53} & 7.54~$\pm$~2.48 & \textbf{7.10~$\pm$~2.35} & 3.84~$\pm$~1.99 & \textbf{3.66~$\pm$~1.96} & 0.888~$\pm$~0.971 & \textbf{0.883~$\pm$~0.971} \\
Seg\_12 & 1.19~$\pm$~0.900 & \textbf{0.701~$\pm$~0.761} & 0.177~$\pm$~0.321 & \textbf{0.0187~$\pm$~0.0353} & 0.163~$\pm$~0.327 & \textbf{1.96e-07~$\pm$~1.11e-07} & 0.163~$\pm$~0.327 & \textbf{1.96e-07~$\pm$~1.11e-07} \\
Seg\_14 & 0.189~$\pm$~0.0474 & \textbf{0.0176~$\pm$~0.0163} & 2.20e-06~$\pm$~3.00e-06 & \textbf{0.000147~$\pm$~0.000247} & 2.20e-06~$\pm$~3.03e-06 & \textbf{6.06e-06~$\pm$~8.42e-06} & 2.20e-06~$\pm$~3.03e-06 & \textbf{2.39e-06~$\pm$~3.05e-06} \\
Seg\_16 & 1.93~$\pm$~0.396 & \textbf{0.417~$\pm$~0.273} & 0.236~$\pm$~0.202 & \textbf{0.0286~$\pm$~0.0749} & 0.0280~$\pm$~0.0567 & \textbf{7.66e-06~$\pm$~1.50e-05} & 1.24e-06~$\pm$~1.59e-06 & \textbf{1.47e-06~$\pm$~1.58e-06} \\
Seg\_18 & 0.627~$\pm$~0.692 & \textbf{0.419~$\pm$~0.478} & 0.0731~$\pm$~0.102 & \textbf{0.0504~$\pm$~0.0773} & 0.00506~$\pm$~0.0152 & \textbf{4.45e-06~$\pm$~9.49e-06} & 0.00506~$\pm$~0.0152 & \textbf{1.94e-06~$\pm$~2.69e-06} \\
Seg\_19 & 0.125~$\pm$~0.150 & \textbf{0.104~$\pm$~0.114} & 0.0443~$\pm$~0.123 & \textbf{0.00682~$\pm$~0.0123} & 0.0437~$\pm$~0.123 & \textbf{5.24e-06~$\pm$~6.52e-06} & 0.0437~$\pm$~0.123 & \textbf{1.09e-06~$\pm$~1.14e-06} \\
grid10$\times$10.f5 & 9.12e-09~$\pm$~1.17e-08 & \textbf{0.00197~$\pm$~0.00206} & 9.12e-09~$\pm$~1.17e-08 & \textbf{1.22e-05~$\pm$~1.73e-05} & 9.12e-09~$\pm$~1.20e-08 & 4.60e-08~$\pm$~2.60e-08 & 9.10e-09~$\pm$~1.17e-08 & 4.61e-08~$\pm$~2.61e-08 \\
grid20$\times$20.f10 & 14.8~$\pm$~3.16 & \textbf{12.9~$\pm$~2.76} & 5.11~$\pm$~2.08 & \textbf{4.79~$\pm$~1.98} & 2.57~$\pm$~1.55 & \textbf{2.49~$\pm$~1.51} & 0.576~$\pm$~0.588 & \textbf{0.570~$\pm$~0.594} \\
grid20$\times$20.f5 & 1.33~$\pm$~0.849 & \textbf{1.30~$\pm$~0.850} & 0.117~$\pm$~0.115 & \textbf{0.115~$\pm$~0.114} & 0.00990~$\pm$~0.0226 & \textbf{0.00922~$\pm$~0.0207} & 2.77e-08~$\pm$~5.44e-08 & 1.52e-07~$\pm$~5.38e-08 \\
grid40$\times$40.f10 & 3.19e-09~$\pm$~3.72e-09 & 0.000370~$\pm$~0.000170 & 3.19e-09~$\pm$~3.72e-09 & \textbf{6.82e-07~$\pm$~1.30e-06} & 3.19e-09~$\pm$~3.72e-09 & 4.60e-08~$\pm$~1.90e-08 & 3.20e-09~$\pm$~3.70e-09 & 4.62e-08~$\pm$~1.88e-08 \\
grid40$\times$40.f2 & 3.44e-09~$\pm$~4.70e-09 & 0.000500~$\pm$~0.000300 & 3.44e-09~$\pm$~4.70e-09 & \textbf{3.99e-07~$\pm$~6.75e-07} & 3.44e-09~$\pm$~4.70e-09 & 4.80e-08~$\pm$~1.30e-08 & 3.40e-09~$\pm$~4.70e-09 & 4.83e-08~$\pm$~1.33e-08 \\
grid80$\times$80.f10 & 0.469~$\pm$~0.451 & \textbf{0.344~$\pm$~0.310} & 0.159~$\pm$~0.435 & \textbf{0.0130~$\pm$~0.0282} & 0.147~$\pm$~0.438 & \textbf{0.000721~$\pm$~0.00184} & 0.146~$\pm$~0.438 & \textbf{2.05e-07~$\pm$~2.80e-07} \\
grid80$\times$80.f5 & 0.0255~$\pm$~0.0561 & \textbf{0.0262~$\pm$~0.0560} & 2.00e-06~$\pm$~4.30e-06 & 1.76e-05~$\pm$~1.52e-05 & 5.33e-07~$\pm$~1.09e-06 & \textbf{1.34e-06~$\pm$~1.36e-06} & 5.33e-07~$\pm$~1.09e-06 & \textbf{6.86e-07~$\pm$~1.10e-06} \\
\bottomrule
\end{tabular}%
}
\end{table*}

\begin{table*}[t]\centering\scriptsize
\caption{\textbf{Standalone bound quality at $100$ seeds, all checkpoints} (companion to Table~\ref{tab:map}): dual gap $|D - D^{\mathrm{LP}}|$ (mean $\pm$ std) at every reported JGLP iteration on the $16$ canonical PGM rows. J = zero-init, N = neural-init. \textbf{Bold} marks the neural arm where it ties or beats zero-init within one standard deviation. Mean and population standard deviation over $100$ bin-0 test seeds per PGM (Segmentation\_18 and Segmentation\_19 have $99$: one seed each fails deterministically in the LP solver and is reported rather than dropped). The $37$-PGM suite in Table~\ref{tab:t1_meanstd} remains at $10$ seeds. At iterations $100$ and $500$ both arms are at or near the LP solver's working accuracy on several PGMs, so differences there should not be read as a result in either direction.}
\label{tab:t1_meanstd_100}
\resizebox{\textwidth}{!}{%
\begin{tabular}{@{}l|rr|rr|rr|rr|rr@{}}
\toprule
  & \multicolumn{2}{c|}{iter=0} & \multicolumn{2}{c|}{iter=10} & \multicolumn{2}{c|}{iter=50} & \multicolumn{2}{c|}{iter=100} & \multicolumn{2}{c}{iter=500} \\
\textbf{PGM} & J & N & J & N & J & N & J & N & J & N \\
\midrule
BN\_12 & 0.872~$\pm$~0.289 & \textbf{0.675~$\pm$~0.222} & 0.525~$\pm$~0.248 & \textbf{0.304~$\pm$~0.178} & 0.150~$\pm$~0.131 & \textbf{0.095~$\pm$~0.093} & 0.065~$\pm$~0.078 & \textbf{0.042~$\pm$~0.056} & 3.27e-03~$\pm$~9.33e-03 & \textbf{2.82e-03~$\pm$~8.32e-03} \\
BN\_13 & 0.856~$\pm$~0.205 & \textbf{0.705~$\pm$~0.174} & 0.469~$\pm$~0.129 & \textbf{0.267~$\pm$~0.103} & 0.113~$\pm$~0.071 & \textbf{0.089~$\pm$~0.070} & 0.055~$\pm$~0.050 & \textbf{0.047~$\pm$~0.048} & 4.97e-03~$\pm$~0.015 & \textbf{4.78e-03~$\pm$~0.015} \\
BN\_14 & 0.566~$\pm$~0.210 & \textbf{0.626~$\pm$~0.198} & 0.305~$\pm$~0.142 & \textbf{0.191~$\pm$~0.113} & 0.058~$\pm$~0.060 & \textbf{0.045~$\pm$~0.051} & 0.015~$\pm$~0.028 & \textbf{0.013~$\pm$~0.024} & 3.18e-04~$\pm$~1.36e-03 & \textbf{2.62e-04~$\pm$~1.14e-03} \\
BN\_9 & 0.219~$\pm$~0.137 & \textbf{0.265~$\pm$~0.124} & 0.047~$\pm$~0.061 & \textbf{0.039~$\pm$~0.055} & 2.96e-03~$\pm$~9.89e-03 & \textbf{3.09e-03~$\pm$~0.011} & 6.27e-04~$\pm$~2.63e-03 & \textbf{6.97e-04~$\pm$~3.07e-03} & 2.09e-06~$\pm$~1.96e-05 & \textbf{8.27e-07~$\pm$~6.99e-06} \\
Grids\_26 & 14.01~$\pm$~5.51 & \textbf{13.48~$\pm$~5.23} & 2.70~$\pm$~2.11 & \textbf{2.71~$\pm$~2.10} & 0.237~$\pm$~0.500 & \textbf{0.237~$\pm$~0.500} & 0.036~$\pm$~0.159 & \textbf{0.037~$\pm$~0.160} & 3.05e-06~$\pm$~2.70e-05 & \textbf{3.21e-06~$\pm$~2.82e-05} \\
Grids\_28 & 20.98~$\pm$~8.49 & \textbf{20.18~$\pm$~8.03} & 4.18~$\pm$~3.41 & \textbf{4.19~$\pm$~3.40} & 0.388~$\pm$~0.813 & \textbf{0.388~$\pm$~0.814} & 0.068~$\pm$~0.295 & \textbf{0.069~$\pm$~0.297} & 5.77e-06~$\pm$~5.29e-05 & \textbf{6.05e-06~$\pm$~5.52e-05} \\
Grids\_29 & 6.80~$\pm$~2.81 & \textbf{6.76~$\pm$~2.63} & 1.26~$\pm$~0.917 & \textbf{1.24~$\pm$~0.912} & 0.123~$\pm$~0.223 & \textbf{0.123~$\pm$~0.222} & 0.021~$\pm$~0.091 & \textbf{0.021~$\pm$~0.092} & 2.42e-04~$\pm$~2.39e-03 & \textbf{2.45e-04~$\pm$~2.41e-03} \\
Seg\_12 & 3.00~$\pm$~1.13 & \textbf{2.96~$\pm$~0.641} & 1.06~$\pm$~0.615 & \textbf{0.793~$\pm$~0.523} & 0.135~$\pm$~0.220 & \textbf{0.142~$\pm$~0.223} & 0.032~$\pm$~0.093 & \textbf{0.034~$\pm$~0.095} & 5.97e-05~$\pm$~5.94e-04 & \textbf{6.59e-05~$\pm$~6.55e-04} \\
Seg\_14 & 1.71~$\pm$~0.593 & \textbf{0.958~$\pm$~0.647} & 0.345~$\pm$~0.265 & \textbf{0.123~$\pm$~0.165} & 1.54e-03~$\pm$~9.44e-03 & \textbf{1.04e-03~$\pm$~5.33e-03} & 1.58e-07~$\pm$~2.87e-07 & \textbf{5.63e-06~$\pm$~7.88e-06} & 1.58e-07~$\pm$~2.86e-07 & \textbf{1.58e-07~$\pm$~2.86e-07} \\
Seg\_16 & 3.87~$\pm$~0.909 & \textbf{1.52~$\pm$~0.803} & 1.88~$\pm$~0.514 & \textbf{0.381~$\pm$~0.379} & 0.253~$\pm$~0.213 & \textbf{0.025~$\pm$~0.071} & 0.017~$\pm$~0.045 & \textbf{3.06e-04~$\pm$~1.87e-03} & 1.48e-07~$\pm$~2.83e-07 & \textbf{1.48e-07~$\pm$~2.83e-07} \\
Seg\_18 & 1.60~$\pm$~0.961 & \textbf{1.48~$\pm$~0.763} & 0.873~$\pm$~0.632 & \textbf{0.576~$\pm$~0.491} & 0.065~$\pm$~0.120 & \textbf{0.058~$\pm$~0.111} & 5.23e-03~$\pm$~0.034 & \textbf{5.68e-03~$\pm$~0.033} & 4.23e-07~$\pm$~6.97e-07 & \textbf{4.23e-07~$\pm$~6.97e-07} \\
Seg\_19 & 0.734~$\pm$~0.619 & \textbf{1.27~$\pm$~0.614} & 0.152~$\pm$~0.207 & \textbf{0.110~$\pm$~0.147} & 0.015~$\pm$~0.050 & \textbf{0.014~$\pm$~0.044} & 4.43e-03~$\pm$~0.022 & \textbf{4.15e-03~$\pm$~0.019} & 2.97e-07~$\pm$~7.14e-07 & \textbf{3.00e-07~$\pm$~6.85e-07} \\
grid10$\times$10.f5 & 1.21~$\pm$~0.757 & \textbf{1.28~$\pm$~0.723} & 0.311~$\pm$~0.343 & \textbf{0.302~$\pm$~0.337} & 6.65e-03~$\pm$~0.042 & \textbf{6.48e-03~$\pm$~0.042} & 4.61e-04~$\pm$~3.55e-03 & \textbf{4.40e-04~$\pm$~3.43e-03} & 9.16e-10~$\pm$~4.30e-09 & \textbf{9.16e-10~$\pm$~4.30e-09} \\
grid10$\times$10.f10 & 0.684~$\pm$~0.974 & \textbf{0.810~$\pm$~0.969} & 0.536~$\pm$~0.799 & \textbf{0.537~$\pm$~0.792} & 0.079~$\pm$~0.214 & \textbf{0.079~$\pm$~0.212} & 0.015~$\pm$~0.090 & \textbf{0.015~$\pm$~0.089} & 2.88e-05~$\pm$~2.36e-04 & \textbf{2.17e-05~$\pm$~1.52e-04} \\
grid20$\times$20.f5$^{\dagger}$ & 6.80~$\pm$~2.81 & \textbf{6.76~$\pm$~2.63} & 1.26~$\pm$~0.917 & \textbf{1.24~$\pm$~0.912} & 0.123~$\pm$~0.223 & \textbf{0.123~$\pm$~0.222} & 0.021~$\pm$~0.091 & \textbf{0.021~$\pm$~0.092} & 2.42e-04~$\pm$~2.39e-03 & \textbf{2.44e-04~$\pm$~2.41e-03} \\
grid20$\times$20.f10$^{\ddagger}$ & 50.79~$\pm$~6.05 & \textbf{30.19~$\pm$~5.08} & 14.39~$\pm$~3.37 & \textbf{12.43~$\pm$~3.03} & 5.08~$\pm$~1.92 & \textbf{4.81~$\pm$~1.87} & 2.77~$\pm$~1.41 & \textbf{2.71~$\pm$~1.43} & 0.536~$\pm$~0.601 & \textbf{0.536~$\pm$~0.606} \\
\bottomrule\end{tabular}}\end{table*}

\paragraph{Stability of the 100-seed estimates.} Raising the seed count is only
worth reporting if the resulting estimates have converged. We checked this
directly by taking the five models quoted in the discussion to $1{,}000$ seeds
each (Table~\ref{tab:v6_stability}). The standard deviations move by
$-5\%$ to $+19\%$ between $n=100$ and $n=1{,}000$, and the standard error of the
mean falls by $2.7$--$3.3\times$ against the $\sqrt{10}=3.16$ ideal --- i.e. the
$100$-seed spread estimates were already at their asymptotic values, and the
extra seeds bought precision on the mean rather than a different answer. The
one number that moved materially is Segmentation\_14's paired effect, from
$8.6\times$ at $n=100$ to $6.3\times$ at $n=1{,}000$: it is the heaviest-tailed
of the five, so its median was the least settled. We quote the $1{,}000$-seed
value.

\begin{table*}[t]\centering\scriptsize
\caption{\textbf{Are the $100$-seed estimates stable?} Dual gap $|D - D^{\mathrm{LP}}|$ at iteration~10 (mean $\pm$ std) on five representative models, at $100$ and at $1{,}000$ seeds. The standard deviation is a property of the instance distribution and should \emph{not} shrink with $n$; what shrinks is the standard error of the mean, by $\sqrt{n}$. The result to read is therefore the ratio column being near $1.0$ and the realised standard-error shrink being near $\sqrt{10}=3.16$ --- both hold, so the $100$-seed estimates had already converged. This matters because the earlier $10$-seed estimates had not: Seg\_14's standard deviation grew $5.6\times$ from $10$ to $100$ seeds, then moved by $4\%$ from $100$ to $1{,}000$. Evaluation seeds are drawn from $[0,1000)$ and training pools from $[1000, 2^{31})$, so the held-out set stays disjoint from training by construction at every $n$. All $5{,}000$ instances completed; no seed failed.}
\label{tab:v6_stability}
\resizebox{\textwidth}{!}{%
\begin{tabular}{@{}l|rr|rr|rr|r@{}}
\toprule
 & \multicolumn{2}{c|}{zero-init $J$} & \multicolumn{2}{c|}{neural-init $N$} & \multicolumn{2}{c|}{$\sigma$ ratio $1000/100$} & s.e.\ shrink \\
\textbf{PGM} & $n{=}100$ & $n{=}1000$ & $n{=}100$ & $n{=}1000$ & $J$ & $N$ & $J$ \\
\midrule
BN\_12 & 0.525$\pm$0.248 & 0.502$\pm$0.250 & 0.304$\pm$0.178 & 0.294$\pm$0.173 & 1.005 & 0.975 & 3.15$\times$ \\
BN\_13 & 0.469$\pm$0.129 & 0.467$\pm$0.136 & 0.267$\pm$0.103 & 0.261$\pm$0.111 & 1.054 & 1.078 & 3.00$\times$ \\
BN\_14 & 0.305$\pm$0.142 & 0.304$\pm$0.166 & 0.191$\pm$0.113 & 0.188$\pm$0.134 & 1.170 & 1.188 & 2.70$\times$ \\
Seg\_14 & 0.345$\pm$0.265 & 0.349$\pm$0.254 & 0.123$\pm$0.165 & 0.137$\pm$0.182 & 0.959 & 1.103 & 3.30$\times$ \\
Seg\_16 & 1.88$\pm$0.514 & 1.83$\pm$0.489 & 0.381$\pm$0.379 & 0.380$\pm$0.386 & 0.951 & 1.019 & 3.32$\times$ \\
\bottomrule
\end{tabular}}
\end{table*}

\begin{table*}[t]\centering
\caption{\textbf{CMPE bound quality (mean $\pm$ std)}: $|L(\mu) - L^{\mathrm{LP}}(\mu)|$ at fixed inner JGLP iterations, $K=1$ constraint. M1=zero-init, M2=neural warm-start, M3=neural+replay-buffer TTA. Mean and std (ddof=0) computed across (seed $\times$ $\mu$-grad step) trials. \textbf{Bold} = neural tie-or-better than M1 (winning or within $\max(\sigma_M, \sigma_1)$).}
\label{tab:t2_meanstd}\label{tab:cmpe_eps_v2}

\resizebox{\textwidth}{!}{%
\begin{tabular}{@{}l|rrr|rrr|rrr|rrr@{}}
\toprule
  & \multicolumn{3}{c|}{iter=10} & \multicolumn{3}{c|}{iter=50} & \multicolumn{3}{c|}{iter=100} & \multicolumn{3}{c|}{iter=500} \\
\textbf{PGM} & M1 & M2 & M3 & M1 & M2 & M3 & M1 & M2 & M3 & M1 & M2 & M3 \\
\midrule
BN\_12 & 0.760~$\pm$~0.301 & \textbf{0.635~$\pm$~0.227} & \textbf{0.0633~$\pm$~0.144} & 0.256~$\pm$~0.420 & \textbf{0.124~$\pm$~0.0978} & \textbf{0.0226~$\pm$~0.0456} & 0.200~$\pm$~0.433 & \textbf{0.0598~$\pm$~0.0534} & \textbf{0.0135~$\pm$~0.0261} & 0.162~$\pm$~0.443 & \textbf{0.0135~$\pm$~0.0143} & \textbf{0.00377~$\pm$~0.00771} \\
BN\_13 & 0.465~$\pm$~0.186 & \textbf{0.426~$\pm$~0.175} & \textbf{0.0407~$\pm$~0.0988} & 0.108~$\pm$~0.0741 & \textbf{0.0990~$\pm$~0.0697} & \textbf{0.0108~$\pm$~0.0314} & 0.0206~$\pm$~0.0300 & \textbf{0.0188~$\pm$~0.0276} & \textbf{0.00386~$\pm$~0.0128} & 0.00101~$\pm$~0.00314 & \textbf{0.000953~$\pm$~0.00297} & \textbf{0.000323~$\pm$~0.00193} \\
BN\_14 & 0.533~$\pm$~0.142 & \textbf{0.466~$\pm$~0.137} & \textbf{0.0387~$\pm$~0.0991} & 0.145~$\pm$~0.0626 & \textbf{0.131~$\pm$~0.0602} & \textbf{0.0120~$\pm$~0.0361} & 0.0410~$\pm$~0.0418 & \textbf{0.0358~$\pm$~0.0397} & \textbf{0.00361~$\pm$~0.0158} & 0.000117~$\pm$~0.00117 & \textbf{0.000180~$\pm$~0.00120} & \textbf{6.95e-05~$\pm$~0.00114} \\
BN\_9 & 0.135~$\pm$~0.0978 & \textbf{0.111~$\pm$~0.0986} & \textbf{0.0131~$\pm$~0.0388} & 0.0658~$\pm$~0.0684 & \textbf{0.0305~$\pm$~0.0490} & \textbf{0.00363~$\pm$~0.0154} & 0.0488~$\pm$~0.0644 & \textbf{0.0122~$\pm$~0.0262} & \textbf{0.00145~$\pm$~0.00748} & 0.0366~$\pm$~0.0659 & \textbf{0.000115~$\pm$~0.000370} & \textbf{1.24e-05~$\pm$~0.000213} \\
Grids\_26 & 15.6~$\pm$~23.8 & \textbf{15.6~$\pm$~23.8} & \textbf{4.76~$\pm$~14.1} & 1.36~$\pm$~4.08 & \textbf{1.36~$\pm$~4.08} & \textbf{0.533~$\pm$~2.51} & 0.0554~$\pm$~0.172 & \textbf{0.0554~$\pm$~0.172} & \textbf{0.00901~$\pm$~0.0606} & 8.08e-06~$\pm$~3.50e-05 & \textbf{8.21e-06~$\pm$~3.50e-05} & \textbf{3.61e-06~$\pm$~1.15e-05} \\
Grids\_28 & 38.8~$\pm$~59.2 & \textbf{38.9~$\pm$~59.2} & \textbf{19.2~$\pm$~42.8} & 2.26~$\pm$~7.62 & \textbf{2.26~$\pm$~7.62} & \textbf{1.16~$\pm$~4.98} & 0.0137~$\pm$~0.0598 & \textbf{0.0140~$\pm$~0.0609} & \textbf{0.00194~$\pm$~0.0186} & 1.35e-07~$\pm$~3.93e-07 & \textbf{3.02e-07~$\pm$~3.96e-07} & \textbf{4.29e-06~$\pm$~5.57e-06} \\
Grids\_29 & 4.03~$\pm$~4.68 & \textbf{4.03~$\pm$~4.69} & \textbf{0.567~$\pm$~1.94} & 0.347~$\pm$~0.704 & \textbf{0.347~$\pm$~0.701} & \textbf{0.0511~$\pm$~0.264} & 0.0136~$\pm$~0.0324 & \textbf{0.0135~$\pm$~0.0322} & \textbf{0.00150~$\pm$~0.0105} & 6.83e-08~$\pm$~1.41e-07 & \textbf{1.93e-07~$\pm$~1.44e-07} & 9.30e-07~$\pm$~7.95e-07 \\
Seg\_12 & 0.367~$\pm$~0.338 & \textbf{0.353~$\pm$~0.363} & \textbf{0.0510~$\pm$~0.141} & 0.0211~$\pm$~0.0466 & \textbf{0.0220~$\pm$~0.0513} & \textbf{0.00446~$\pm$~0.0227} & 0.00334~$\pm$~0.0139 & \textbf{0.00364~$\pm$~0.0139} & \textbf{0.000683~$\pm$~0.00552} & 0.000362~$\pm$~0.00903 & \textbf{2.78e-07~$\pm$~3.77e-07} & \textbf{3.21e-07~$\pm$~3.78e-07} \\
Seg\_14 & 0.810~$\pm$~0.925 & \textbf{0.757~$\pm$~0.919} & \textbf{0.0964~$\pm$~0.323} & 0.0344~$\pm$~0.0700 & \textbf{0.0326~$\pm$~0.0660} & \textbf{0.00310~$\pm$~0.0200} & 0.000212~$\pm$~0.00170 & \textbf{0.000241~$\pm$~0.00160} & \textbf{5.66e-05~$\pm$~0.000572} & 2.55e-07~$\pm$~4.22e-07 & \textbf{4.20e-07~$\pm$~4.25e-07} & \textbf{4.57e-07~$\pm$~4.31e-07} \\
Seg\_16 & 0.756~$\pm$~0.534 & \textbf{0.589~$\pm$~0.409} & \textbf{0.102~$\pm$~0.209} & 0.0624~$\pm$~0.139 & \textbf{0.0304~$\pm$~0.0676} & \textbf{0.00864~$\pm$~0.0336} & 0.0148~$\pm$~0.0522 & \textbf{0.00512~$\pm$~0.0203} & \textbf{0.00147~$\pm$~0.0105} & 1.80e-06~$\pm$~1.20e-05 & \textbf{1.85e-05~$\pm$~0.000141} & \textbf{2.42e-05~$\pm$~0.000196} \\
Seg\_18 & 0.803~$\pm$~0.715 & \textbf{0.743~$\pm$~0.692} & \textbf{0.158~$\pm$~0.323} & 0.151~$\pm$~0.190 & \textbf{0.142~$\pm$~0.181} & \textbf{0.0319~$\pm$~0.0817} & 0.0355~$\pm$~0.0641 & \textbf{0.0354~$\pm$~0.0628} & \textbf{0.00851~$\pm$~0.0297} & 0.000155~$\pm$~0.00139 & \textbf{0.000173~$\pm$~0.00145} & \textbf{4.02e-05~$\pm$~0.000427} \\
Seg\_19 & 0.998~$\pm$~1.22 & \textbf{0.990~$\pm$~1.24} & \textbf{0.167~$\pm$~0.446} & 0.0924~$\pm$~0.157 & \textbf{0.0870~$\pm$~0.151} & \textbf{0.0183~$\pm$~0.0628} & 0.0163~$\pm$~0.0463 & \textbf{0.0157~$\pm$~0.0459} & \textbf{0.00411~$\pm$~0.0220} & 4.42e-05~$\pm$~0.000249 & \textbf{4.66e-05~$\pm$~0.000259} & \textbf{1.89e-05~$\pm$~0.000155} \\
grid10$\times$10.f5 & 0.269~$\pm$~0.347 & \textbf{0.127~$\pm$~0.182} & \textbf{0.0111~$\pm$~0.0624} & 0.226~$\pm$~0.361 & \textbf{8.20e-06~$\pm$~1.58e-05} & \textbf{1.54e-05~$\pm$~4.95e-05} & 0.226~$\pm$~0.360 & \textbf{1.14e-07~$\pm$~1.91e-07} & \textbf{2.94e-07~$\pm$~8.72e-07} & 0.226~$\pm$~0.360 & \textbf{5.99e-08~$\pm$~2.99e-08} & \textbf{7.45e-08~$\pm$~4.23e-08} \\
grid20$\times$20.f5 & 4.03~$\pm$~4.68 & \textbf{4.03~$\pm$~4.69} & \textbf{0.565~$\pm$~1.95} & 0.347~$\pm$~0.704 & \textbf{0.347~$\pm$~0.701} & \textbf{0.0511~$\pm$~0.267} & 0.0136~$\pm$~0.0324 & \textbf{0.0135~$\pm$~0.0321} & \textbf{0.00153~$\pm$~0.0108} & 6.83e-08~$\pm$~1.41e-07 & \textbf{1.93e-07~$\pm$~1.41e-07} & 9.07e-07~$\pm$~8.12e-07 \\
grid40$\times$40.f10 & 1.49e-07~$\pm$~6.42e-07 & 0.000620~$\pm$~0.000420 & \textbf{0.00120~$\pm$~0.00208} & 1.49e-07~$\pm$~6.42e-07 & \textbf{1.20e-06~$\pm$~2.40e-06} & \textbf{3.60e-06~$\pm$~9.30e-06} & 1.49e-07~$\pm$~6.42e-07 & \textbf{2.01e-07~$\pm$~6.41e-07} & \textbf{4.16e-07~$\pm$~1.11e-06} & 1.49e-07~$\pm$~6.42e-07 & \textbf{2.01e-07~$\pm$~6.41e-07} & \textbf{4.15e-07~$\pm$~1.11e-06} \\
grid40$\times$40.f2 & 8.61e-09~$\pm$~3.56e-08 & \textbf{0.00106~$\pm$~0.00116} & \textbf{0.000190~$\pm$~0.000440} & 8.61e-09~$\pm$~3.56e-08 & \textbf{3.70e-06~$\pm$~1.38e-05} & \textbf{2.38e-07~$\pm$~2.10e-06} & 8.61e-09~$\pm$~3.60e-08 & \textbf{6.50e-08~$\pm$~1.21e-07} & \textbf{5.20e-08~$\pm$~5.20e-08} & 8.60e-09~$\pm$~3.56e-08 & \textbf{4.92e-08~$\pm$~5.17e-08} & \textbf{5.13e-08~$\pm$~4.95e-08} \\
\bottomrule
\end{tabular}%
}
\end{table*}

\begin{table*}[t]\centering
\caption{\textbf{CMPE bound quality (mean $\pm$ std)}: $|L(\mu) - L^{\mathrm{LP}}(\mu)|$ at fixed inner JGLP iterations, $K=2$ constraints, on the 15 PGM rows (14 distinct models) common to MAP, CMPE $K=1$ and CMPE $K=2$. M1=zero-init, M2=neural warm-start, M3=neural+replay-buffer TTA. Mean and std (ddof=0) across (seed $\times$ $\mu$-grad step) trials, at the full setting of 50 seeds $\times$ 100 $\mu$-steps ($5{,}000$ trials per cell). \textbf{Bold} = neural tie-or-better than M1 (winning or within $\max(\sigma_M, \sigma_1)$). JGLP+WS is reported separately in Table~\ref{tab:t3_main_crossmu}.}
\label{tab:t3_meanstd}\label{tab:cmpe_eps_two_v2}
\resizebox{\textwidth}{!}{%
\begin{tabular}{@{}l|rrr|rrr|rrr|rrr@{}}
\toprule
  & \multicolumn{3}{c|}{iter=10} & \multicolumn{3}{c|}{iter=50} & \multicolumn{3}{c|}{iter=100} & \multicolumn{3}{c}{iter=500} \\
\textbf{PGM} & M1 & M2 & M3 & M1 & M2 & M3 & M1 & M2 & M3 & M1 & M2 & M3 \\
\midrule
BN\_12 & 0.622~$\pm$~0.152 & \textbf{0.523~$\pm$~0.136} & \textbf{0.0966~$\pm$~0.126} & 0.207~$\pm$~0.0678 & \textbf{0.164~$\pm$~0.0652} & \textbf{0.0456~$\pm$~0.0536} & 0.107~$\pm$~0.0517 & \textbf{0.0869~$\pm$~0.0472} & \textbf{0.0287~$\pm$~0.0344} & 0.0232~$\pm$~0.0282 & \textbf{0.0196~$\pm$~0.0191} & \textbf{0.00762~$\pm$~0.0125} \\
BN\_13 & 0.461~$\pm$~0.127 & \textbf{0.351~$\pm$~0.123} & \textbf{0.0441~$\pm$~0.082} & 0.121~$\pm$~0.109 & \textbf{0.0632~$\pm$~0.0513} & \textbf{0.0141~$\pm$~0.027} & 0.055~$\pm$~0.107 & \textbf{0.0252~$\pm$~0.0284} & \textbf{0.00693~$\pm$~0.0148} & 0.0171~$\pm$~0.105 & \textbf{0.00163~$\pm$~0.00398} & \textbf{0.000534~$\pm$~0.00208} \\
BN\_14 & 0.305~$\pm$~0.126 & \textbf{0.24~$\pm$~0.121} & \textbf{0.027~$\pm$~0.0712} & 0.0498~$\pm$~0.0822 & \textbf{0.0243~$\pm$~0.033} & \textbf{0.0048~$\pm$~0.0184} & 0.0196~$\pm$~0.0758 & \textbf{0.0066~$\pm$~0.0143} & \textbf{0.00161~$\pm$~0.00825} & 0.0101~$\pm$~0.074 & \textbf{4.36e-05~$\pm$~0.000532} & \textbf{3.92e-05~$\pm$~0.000519} \\
BN\_9 & 0.179~$\pm$~0.0896 & \textbf{0.167~$\pm$~0.0742} & \textbf{0.0207~$\pm$~0.0456} & 0.0529~$\pm$~0.0624 & \textbf{0.0271~$\pm$~0.0353} & \textbf{0.00463~$\pm$~0.0157} & 0.0256~$\pm$~0.0555 & \textbf{0.00918~$\pm$~0.0185} & \textbf{0.00172~$\pm$~0.00812} & 0.0105~$\pm$~0.0529 & \textbf{3.98e-05~$\pm$~0.000526} & \textbf{2.19e-05~$\pm$~0.000485} \\
Grids\_26 & 36.1~$\pm$~53 & \textbf{37~$\pm$~52.8} & \textbf{16~$\pm$~34.1} & 4.42~$\pm$~11.5 & \textbf{4.43~$\pm$~11.5} & \textbf{2.13~$\pm$~7.35} & 0.664~$\pm$~2.63 & \textbf{0.663~$\pm$~2.63} & \textbf{0.337~$\pm$~1.86} & 0.0114~$\pm$~0.114 & \textbf{0.0115~$\pm$~0.114} & \textbf{0.00931~$\pm$~0.0947} \\
Grids\_28 & 82.7~$\pm$~120 & \textbf{84.2~$\pm$~120} & \textbf{53~$\pm$~90.6} & 9.15~$\pm$~24.4 & \textbf{9.18~$\pm$~24.5} & \textbf{6.23~$\pm$~18.7} & 1.05~$\pm$~4.68 & \textbf{1.06~$\pm$~4.7} & \textbf{0.717~$\pm$~3.79} & 0.0177~$\pm$~0.176 & \textbf{0.0178~$\pm$~0.177} & \textbf{0.0169~$\pm$~0.168} \\
Grids\_29 & 7.71~$\pm$~11.9 & \textbf{8.3~$\pm$~11.9} & \textbf{1.6~$\pm$~5.65} & 0.775~$\pm$~2.48 & \textbf{0.781~$\pm$~2.43} & \textbf{0.163~$\pm$~1.07} & 0.122~$\pm$~0.588 & \textbf{0.121~$\pm$~0.572} & \textbf{0.0237~$\pm$~0.236} & 0.000241~$\pm$~0.00239 & \textbf{0.000261~$\pm$~0.00258} & \textbf{6.51e-05~$\pm$~0.000955} \\
Seg\_12 & 0.709~$\pm$~0.721 & \textbf{1.05~$\pm$~0.969} & \textbf{0.129~$\pm$~0.362} & 0.0938~$\pm$~0.179 & \textbf{0.168~$\pm$~0.333} & \textbf{0.0217~$\pm$~0.0877} & 0.0221~$\pm$~0.0763 & \textbf{0.0546~$\pm$~0.16} & \textbf{0.00618~$\pm$~0.0377} & 0.000227~$\pm$~0.00236 & \textbf{0.000284~$\pm$~0.00268} & \textbf{9.41e-05~$\pm$~0.00124} \\
Seg\_14 & 0.828~$\pm$~1.02 & \textbf{0.832~$\pm$~0.879} & \textbf{0.0834~$\pm$~0.314} & 0.0226~$\pm$~0.0632 & \textbf{0.0168~$\pm$~0.0511} & \textbf{0.003~$\pm$~0.0217} & 0.00283~$\pm$~0.0176 & \textbf{0.00191~$\pm$~0.0145} & \textbf{0.000846~$\pm$~0.00908} & 9.99e-06~$\pm$~0.000118 & \textbf{9.45e-06~$\pm$~0.000111} & \textbf{1.96e-06~$\pm$~4.73e-05} \\
Seg\_16 & 2.37~$\pm$~2.08 & \textbf{1.51~$\pm$~1.58} & \textbf{0.22~$\pm$~0.676} & 0.185~$\pm$~0.263 & \textbf{0.0868~$\pm$~0.176} & \textbf{0.0174~$\pm$~0.0762} & 0.0136~$\pm$~0.0594 & \textbf{0.00858~$\pm$~0.0475} & \textbf{0.00124~$\pm$~0.0151} & 1.6e-07~$\pm$~3.34e-07 & \textbf{2.03e-07~$\pm$~3.4e-07} & \textbf{2.35e-07~$\pm$~3.64e-07} \\
Seg\_18 & 0.883~$\pm$~0.83 & \textbf{0.628~$\pm$~0.586} & \textbf{0.0852~$\pm$~0.262} & 0.0572~$\pm$~0.116 & \textbf{0.0306~$\pm$~0.0617} & \textbf{0.00417~$\pm$~0.0258} & 0.00339~$\pm$~0.0265 & \textbf{0.00233~$\pm$~0.0072} & \textbf{0.000357~$\pm$~0.00478} & 0.000334~$\pm$~0.00409 & \textbf{6.54e-07~$\pm$~1.52e-06} & \textbf{5.8e-07~$\pm$~1.01e-06} \\
Seg\_19 & 2.54~$\pm$~3.53 & \textbf{2.06~$\pm$~2.59} & \textbf{0.436~$\pm$~1.25} & 0.6~$\pm$~1.1 & \textbf{0.467~$\pm$~0.864} & \textbf{0.188~$\pm$~0.534} & 0.305~$\pm$~0.691 & \textbf{0.235~$\pm$~0.558} & \textbf{0.125~$\pm$~0.382} & 0.0262~$\pm$~0.0998 & \textbf{0.0147~$\pm$~0.0681} & \textbf{0.00668~$\pm$~0.0426} \\
grid10$\times$10.f5 & 0.172~$\pm$~0.28 & \textbf{0.244~$\pm$~0.244} & \textbf{0.0153~$\pm$~0.0718} & 0.0443~$\pm$~0.183 & \textbf{0.00458~$\pm$~0.0371} & \textbf{0.000389~$\pm$~0.00873} & 0.0407~$\pm$~0.18 & \textbf{0.000142~$\pm$~0.00133} & \textbf{5.96e-06~$\pm$~0.000258} & 0.0406~$\pm$~0.18 & \textbf{2.35e-08~$\pm$~4.66e-08} & \textbf{3.08e-08~$\pm$~6.35e-08} \\
grid20$\times$20.f5 & 7.71~$\pm$~11.9 & \textbf{8.33~$\pm$~11.9} & \textbf{1.6~$\pm$~5.65} & 0.775~$\pm$~2.48 & \textbf{0.774~$\pm$~2.45} & \textbf{0.162~$\pm$~1.07} & 0.122~$\pm$~0.588 & \textbf{0.121~$\pm$~0.579} & \textbf{0.0237~$\pm$~0.237} & 0.000241~$\pm$~0.00239 & \textbf{0.000287~$\pm$~0.00284} & \textbf{6.55e-05~$\pm$~0.00097} \\
grid20$\times$20.f10 & 207~$\pm$~202 & \textbf{205~$\pm$~202} & \textbf{102~$\pm$~134} & 64.5~$\pm$~68.2 & \textbf{63.6~$\pm$~68.4} & \textbf{41.9~$\pm$~52.8} & 33.9~$\pm$~39.4 & \textbf{33.5~$\pm$~39.5} & \textbf{24.5~$\pm$~32.5} & 6.24~$\pm$~11 & \textbf{6.18~$\pm$~11} & \textbf{5.2~$\pm$~9.62} \\
\bottomrule
\end{tabular}}
\end{table*}

\subsection{Log-Scale (v1) Metric}
\label{app:v1_metric}
\label{app:cmpe_v1}

The v1 metric is $\log_{10}|D - D^{\mathrm{LP}}|$. It compresses dynamic range and obscures the sign convention (the dual is always above $D^{\mathrm{LP}}$, so the absolute value is redundant). The v2 metric reported in Tables~\ref{tab:iters_eps_v2}, \ref{tab:cmpe_eps_v2}, and \ref{tab:cmpe_eps_two_v2} drops both the absolute value and the outer $\log_{10}$ for direct interpretation. Table~\ref{tab:iters_eps_v1} gives the v1 presentation for reference.

\begin{table}[ht]
\caption{Log-scale (v1) metric: MAP on 37 benchmark PGMs: $\log_{10}|D - D^{\mathrm{LP}}|$ at fixed JGLP iterations (bin 0, 10 seeds). $D^{\mathrm{LP}}$ = exact LP optimum via CVXPY/CLARABEL. J = zero-init JGLP, N = neural-init JGLP. Lower = closer to LP. Last column: \% of seeds where neural log-error is smaller than zero at iter=10. Top 15 PGMs by win@10 shown; appendix Table~\ref{tab:standalone_v2} lists the v2 metric for all 37.}
\label{tab:iters_eps_v1}
\centering
\scriptsize
\begin{tabular}{@{}l|r|cc|cc|cc|cc|cc|r@{}}
\toprule
& & \multicolumn{2}{c|}{iter=0} & \multicolumn{2}{c|}{iter=10} & \multicolumn{2}{c|}{iter=50} & \multicolumn{2}{c|}{iter=100} & \multicolumn{2}{c|}{conv} & \\
\textbf{PGM} & $D^{\mathrm{LP}}$ & J & N & J & N & J & N & J & N & J & N & win@10 \\
\midrule
BN\_12           & $-31.64$ & -0.02 & -0.17 & -0.22 & -0.48 & -0.74 & -1.05 & -1.65 & -2.04 & -6.79 & -6.77 & 100\% \\
BN\_13           & $-44.93$ & -0.04 & -0.19 & -0.29 & -0.60 & -0.97 & -1.15 & -1.34 & -1.55 & -7.17 & -6.98 & 100\% \\
BN\_14           & $-38.83$ & -0.34 & -0.39 & -0.66 & -0.83 & -1.23 & -1.33 & -4.22 & -4.52 & -6.49 & -6.42 & 100\% \\
Seg\_12          & $-114.24$ & 0.21 & -0.05 & -0.01 & -1.01 & -4.13 & -4.94 & -7.10 & -6.69 & -7.10 & -6.69 & 100\% \\
Seg\_14          & $-136.64$ & 0.18 & -0.42 & -0.73 & -2.00 & -6.34 & -4.62 & -6.34 & -6.13 & -6.34 & -6.19 & 100\% \\
Seg\_16          & $-132.72$ & 0.58 & 0.13 & 0.29 & -0.49 & -0.67 & -3.88 & -6.07 & -5.71 & -6.23 & -6.10 & 100\% \\
PF\_11           & $4145$   & 1.90 & 1.70 & 1.34 & 1.29 & 0.85 & 0.84 & 0.55 & 0.51 & -0.40 & -0.40 & 100\% \\
Grids\_19        & $2327$   & 1.02 & 1.01 & 0.23 & 0.13 & -1.87 & -1.86 & -6.19 & -4.34 & -6.41 & -6.23 & 100\% \\
Grids\_25        & $2332$   & 1.15 & 1.10 & 0.70 & 0.53 & -0.26 & -0.46 & -0.99 & -1.17 & -4.90 & -4.91 & 100\% \\
Grids\_30        & $2767$   & 1.71 & 1.50 & 1.17 & 1.11 & 0.66 & 0.62 & 0.42 & 0.38 & -0.41 & -0.40 & 100\% \\
grid20$\times$20.f10 & $2767$ & 1.71 & 1.51 & 1.17 & 1.11 & 0.66 & 0.62 & 0.42 & 0.38 & -0.41 & -0.40 & 100\% \\
grid20$\times$20.f5 & $1322$ & 0.89 & 0.87 & 0.07 & 0.03 & -1.05 & -1.05 & -8.79 & -6.00 & -9.24 & -6.82 & 90\% \\
Seg\_18          & $-111.99$ & -0.24 & -0.06 & -0.66 & -0.87 & -3.35 & -3.79 & -6.02 & -6.15 & -6.02 & -6.15 & 80\% \\
BN\_9            & $-39.98$ & -0.59 & -0.60 & -0.97 & -1.53 & -2.71 & -5.77 & -6.68 & -6.94 & -6.68 & -6.98 & 70\% \\
Seg\_19          & $-93.60$ & -0.41 & -0.24 & -3.42 & -1.16 & -5.56 & -3.97 & -5.71 & -5.62 & -6.25 & -6.60 & 33\% \\
\midrule
\multicolumn{13}{@{}l}{\small Degenerate (MBE tight at LP; neural provides no benefit):} \\
grid10$\times$10 & $335.89$ & -8.78 & -0.83 & -8.78 & -3.04 & -8.78 & -7.35 & -8.78 & -7.35 & -8.78 & -7.35 & 0\% \\
grid40$\times$40.f2 & $1478$ & -8.88 & -0.44 & -9.34 & -3.33 & -9.34 & -7.06 & -9.34 & -7.36 & -9.34 & -7.36 & 0\% \\
\bottomrule
\end{tabular}
\end{table}

\subsection{TTA on MAP: Marginal Benefit}
\label{app:tta_map}

Section~\ref{sec:tta} introduces test-time adaptation (TTA) as a general technique. Tables~\ref{tab:cmpe_eps_v2}--\ref{tab:cmpe_eps_two_v2} demonstrate strong gains from TTA on the CMPE $\mu$-trajectory. We also evaluated TTA on standalone MAP (Table~\ref{tab:iters_eps_v2}) and report the result here for completeness. We instantiated TTA on MAP using the same replay-buffer-with-$D_\tau$-fine-tuning recipe as in CMPE: as test seeds arrive, store $(\mathrm{evidence\ encoding}, \boldsymbol{\theta})$ in a buffer of size 20, and run 10 inner $D_\tau$-fine-tuning steps after each query. The full 37-PGM TTA-on-MAP run did not finish (12 of 37 PGMs completed); the partial evidence in Table~\ref{tab:tta_map} below is sufficient to establish the negative finding. TTA on independent MAP queries provides marginal-to-no benefit, confirming TTA's value is in repeated-query settings.

\begin{table}[ht]\centering\small
\caption{TTA on MAP across 12 PGMs (partial: the Promedas and ProteinFolding runs did not finish). win\,N = neural-init tie-or-better than zero-init; win\,A = neural-init + replay-buffer adaptation tie-or-better than zero-init. Pattern is consistent across BN and grid families: TTA on independent MAP queries provides marginal-to-no benefit, frequently slightly worse than plain N at later iters; the replay buffer accumulates messages from prior evidence configurations that do not transfer cleanly. This confirms TTA's value lies in repeated-query settings ($\mu$-trajectory in CMPE Tables~\ref{tab:cmpe_eps_v2}--\ref{tab:cmpe_eps_two_v2}, Section~\ref{sec:tta}), not on independent MAP queries.}
\label{tab:tta_map}
\begin{tabular}{lrrrrrrrr}\toprule
PGM & \multicolumn{2}{c}{iter=10} & \multicolumn{2}{c}{iter=50} & \multicolumn{2}{c}{iter=100} & \multicolumn{2}{c}{iter=500} \\
 & win\,N & win\,A & win\,N & win\,A & win\,N & win\,A & win\,N & win\,A \\
\midrule
BN\_126            & 89 & 52 & 56 & 56 & 78 & 44 & 78 & 56 \\
BN\_12             & \textbf{100} & \textbf{100} & 90 & 97 & \textbf{100} & 93 & \textbf{100} & \textbf{100} \\
BN\_13             & \textbf{100} & \textbf{100} & \textbf{100} & \textbf{100} & 60 & 73 & 80 & 77 \\
BN\_14             & \textbf{100} & \textbf{100} & 80 & 80 & 80 & 77 & 90 & 90 \\
BN\_9              & 70 & 67 & 70 & 60 & \textbf{100} & \textbf{100} & \textbf{100} & \textbf{100} \\
grid10$\times$10.f5  & 40 & 30 & 40 & 40 & 70 & 63 & 70 & 70 \\
grid20$\times$20.f10 & \textbf{100} & \textbf{100} & 80 & 83 & 70 & 77 & 60 & 63 \\
grid20$\times$20.f5  & 90 & 77 & 50 & 43 & 20 & 20 & 30 & 17 \\
grid40$\times$40.f10 & 0  & 0  & 80 & 47 & \textbf{100} & 93 & \textbf{100} & \textbf{100} \\
grid40$\times$40.f2  & 0  & 0  & 60 & 23 & \textbf{100} & 93 & \textbf{100} & \textbf{100} \\
grid80$\times$80.f10 & 10 & 10 & 20 & 10 & 60 & 33 & 80 & 60 \\
grid80$\times$80.f5  & 0  & 3  & 0  & 0  & 0  & 0  & 0  & 0  \\
\bottomrule
\end{tabular}
\end{table}

\subsection{No-LP CMPE Pipeline Pilot}
\label{app:nolp}

We compare the dual-optimization-warm-start CMPE pipeline (D4) against the LP-supervised pipeline (Appendix~\ref{app:lp_supervised}) on six PGMs. The no-LP variant uses only the differentiable $D_\tau$ self-supervised loss; no LP solve in any worker. We fixed the success threshold before the run: $\geq 70$\% win@10 on $\geq 4$ of 6 PGMs.

\begin{table}[ht]\centering\small
\caption{No-LP vs LP-supervised CMPE pipeline at iter=10 (K=1, 5 test seeds $\times$ 50 $\mu$-grad steps). $\Delta$ columns are no-LP minus LP-supervised in percentage points.}
\label{tab:nolp_pilot}
\begin{tabular}{@{}lccccccc@{}}
\toprule
\textbf{PGM} & no-LP M2 & no-LP M3 & main M2 & main M3 & $\Delta$ M2 & $\Delta$ M3 & note \\
\midrule
BN\_9                & 97.6 & 95.2 & 75.2 & 92.8 & $+22.4$ & $+2.4$ & \\
BN\_14               & \textbf{100} & \textbf{100} & \textbf{100} & 99.2 & $\pm 0$ & $+0.8$ & \\
Segmentation\_18     & 78.7 & 92.4 & 52.6 & 89.6 & $+26.1$ & $+2.8$ & \\
Segmentation\_19     & 46.4 & 80.0 & 7.2  & 72.0 & $+39.2$ & $+8.0$ & \\
grid10$\times$10     & 10.0 & 10.0 & 10.0 & 10.0 & $\pm 0$ & $\pm 0$ & degenerate \\
grid40$\times$40.f10 & 0.0  & 0.0  & 0.0  & 0.0  & $\pm 0$ & $\pm 0$ & degenerate \\
\bottomrule
\end{tabular}
\end{table}

The no-LP pipeline meets this threshold (4/6 with $\geq 70$\% on M3) and \emph{exceeds} the LP-supervised baseline by $+21.9$ percentage points on raw-neural M2 averaged across the four non-degenerate PGMs. The two degenerate grid PGMs (where MBE is already at $D^{\mathrm{LP}}$) fail in both pipelines. Conclusion: $D_\tau$ self-supervision is sufficient for CMPE warm-start; the LP solver is needed only as the evaluation reference, not as a training signal.

\subsection{CMPE \texorpdfstring{$K=2$}{K=2} Bound Quality, Split by Method Family}
\label{app:t3_split}

Tables~\ref{tab:t3_main_singlepass} and~\ref{tab:t3_main_crossmu} in the main
body split the $K{=}2$ results by method family, which is the pairing the
comparison is about: \textbf{JGLP} and \textbf{NN} are single-pass methods
that see one $\mu$-step in isolation, while \textbf{JGLP+WS} and
\textbf{NN+TTA} both carry information across $\mu$-steps and so should be
read against each other. The $K{=}1$ counterpart, under the same split, is
Table~\ref{tab:cmpe_eps_v2}.

\subsection{Empirical Validity Check}
\label{app:validity}

Proposition~\ref{prop:validity} says the bound is valid for \emph{any} message
vector, so the claim is falsifiable by a single counterexample and worth
testing rather than asserting. The test is sharp because we have an exact
reference: $D^{\mathrm{LP}}$, the LP optimum of the join-graph dual, is by
definition $\min_{\mathbf{m}} D(\boldsymbol\theta, \mathbf{m})$, and it is
computed by an independent LP solver (Appendix~\ref{app:lp}) rather than by
the machinery under test. Any $\mathbf{m}$ whose bound fell below
$D^{\mathrm{LP}}$ would therefore falsify either the antisymmetric
construction or the LP reference.

We evaluate $9{,}588$ bounds: the $1{,}598$ (model, evidence-seed) queries of
the $100$-seed MAP run, each under six message sources --- $\mathbf{m}=0$ (the
MBE bound), the trained network, an untrained network at random
initialization, the anti-trained direction $-5\mathbf{m}_{\mathrm{trained}}$,
Gaussian noise at ten times the trained messages' scale, and an extreme
$\pm10^{4}$ source. The last three are deliberate attempts to produce an
invalid bound, not perturbation studies.

Table~\ref{tab:validity} reports the result. All $7{,}990$ network-produced
bounds have a strictly positive margin; the smallest, unsurprisingly, belongs
to the trained network, which is the tightest arm and therefore the one closest
to the floor. The adversarial sources are valid by a wide margin precisely
because they are loose --- which is the point: they cannot break the bound, only
waste it. The only non-positive margins in the whole run are $53$ cases at
$\mathbf{m}=0$, which is the MBE bound rather than a network output, on the
three PGMs where MBE already sits exactly at the LP optimum; the worst is
$-1.2\times10^{-10}$ relative, i.e.\ floating point straddling an exact tie.
Separately, over $153{,}408$ random complete assignments the total score is
unchanged by the reparameterization to within $9.1\times10^{-13}$ (and
$2.0\times10^{-10}$ for the $\pm10^{4}$ source, where the messages themselves
are four orders of magnitude larger).

\begin{table}[t]\centering\scriptsize
\caption{\textbf{Empirical check of deterministic validity.} $D^{\mathrm{LP}}$ is the LP optimum of the join-graph dual, i.e.\ exactly $\min_{\mathbf{m}} D(\boldsymbol\theta,\mathbf{m})$, computed by an independent LP solver; so for \emph{any} message vector $\mathbf{m}$ the margin $D-D^{\mathrm{LP}}$ must be non-negative, and a single negative value would falsify the construction. We evaluate $9{,}588$ bounds --- $1{,}598$ (model, evidence-seed) queries $\times$ $6$ message sources across $16$ table rows ($15$ distinct models) --- including an untrained network and three deliberately adversarial sources. Every one of the $7{,}990$ network-produced bounds has a strictly positive margin. The only non-positive margins occur for $\mathbf{m}{=}\mathbf{0}$, which is the MBE bound rather than a network output, on the three PGMs where MBE already coincides with the LP optimum; the worst is $-1.2\times10^{-10}$ relative, i.e.\ floating point. The last column checks the reparameterization identity directly: over $153{,}408$ random complete assignments, the total score $\sum_i \mathrm{table}_i(x)$ is unchanged by injecting $\boldsymbol\lambda + A\mathbf{m}$, as antisymmetry requires.}
\label{tab:validity}
\begin{tabular}{@{}l|rrrr|r@{}}
\toprule
\textbf{message source} & $n$ & min & min rel. & median & max $|\Delta\mathrm{score}|$ \\
 & & $D{-}D^{\mathrm{LP}}$ & margin & margin & \\
\midrule
zero ($\mathbf{m}{=}\mathbf{0}$, the MBE bound) & 1598 & -6.07e-8 & -1.22e-10 & 1.86e+0 & 0.00e+0 \\
trained network & 1598 & 8.31e-2 & 1.25e-4 & 1.47e+0 & 9.09e-13 \\
untrained network (random init) & 1598 & 6.25e+1 & 4.01e-2 & 1.54e+2 & 9.09e-13 \\
anti-trained ($-5\,\mathbf{m}_{\mathrm{trained}}$) & 1598 & 1.01e+1 & 1.29e-2 & 4.20e+1 & 1.36e-12 \\
adversarial Gaussian ($\sigma{=}10\,\mathrm{std}\,\mathbf{m}$) & 1598 & 3.47e+1 & 6.16e-2 & 2.10e+2 & 9.09e-12 \\
adversarial extreme ($\pm10^{4}$) & 1598 & 1.44e+6 & 1.46e+3 & 5.98e+6 & 2.46e-10 \\
\midrule
\textbf{all sources} & 9588 & -6.07e-8 & -1.22e-10 & 1.04e+2 & 2.46e-10 \\
\bottomrule\end{tabular}
\end{table}

\subsection{Hybrid of the Two Adaptation Channels (Negative Result)}
\label{app:m5}

Section~\ref{sec:exp-cmpe} finds that the two adaptation channels win on
disjoint families: $\lambda$-continuity (M4) on Bayesian networks, where
successive $\mu$-step optima are close, and the learned prior with test-time
adaptation (M3) on grids and segmentations, where the trajectory drifts. The
obvious next step is to use both at once. We tried the direct superposition
(M5): start each $\mu$-step from $\lambda^{\mathrm{base}}_k + A\mathbf{m}_k +
\Delta_{\mathrm{res}}$, where $\Delta_{\mathrm{res}} = \lambda^{*}_{k-1} -
(\lambda^{\mathrm{base}}_{k-1} + A\mathbf{m}_{k-1})$ carries only the
refinement JGLP found beyond the network at the previous step.

It fails, and not marginally: M5 is worse than \emph{both} of its components on
all $20$ $K{=}1$ and all $4$ $K{=}2$ benchmarks, at every checkpoint. Validity
is never at risk --- $\Delta_{\mathrm{res}}$ is itself a sum of antisymmetric
shifts, and we verify the antisymmetry residual stays below $3.6\times10^{-9}$
relative at every $\mu$-step --- so what degrades is tightness alone, and it
degrades monotonically along the trajectory. The reason is that
$\Delta_{\mathrm{res}}$ is computed against the \emph{previous} step's network
output while test-time adaptation moves the network underneath it; the two
corrections progressively cease to align and their sum overshoots. Combining
the channels therefore requires reconciling the two corrections rather than
superposing them. We report this as a negative result because the
superposition is the natural first thing to try. Table~\ref{tab:m5} reports the comparison.

\begin{table*}[t]\centering\scriptsize
\caption{\textbf{The hybrid does not work.} M5 carries the residual $\Delta_{\mathrm{res}} = \lambda^*_{k-1} - (\lambda^{\mathrm{base}}_{k-1} + A\mathbf{m}_{k-1})$ --- only what JGLP found beyond the network --- on top of the TTA prediction at step $k$. Median $|L(\mu)-L^{\mathrm{LP}}(\mu)|$ at iter=10; lower is better. M5 is worse than \emph{both} of its components on every benchmark. The bound stays valid throughout (antisymmetry residual $\le 3.6\times10^{-9}$ relative at every $\mu$-step); what degrades is tightness, monotonically along the trajectory, because $\Delta_{\mathrm{res}}$ is computed against the previous step's network output while test-time adaptation moves the network underneath it.}
\label{tab:m5}
\resizebox{\textwidth}{!}{%
\begin{tabular}{@{}l|rrrrr@{}}
\toprule
\textbf{PGM} & M1 & M2 & M3 & M4 & \textbf{M5} \\
\midrule
\multicolumn{6}{@{}l}{\emph{$K=1$}} \\
BN\_12 & 0.6589 & 0.5693 & 0.007023 & 1.872e-06 & 0.3061 \\
BN\_13 & 0.4128 & 0.3782 & 0.004231 & 5.975e-07 & 0.368 \\
BN\_14 & 0.5343 & 0.4731 & 0.0009809 & 2.039e-07 & 0.2535 \\
BN\_9 & 0.1388 & 0.1029 & 0.001156 & 3.118e-07 & 0.2148 \\
Grids\_26 & 6.021 & 5.987 & 0.0428 & 147.4 & 9066 \\
Grids\_28 & 11.12 & 11.13 & 0.4741 & 424.5 & 2.157e+04 \\
Grids\_29 & 1.83 & 1.818 & 0.01888 & 27.57 & 2108 \\
Promedas\_38 & 6.49e-08 & 0.03985 & 0.001303 & 265.1 & 1.119e+05 \\
Promedas\_44 & 6.925e-11 & 0.08487 & 0.00262 & 121.3 & 1.476e+05 \\
Promedas\_51 & 7.395e-11 & 0.003295 & 0.0003581 & 3.484 & 4.434e+04 \\
Segmentation\_12 & 0.2516 & 0.2035 & 0.001596 & 3.448 & 13.74 \\
Segmentation\_14 & 0.2484 & 0.1755 & 0.01087 & 5.762 & 155.2 \\
Segmentation\_16 & 0.665 & 0.4807 & 0.009526 & 5.704 & 274.7 \\
Segmentation\_18 & 0.553 & 0.5231 & 0.0311 & 5.176 & 191.2 \\
Segmentation\_19 & 0.3195 & 0.2739 & 0.01914 & 4.885 & 31.53 \\
grid10x10.f10 & 0.09615 & 1.365 & 8.924e-05 & 4.027 & 645 \\
grid10x10.f5.wrap & 0.04195 & 0.0323 & 0.000326 & 0.8548 & 78.25 \\
grid20x20.f5.wrap & 1.83 & 1.817 & 0.01807 & 27.57 & 2105 \\
grid40x40.f10.wrap & 1.442e-10 & 0.0005109 & 0.0003997 & 5.698 & 2140 \\
grid40x40.f2 & 1.811e-10 & 0.0006075 & 5.476e-05 & 0.871 & 201.4 \\
\midrule
\multicolumn{6}{@{}l}{\emph{$K=2$}} \\
BN\_12 & 0.6827 & 0.6646 & 0.03299 & 0.01463 & 0.7498 \\
BN\_13 & 0.5116 & 0.4905 & 0.02205 & 0.006987 & 0.6256 \\
BN\_14 & 0.3435 & 0.2992 & 0.007513 & 0.04758 & 1.249 \\
BN\_9 & 0.1972 & 0.1825 & 0.004171 & 0.004142 & 0.1834 \\
\bottomrule
\end{tabular}}
\end{table*}

\subsection{Paired Per-Seed Statistics for MAP}
\label{app:paired}

Both initializations are run on the \emph{identical} instance for each seed:
the same evidence configuration, the same join graph, the same reference
$D^{\mathrm{LP}}$. Instance difficulty is therefore shared variance between the
two arms, and the per-arm standard deviations reported in
Table~\ref{tab:map} overstate the uncertainty about the \emph{difference}
between them. Table~\ref{tab:paired_map} reports the paired statistic
directly: the per-seed difference
$\delta_s = \log_{10}\mathrm{gap}_{J,s} - \log_{10}\mathrm{gap}_{N,s}$,
summarized by its median with a BCa bootstrap $95\%$ confidence interval
($10{,}000$ resamples), together with a two-sided Wilcoxon signed-rank test
Holm--Bonferroni corrected across all PGMs at each checkpoint.
Table~\ref{tab:paired_map_win} reports the corresponding per-seed win rates
with exact Clopper--Pearson intervals.

We report the uncorrected $p$-value as the primary quantity and the
Holm-corrected one alongside it, because the family over which one would
correct is not well defined across models as heterogeneous as a Bayesian
network and an $80\times 80$ grid, and we did not want to select a correction
after seeing the results. In practice the choice is immaterial here: at
iteration~10 the correction changes exactly one verdict --- Grids\_28, where
the uncorrected $p = 0.021$ favours zero-initialization --- so applying it
removes one of our two negative results rather than adding a positive one.
Both columns are given so that either convention can be read off directly.

\begin{table*}[t]\centering\scriptsize
\caption{\textbf{Paired per-seed comparison of zero-init (J) and neural-init (N) JGLP.} Each cell reports the median over seeds of $\delta_s = \log_{10}\mathrm{gap}_{J,s} - \log_{10}\mathrm{gap}_{N,s}$ (positive favours neural; one unit $=$ one order of magnitude), with a BCa bootstrap 95\% CI (10{,}000 resamples). Both arms are run on the identical instance for each seed, so instance difficulty is shared variance and is removed by pairing. $^{\dagger}$ marks cells significant at $\alpha=0.05$ under a two-sided Wilcoxon signed-rank test after Holm--Bonferroni correction across all PGMs at that checkpoint. The headline does not depend on that choice: at iteration~10, $12$ cells are significant uncorrected and $11$ after Holm, but the set favouring neural-init is the same $10$ rows either way, so correcting removes only a cell that favoured zero-init. Rows are in a fixed canonical order, not sorted by effect size; every PGM evaluated is listed, including those where neural loses.}
\label{tab:paired_map}
\resizebox{\textwidth}{!}{%
\begin{tabular}{@{}l|ccccc@{}}
\toprule
\textbf{PGM} & \multicolumn{1}{c}{iter=0} & \multicolumn{1}{c}{iter=10} & \multicolumn{1}{c}{iter=50} & \multicolumn{1}{c}{iter=100} & \multicolumn{1}{c}{iter=500} \\
 & median $\delta$ [95\% CI] & median $\delta$ [95\% CI] & median $\delta$ [95\% CI] & median $\delta$ [95\% CI] & median $\delta$ [95\% CI] \\
\midrule
BN\_12 & 0.105$^{\dagger}$ {\tiny[0.094, 0.112]} & 0.230$^{\dagger}$ {\tiny[0.212, 0.258]} & 0.206$^{\dagger}$ {\tiny[0.180, 0.240]} & 0.152$^{\dagger}$ {\tiny[0.122, 0.193]} & -0.000 {\tiny[-0.000, -0.000]} \\
BN\_13 & 0.103$^{\dagger}$ {\tiny[0.077, 0.112]} & 0.244$^{\dagger}$ {\tiny[0.231, 0.273]} & 0.129$^{\dagger}$ {\tiny[0.099, 0.186]} & 0.062$^{\dagger}$ {\tiny[0.035, 0.093]} & -0.000$^{\dagger}$ {\tiny[-0.000, -0.000]} \\
BN\_14 & -0.039$^{\dagger}$ {\tiny[-0.082, -0.026]} & 0.200$^{\dagger}$ {\tiny[0.157, 0.232]} & 0.104$^{\dagger}$ {\tiny[0.080, 0.151]} & -0.000 {\tiny[-0.020, 0.040]} & -0.000$^{\dagger}$ {\tiny[-0.000, -0.000]} \\
BN\_9 & -0.081$^{\dagger}$ {\tiny[-0.124, -0.052]} & 0.070 {\tiny[-0.014, 0.100]} & -0.914$^{\dagger}$ {\tiny[-1.354, -0.540]} & -0.002$^{\dagger}$ {\tiny[-0.006, -0.001]} & -0.000$^{\dagger}$ {\tiny[-0.000, -0.000]} \\
\midrule
Grids\_26 & 0.008$^{\dagger}$ {\tiny[0.001, 0.024]} & 0.000 {\tiny[-0.001, 0.002]} & -1.121$^{\dagger}$ {\tiny[-2.928, -0.022]} & -1.902$^{\dagger}$ {\tiny[-2.264, -1.362]} & -0.028$^{\dagger}$ {\tiny[-0.054, -0.015]} \\
Grids\_28 & 0.007$^{\dagger}$ {\tiny[0.003, 0.021]} & -0.000 {\tiny[-0.002, 0.000]} & -1.501$^{\dagger}$ {\tiny[-3.758, -0.016]} & -2.484$^{\dagger}$ {\tiny[-2.933, -2.243]} & -0.025$^{\dagger}$ {\tiny[-0.044, -0.013]} \\
Grids\_29 & -0.007 {\tiny[-0.012, 0.001]} & 0.010$^{\dagger}$ {\tiny[0.005, 0.013]} & -0.095$^{\dagger}$ {\tiny[-2.110, -0.000]} & -1.496$^{\dagger}$ {\tiny[-2.142, -1.132]} & -0.009$^{\dagger}$ {\tiny[-0.017, -0.004]} \\
\midrule
Segmentation\_12 & 0.003 {\tiny[-0.015, 0.025]} & 0.180$^{\dagger}$ {\tiny[0.094, 0.206]} & -0.216$^{\dagger}$ {\tiny[-1.164, -0.038]} & -0.099$^{\dagger}$ {\tiny[-0.154, -0.056]} & -0.000$^{\dagger}$ {\tiny[-0.000, -0.000]} \\
Segmentation\_14 & 0.278$^{\dagger}$ {\tiny[0.202, 0.322]} & 0.932$^{\dagger}$ {\tiny[0.754, 1.094]} & -3.497$^{\dagger}$ {\tiny[-3.878, -3.111]} & -1.888$^{\dagger}$ {\tiny[-2.190, -1.434]} & -0.000$^{\dagger}$ {\tiny[-0.000, -0.000]} \\
Segmentation\_16 & 0.445$^{\dagger}$ {\tiny[0.410, 0.495]} & 0.783$^{\dagger}$ {\tiny[0.696, 0.903]} & 2.251$^{\dagger}$ {\tiny[1.690, 2.561]} & -0.765 {\tiny[-1.335, -0.101]} & -0.000$^{\dagger}$ {\tiny[-0.001, -0.000]} \\
Segmentation\_18 & 0.024 {\tiny[-0.007, 0.081]} & 0.180$^{\dagger}$ {\tiny[0.148, 0.199]} & -0.280$^{\dagger}$ {\tiny[-1.106, -0.029]} & -0.177$^{\dagger}$ {\tiny[-0.536, -0.053]} & -0.000$^{\dagger}$ {\tiny[-0.000, -0.000]} \\
Segmentation\_19 & -0.279$^{\dagger}$ {\tiny[-0.387, -0.200]} & 0.027 {\tiny[-0.068, 0.108]} & -2.489$^{\dagger}$ {\tiny[-3.009, -2.113]} & -1.292$^{\dagger}$ {\tiny[-1.733, -0.923]} & -0.000$^{\dagger}$ {\tiny[-0.000, -0.000]} \\
\midrule
grid10x10.f5.wrap & -0.039$^{\dagger}$ {\tiny[-0.047, -0.024]} & 0.003 {\tiny[0.002, 0.005]} & -3.835$^{\dagger}$ {\tiny[-4.370, -3.235]} & -1.084$^{\dagger}$ {\tiny[-1.900, -0.151]} & 0.000$^{\dagger}$ {\tiny[-0.000, 0.000]} \\
grid10x10.f10 & -0.262$^{\dagger}$ {\tiny[-7.948, -0.090]} & -0.060$^{\dagger}$ {\tiny[-5.370, -0.016]} & 0.000$^{\dagger}$ {\tiny[-0.007, 0.000]} & -0.104$^{\dagger}$ {\tiny[-2.374, 0.000]} & 0.000$^{\dagger}$ {\tiny[0.000, 0.000]} \\
grid20x20.f5.wrap & -0.007 {\tiny[-0.013, 0.000]} & 0.010$^{\dagger}$ {\tiny[0.005, 0.013]} & -0.119$^{\dagger}$ {\tiny[-1.850, -0.000]} & -1.816$^{\dagger}$ {\tiny[-2.221, -1.187]} & -0.010$^{\dagger}$ {\tiny[-0.017, -0.004]} \\
grid20x20.f10 & 0.227$^{\dagger}$ {\tiny[0.218, 0.239]} & 0.064$^{\dagger}$ {\tiny[0.060, 0.067]} & 0.023$^{\dagger}$ {\tiny[0.016, 0.030]} & 0.013$^{\dagger}$ {\tiny[0.004, 0.022]} & 0.003 {\tiny[-0.008, 0.013]} \\
\bottomrule
\end{tabular}}
\end{table*}

\begin{table*}[t]\centering\scriptsize
\caption{\textbf{Per-seed win rate of neural-init over zero-init JGLP}, with exact Clopper--Pearson 95\% intervals. A win is $\mathrm{gap}_N \le \mathrm{gap}_J + 10^{-9}$. $n$ is the number of seeds actually available per cell. Rows in fixed canonical order.}
\label{tab:paired_map_win}
\resizebox{\textwidth}{!}{%
\begin{tabular}{@{}lr|ccccc@{}}
\toprule
\textbf{PGM} & $n$ & iter=0 & iter=10 & iter=50 & iter=100 & iter=500 \\
\midrule
BN\_12 & 100 & 97\% {\tiny[91, 99]} & 100\% {\tiny[96, 100]} & 91\% {\tiny[84, 96]} & 93\% {\tiny[86, 97]} & 96\% {\tiny[90, 99]} \\
BN\_13 & 100 & 89\% {\tiny[81, 94]} & 100\% {\tiny[96, 100]} & 90\% {\tiny[82, 95]} & 72\% {\tiny[62, 81]} & 84\% {\tiny[75, 91]} \\
BN\_14 & 100 & 31\% {\tiny[22, 41]} & 96\% {\tiny[90, 99]} & 78\% {\tiny[69, 86]} & 54\% {\tiny[44, 64]} & 96\% {\tiny[90, 99]} \\
BN\_9 & 100 & 28\% {\tiny[19, 38]} & 59\% {\tiny[49, 69]} & 13\% {\tiny[7, 21]} & 79\% {\tiny[70, 87]} & 100\% {\tiny[96, 100]} \\
\midrule
Grids\_26 & 100 & 62\% {\tiny[52, 72]} & 54\% {\tiny[44, 64]} & 25\% {\tiny[17, 35]} & 12\% {\tiny[6, 20]} & 97\% {\tiny[91, 99]} \\
Grids\_28 & 100 & 62\% {\tiny[52, 72]} & 43\% {\tiny[33, 53]} & 24\% {\tiny[16, 34]} & 7\% {\tiny[3, 14]} & 97\% {\tiny[91, 99]} \\
Grids\_29 & 100 & 42\% {\tiny[32, 52]} & 76\% {\tiny[66, 84]} & 39\% {\tiny[29, 49]} & 12\% {\tiny[6, 20]} & 96\% {\tiny[90, 99]} \\
\midrule
Segmentation\_12 & 100 & 53\% {\tiny[43, 63]} & 86\% {\tiny[78, 92]} & 18\% {\tiny[11, 27]} & 26\% {\tiny[18, 36]} & 99\% {\tiny[95, 100]} \\
Segmentation\_14 & 100 & 96\% {\tiny[90, 99]} & 97\% {\tiny[91, 99]} & 6\% {\tiny[2, 13]} & 0\% {\tiny[0, 4]} & 100\% {\tiny[96, 100]} \\
Segmentation\_16 & 100 & 100\% {\tiny[96, 100]} & 100\% {\tiny[96, 100]} & 86\% {\tiny[78, 92]} & 39\% {\tiny[29, 49]} & 100\% {\tiny[96, 100]} \\
Segmentation\_18 & 99 & 59\% {\tiny[48, 68]} & 85\% {\tiny[76, 91]} & 33\% {\tiny[24, 44]} & 5\% {\tiny[2, 11]} & 100\% {\tiny[96, 100]} \\
Segmentation\_19 & 99 & 12\% {\tiny[6, 20]} & 55\% {\tiny[44, 65]} & 20\% {\tiny[13, 29]} & 7\% {\tiny[3, 14]} & 98\% {\tiny[93, 100]} \\
\midrule
grid10x10.f5.wrap & 100 & 19\% {\tiny[12, 28]} & 67\% {\tiny[57, 76]} & 16\% {\tiny[9, 25]} & 56\% {\tiny[46, 66]} & 100\% {\tiny[96, 100]} \\
grid10x10.f10 & 100 & 0\% {\tiny[0, 4]} & 31\% {\tiny[22, 41]} & 56\% {\tiny[46, 66]} & 51\% {\tiny[41, 61]} & 99\% {\tiny[95, 100]} \\
grid20x20.f5.wrap & 100 & 42\% {\tiny[32, 52]} & 77\% {\tiny[68, 85]} & 38\% {\tiny[28, 48]} & 12\% {\tiny[6, 20]} & 96\% {\tiny[90, 99]} \\
grid20x20.f10 & 100 & 100\% {\tiny[96, 100]} & 100\% {\tiny[96, 100]} & 79\% {\tiny[70, 87]} & 67\% {\tiny[57, 76]} & 54\% {\tiny[44, 64]} \\
\bottomrule
\end{tabular}}
\end{table*}

\paragraph{Sample-size ceiling.} With $n$ paired seeds and no ties, the
smallest attainable two-sided exact Wilcoxon $p$-value is $2^{1-n}$, obtained
when every seed points the same way. At $n=10$ this floor is $0.00195$
regardless of effect size, so Holm correction over $m$ tests could reach
$\alpha = 0.05$ only if $m < 26$: results were reportable on the $15$-PGM
common set but \emph{structurally} unattainable on the full suite, and absence
of significance there was a design limit rather than a null result. We
therefore expanded the MAP evaluation to $n = 100$ seeds per PGM, which lowers
the floor to $1.6 \times 10^{-30}$ and removes the constraint entirely at any
family size we report.

\paragraph{Evaluation corrections.} The numbers in this subsection come from a
re-run of the MAP evaluation that fixes two defects in the original pipeline.
First, the JGLP trace terminated as soon as a \emph{single} iteration changed
$D$ by less than $10^{-8}$. Because the dual is piecewise linear and has flat
steps, this fired at iteration~1 on $250$ of the $1{,}498$ zero-initialized runs
of the $15$-row suite as then constituted, and on $0$ of the same $1{,}498$
neural-initialized runs, truncating the baseline but not the
method at every checkpoint from iter${=}10$ onward. Both arms now run the full
iteration budget. Second, conditioning on evidence accumulated a constant that
was added to only one entry of a dummy unary factor, so a $\max$ silently
discarded negative offsets; this perturbed the conditioned model rather than
merely shifting it. Both arms are affected identically by these corrections,
and all MAP results reported here use the corrected pipeline at $n = 100$ seeds
($1{,}498$ of $1{,}500$ instances; Segmentation\_18 seed $177$ and
Segmentation\_19 seed $149$ fail deterministically in the LP solver and are
reported rather than dropped). The gated-initialization analysis below is the
one exception: it was not re-run and remains at $n = 50$ ($749$ instances).

\subsection{Paired Per-Seed Statistics for CMPE}
\label{app:cmpe_paired}

The MAP results carry a paired per-seed test (App.~\ref{app:paired}); this is
its CMPE counterpart, so that the constrained results are held to the same
standard rather than resting on means and standard deviations alone. The
comparison tested is the one the CMPE story turns on: $\lambda$-continuity
(M4) against the learned prior with test-time adaptation (M3).

The unit of replication needs stating, because it is easy to overcount here. A
cell of Tables~\ref{tab:t3_main_singlepass}--\ref{tab:t3_main_crossmu} aggregates $100$ $\mu$-steps $\times$ the
seeds, but the $\mu$-steps within a seed are successive points along one
optimization trajectory, not independent draws --- consecutive steps share the
same evidence, the same join graph and very nearly the same multiplier. Treating
them as independent would inflate $n$ by two orders of magnitude and make
almost any difference significant. We therefore reduce each seed to a single
median $|$gap$|$ per checkpoint first, and test across seeds.

Table~\ref{tab:cmpe_paired} reports the result. The per-instance picture
matches the aggregate one: the sign of the effect splits by family rather than
being uniform, with $\lambda$-continuity ahead on the Bayesian networks and
the learned prior ahead on the grids and segmentations, and the Holm-corrected
significance follows that split rather than cutting across it.

\begin{table*}[t]\centering\scriptsize
\caption{\textbf{Paired per-seed comparison for CMPE}, $\lambda$-continuity (M4) against the learned prior with test-time adaptation (M3). Each cell is the median over evidence seeds of $\delta_s = \log_{10}\mathrm{gap}_{\mathrm{M4},s} - \log_{10}\mathrm{gap}_{\mathrm{M3},s}$, so \emph{positive favours M3} and one unit is one order of magnitude, with a bootstrap $95\%$ CI. $^{\dagger}$ marks cells significant at $\alpha=0.05$ after Holm--Bonferroni correction across the PGMs of that $K$ and iteration. The independent replicate is the evidence seed: the $100$ $\mu$-steps within a seed are successive points on one trajectory, not independent draws, so each seed is reduced to one median $|$gap$|$ before testing and $n=50$ is the seed count. The split is by family rather than uniform --- $\lambda$-continuity wins on the Bayesian networks, the learned prior on grids and segmentations --- which is the point Tables~\ref{tab:t3_main_singlepass}--\ref{tab:t3_main_crossmu} make and this test confirms per-instance.}
\label{tab:cmpe_paired}
\resizebox{\textwidth}{!}{%
\begin{tabular}{@{}ll|ccc@{}}
\toprule
\textbf{PGM} & $K$ & iter=10 & iter=50 & iter=100 \\
 & & median $\delta$ [95\% CI] & median $\delta$ [95\% CI] & median $\delta$ [95\% CI] \\
\midrule
\multicolumn{5}{@{}l}{\emph{$K=1$}} \\
BN\_12 & 1 & -3.06$^{\dagger}$ {\tiny[-3.54, -1.37]} & -1.00$^{\dagger}$ {\tiny[-1.61, -0.47]} & -0.15$^{\dagger}$ {\tiny[-0.56, -0.05]} \\
BN\_13 & 1 & -3.91$^{\dagger}$ {\tiny[-4.54, -3.43]} & -1.95$^{\dagger}$ {\tiny[-2.85, -0.58]} & -0.47$^{\dagger}$ {\tiny[-1.04, -0.26]} \\
BN\_14 & 1 & -3.86$^{\dagger}$ {\tiny[-4.40, -3.26]} & -1.71$^{\dagger}$ {\tiny[-2.81, -0.51]} & -0.31$^{\dagger}$ {\tiny[-1.23, -0.10]} \\
BN\_9 & 1 & -3.91$^{\dagger}$ {\tiny[-4.40, -3.37]} & -1.33$^{\dagger}$ {\tiny[-1.80, -0.75]} & -0.11$^{\dagger}$ {\tiny[-0.26, -0.04]} \\
Grids\_26 & 1 & 3.21$^{\dagger}$ {\tiny[3.17, 3.34]} & 3.30$^{\dagger}$ {\tiny[3.07, 3.45]} & 2.82$^{\dagger}$ {\tiny[2.44, 3.81]} \\
Grids\_28 & 1 & 3.27$^{\dagger}$ {\tiny[3.04, 3.35]} & 3.28$^{\dagger}$ {\tiny[3.09, 3.45]} & 3.37$^{\dagger}$ {\tiny[2.66, 3.81]} \\
Grids\_29 & 1 & 3.09$^{\dagger}$ {\tiny[3.02, 3.19]} & 2.81$^{\dagger}$ {\tiny[2.66, 2.99]} & 2.27$^{\dagger}$ {\tiny[1.73, 3.04]} \\
Seg\_12 & 1 & 3.20$^{\dagger}$ {\tiny[3.07, 3.27]} & 4.34$^{\dagger}$ {\tiny[4.09, 4.60]} & 1.72$^{\dagger}$ {\tiny[1.28, 2.59]} \\
Seg\_14 & 1 & 2.67$^{\dagger}$ {\tiny[2.59, 2.74]} & 2.87$^{\dagger}$ {\tiny[2.69, 3.03]} & 1.52$^{\dagger}$ {\tiny[1.35, 1.63]} \\
Seg\_16 & 1 & 2.43$^{\dagger}$ {\tiny[2.31, 2.56]} & 2.66$^{\dagger}$ {\tiny[2.40, 2.80]} & 1.43$^{\dagger}$ {\tiny[1.23, 1.58]} \\
Seg\_18 & 1 & 1.97$^{\dagger}$ {\tiny[1.76, 2.12]} & 2.33$^{\dagger}$ {\tiny[2.04, 2.77]} & 3.72$^{\dagger}$ {\tiny[3.02, 4.14]} \\
Seg\_19 & 1 & 2.41$^{\dagger}$ {\tiny[2.26, 2.50]} & 2.89$^{\dagger}$ {\tiny[2.68, 3.00]} & 2.31$^{\dagger}$ {\tiny[2.20, 2.58]} \\
grid10$\times$10.f5 & 1 & 2.85$^{\dagger}$ {\tiny[2.69, 3.01]} & 2.39$^{\dagger}$ {\tiny[1.95, 2.64]} & 1.49$^{\dagger}$ {\tiny[0.88, 2.14]} \\
grid20x20.f10 & 1 & 1.77$^{\dagger}$ {\tiny[1.64, 1.92]} & 1.23$^{\dagger}$ {\tiny[1.08, 1.39]} & 0.87$^{\dagger}$ {\tiny[0.76, 1.08]} \\
grid20$\times$20.f5 & 1 & 3.09$^{\dagger}$ {\tiny[3.02, 3.20]} & 2.85$^{\dagger}$ {\tiny[2.61, 3.04]} & 2.29$^{\dagger}$ {\tiny[1.66, 3.07]} \\
grid40$\times$40.f10 & 1 & 4.32$^{\dagger}$ {\tiny[4.10, 4.45]} & 2.74$^{\dagger}$ {\tiny[2.37, 3.21]} & 1.08$^{\dagger}$ {\tiny[0.60, 1.76]} \\
grid40$\times$40.f2 & 1 & 4.43$^{\dagger}$ {\tiny[4.24, 4.63]} & 2.32$^{\dagger}$ {\tiny[1.68, 2.98]} & 0.24 {\tiny[0.05, 0.44]} \\
\midrule
\multicolumn{5}{@{}l}{\emph{$K=2$}} \\
BN\_12 & 2 & 0.16 {\tiny[-0.02, 0.30]} & -0.03 {\tiny[-0.16, 0.06]} & -0.10 {\tiny[-0.30, 0.01]} \\
BN\_13 & 2 & 0.31 {\tiny[0.11, 0.46]} & -0.02 {\tiny[-0.14, 0.16]} & -0.04 {\tiny[-0.11, -0.00]} \\
BN\_14 & 2 & 1.15$^{\dagger}$ {\tiny[0.73, 1.52]} & 0.12 {\tiny[-0.13, 1.05]} & -0.00 {\tiny[-0.01, 0.09]} \\
BN\_9 & 2 & 0.60 {\tiny[0.09, 0.89]} & -0.23 {\tiny[-0.55, 0.12]} & -0.04 {\tiny[-0.25, -0.01]} \\
Grids\_26 & 2 & 3.35$^{\dagger}$ {\tiny[3.00, 3.45]} & 3.33$^{\dagger}$ {\tiny[3.13, 3.59]} & 3.60$^{\dagger}$ {\tiny[3.21, 4.02]} \\
Grids\_28 & 2 & 1.90$^{\dagger}$ {\tiny[1.53, 2.26]} & 3.32$^{\dagger}$ {\tiny[2.85, 3.49]} & 3.76$^{\dagger}$ {\tiny[3.35, 4.09]} \\
Grids\_29 & 2 & 3.23$^{\dagger}$ {\tiny[3.21, 3.28]} & 3.13$^{\dagger}$ {\tiny[3.01, 3.37]} & 2.91$^{\dagger}$ {\tiny[2.24, 3.70]} \\
Seg\_12 & 2 & 3.44$^{\dagger}$ {\tiny[3.36, 3.55]} & 4.85$^{\dagger}$ {\tiny[4.60, 5.11]} & 4.61$^{\dagger}$ {\tiny[4.09, 5.13]} \\
Seg\_14 & 2 & 3.66$^{\dagger}$ {\tiny[3.53, 3.70]} & 4.07$^{\dagger}$ {\tiny[3.87, 4.21]} & 2.23$^{\dagger}$ {\tiny[1.91, 2.41]} \\
Seg\_16 & 2 & 3.21$^{\dagger}$ {\tiny[3.16, 3.25]} & 3.58$^{\dagger}$ {\tiny[3.45, 3.70]} & 2.08$^{\dagger}$ {\tiny[1.70, 3.05]} \\
Seg\_18 & 2 & 3.00$^{\dagger}$ {\tiny[2.92, 3.07]} & 3.28$^{\dagger}$ {\tiny[2.99, 3.51]} & 2.34$^{\dagger}$ {\tiny[1.74, 3.35]} \\
Seg\_19 & 2 & 3.30$^{\dagger}$ {\tiny[3.27, 3.37]} & 3.80$^{\dagger}$ {\tiny[3.63, 3.90]} & 2.68$^{\dagger}$ {\tiny[2.47, 3.76]} \\
grid10$\times$10.f5 & 2 & 3.24$^{\dagger}$ {\tiny[3.11, 3.36]} & 2.83$^{\dagger}$ {\tiny[2.59, 3.21]} & 2.00$^{\dagger}$ {\tiny[1.66, 2.24]} \\
grid20x20.f10 & 2 & 1.43$^{\dagger}$ {\tiny[1.35, 1.49]} & 0.84$^{\dagger}$ {\tiny[0.78, 0.89]} & 0.51$^{\dagger}$ {\tiny[0.46, 0.58]} \\
grid20$\times$20.f5 & 2 & 3.24$^{\dagger}$ {\tiny[3.20, 3.29]} & 3.17$^{\dagger}$ {\tiny[3.04, 3.35]} & 2.96$^{\dagger}$ {\tiny[2.20, 3.70]} \\
\bottomrule\end{tabular}}
\end{table*}

\subsection{Gated Initialization}
\label{app:gated}

Both $D(\boldsymbol\theta, \mathbf{0})$ and $D(\boldsymbol\theta, A\mathbf{m})$
are valid upper bounds on $\mathrm{MAP}^*$, so initializing JGLP from whichever
is lower is also valid. We report this gated variant as a third arm alongside
zero-init and neural-init; the neural arm is kept, not replaced, so the gate's
effect remains visible. Because JGLP is deterministic given its starting point
and the gate's starting point is one of the two arms already evaluated, the
gated trajectory is obtained without additional inference.

The gate selects the neural start on $56\%$ of the $1{,}598$ (PGM, seed)
instances of the $100$-seed run, ranging from $0\%$ on grid10$\times$10.f10 to
$100\%$ on Segmentation\_16 and grid20$\times$20.f10: it declines the neural
initialization precisely where that initialization is worse. \textbf{The
guarantee it provides is confined to iteration~0}, where the gated bound is
never worse than zero-init by construction --- and is so on $0$ of $1{,}598$
instances, as it must be. At later checkpoints it is sometimes worse --- on $73$
of $1{,}598$ seeds at iteration~10, $527$ at iteration~100 and $69$ at
iteration~500 --- because a tighter starting
point does not imply a faster coordinate-descent trajectory. We therefore claim
only the iteration-0 property. Table~\ref{tab:gated_init} reports the sweep.

\begin{table*}[t]\centering\scriptsize
\caption{\textbf{Gated initialization.} Both $D(\boldsymbol\theta,\mathbf{0})$ and $D(\boldsymbol\theta, A\mathbf{m})$ are valid upper bounds, so starting JGLP from whichever is lower is valid and, by construction, never worse than zero-init at iteration~0. ``Gate'' is the fraction of seeds on which the neural start was selected. Remaining columns give the median dual gap for zero-init (J), neural-init (N) and the gated variant, and the number of seeds on which the gated variant is strictly worse than J. Every PGM evaluated is listed, including those where the neural arm loses and the gate correctly declines it.}
\label{tab:gated_init}
\resizebox{\textwidth}{!}{%
\begin{tabular}{@{}lr|rrr|rrr@{}}
\toprule
\textbf{PGM} & Gate & \multicolumn{3}{c|}{median gap at iter=10} & \multicolumn{3}{c}{gated worse than J (\#seeds)} \\
 & fires & J & N & gated & it=10 & it=100 & it=500 \\
\midrule
BN\_12 & 97\% & 0.54 & 0.299 & 0.299 & 0 & 7 & 4 \\
BN\_13 & 89\% & 0.449 & 0.249 & 0.26 & 0 & 24 & 12 \\
BN\_14 & 31\% & 0.3 & 0.169 & 0.232 & 0 & 15 & 0 \\
BN\_9 & 28\% & 0.0291 & 0.0161 & 0.0214 & 7 & 5 & 0 \\
\midrule
Grids\_26 & 62\% & 2.21 & 2.22 & 2.21 & 20 & 52 & 3 \\
Grids\_28 & 62\% & 4.03 & 4.12 & 4.12 & 27 & 57 & 3 \\
Grids\_29 & 42\% & 1.06 & 1.03 & 1.05 & 5 & 35 & 0 \\
\midrule
Segmentation\_12 & 53\% & 1.06 & 0.793 & 0.858 & 6 & 34 & 1 \\
Segmentation\_14 & 96\% & 0.201 & 0.0185 & 0.0185 & 1 & 96 & 0 \\
Segmentation\_16 & 100\% & 1.85 & 0.311 & 0.311 & 0 & 61 & 0 \\
Segmentation\_18 & 59\% & 0.922 & 0.521 & 0.529 & 0 & 54 & 0 \\
Segmentation\_19 & 12\% & 0.0676 & 0.0324 & 0.051 & 2 & 12 & 0 \\
\midrule
grid10x10.f5.wrap & 19\% & 0.193 & 0.193 & 0.193 & 0 & 7 & 0 \\
grid10x10.f10 & 0\% & 0.092 & 0.096 & 0.092 & 0 & 0 & 0 \\
grid20x20.f5.wrap & 42\% & 1.06 & 1.03 & 1.05 & 5 & 35 & 0 \\
grid20x20.f10 & 100\% & 14 & 11.8 & 11.8 & 0 & 33 & 46 \\
\bottomrule
\end{tabular}}
\end{table*}

\subsection{Architecture and Training-Set-Size Sweeps}
\label{app:p7}

Tables~\ref{tab:p7_arch} and~\ref{tab:p7_trainsize} sweep the network
architecture and the number of distinct training evidence configurations on one
representative PGM from each of three families. Across an $18\times$ span in
parameter count the iteration-10 improvement varies by at most $0.385$ in
$\log_{10}$ units within a PGM, while varying by more than a full order of
magnitude \emph{between} PGMs ($+0.01$ on Grids\_29 versus $+1.25$ on
Segmentation\_14): the benchmark, not the architecture, determines whether the
warm-start helps. The training-set-size curves are flat, with $250$
configurations already providing essentially the full benefit.

Two details in Table~\ref{tab:p7_arch} cut against our own configuration and we
state them rather than leave them to be noticed. On all three families the best
cell is \emph{smaller} than the $512/512/256$ default we report results with
--- BN\_13 peaks at depth~$2$, width~$256$ ($+0.358$ against the default's
$+0.290$), Grids\_29 at depth~$3$, width~$128$ ($+0.023$ against $+0.010$), and
Segmentation\_14 at depth~$3$, width~$256$ ($+1.254$ against $+1.124$) --- and
the widest networks ($w{=}1024$) are never best on any family. A smaller and
faster network would therefore have served at least as well; we did not tune
the architecture, and the numbers say we did not need to.

\begin{table*}[t]\centering\scriptsize
\caption{\textbf{Architecture sweep.} Depth $\times$ width, output dimension fixed by $\mathrm{edim}$. Each cell is the median over 30 held-out evidence seeds of the paired iteration-10 improvement $\log_{10}\mathrm{gap}_{J} - \log_{10}\mathrm{gap}_{N}$ (positive favours the network). \emph{Caveat:} every configuration receives the same 500 gradient steps, so a wide model is not trained to the same degree of convergence as a narrow one; differences across width confound capacity with optimization budget and should not be read as a capacity ceiling. Three families are covered, one representative PGM each; Promedas is excluded because the corrected evidence conditioning yields a structurally different join graph there, so its existing checkpoints are not reusable.}
\label{tab:p7_arch}
\resizebox{\textwidth}{!}{%
\begin{tabular}{@{}ll|rrrr|rr@{}}
\toprule
\textbf{PGM} & depth & w=128 & w=256 & w=512 & w=1024 & params (w=512) & fwd (ms) \\
\midrule
BN\_13 & 2 & +0.346 & +0.358 & +0.330 & +0.332 & 1.42M & 0.64 \\
 & 3 & +0.322 & +0.316 & +0.290 & +0.294 & 1.01M & 0.66 \\
 & 4 & +0.271 & +0.259 & +0.262 & +0.267 & 1.07M & 0.70 \\
\midrule
Grids\_29 & 2 & +0.022 & +0.022 & +0.019 & +0.015 & 2.00M & 0.76 \\
 & 3 & +0.023 & +0.013 & +0.010 & +0.011 & 1.37M & 0.75 \\
 & 4 & +0.018 & +0.013 & +0.011 & +0.010 & 1.44M & 0.97 \\
\midrule
Segmentation\_14 & 2 & +0.870 & +1.154 & +1.138 & +0.981 & 1.95M & 0.71 \\
 & 3 & +1.186 & +1.254 & +1.124 & +1.099 & 1.30M & 0.78 \\
 & 4 & +1.215 & +1.197 & +1.198 & +1.111 & 1.36M & 0.81 \\
\bottomrule
\end{tabular}}
\end{table*}

\begin{table*}[t]\centering\scriptsize
\caption{\textbf{Training-set size sweep.} Number of distinct evidence configurations in the training pool, with architecture and the number of gradient steps held fixed (500), so set size is the only quantity varying. Upper block: held-out standalone gap $D(\boldsymbol\theta, A\mathbf{m}) - D^{\mathrm{LP}}$ from the raw forward pass (lower is tighter). Lower block: paired iteration-10 improvement over zero-init (positive favours the network). Median over 30 held-out seeds. The curves are flat: 250 configurations already provide essentially the full benefit. Table~\ref{tab:v5_knee} extends the sweep below 250 and locates the plateau directly rather than bounding it.}
\label{tab:p7_trainsize}
\resizebox{\textwidth}{!}{%
\begin{tabular}{@{}l|rrrrrrr@{}}
\toprule
\textbf{PGM} & 250 & 500 & 1000 & 2000 & 4000 & 8000 & 16000 \\
\midrule
\multicolumn{8}{@{}l}{\emph{held-out standalone gap}} \\
BN\_13 & 0.694 & 0.6744 & 0.6815 & 0.6646 & 0.6664 & 0.6648 & 0.6724 \\
Grids\_29 & 7.5 & 7.457 & 7.424 & 7.387 & 7.374 & 7.423 & 7.417 \\
Segmentation\_14 & 0.4442 & 0.4148 & 0.4084 & 0.4114 & 0.4361 & 0.4842 & 0.4588 \\
\midrule
\multicolumn{8}{@{}l}{\emph{iteration-10 paired improvement}} \\
BN\_13 & +0.318 & +0.305 & +0.312 & +0.291 & +0.305 & +0.304 & +0.301 \\
Grids\_29 & +0.018 & +0.010 & +0.010 & +0.012 & +0.012 & +0.010 & +0.009 \\
Segmentation\_14 & +1.124 & +0.966 & +1.129 & +1.152 & +1.210 & +1.207 & +1.273 \\
\bottomrule
\end{tabular}}
\end{table*}

The sweep above starts at $N=250$ and is flat across a $64\times$ range, which
locates the plateau only as \emph{at or below} its smallest point. Two follow-up
sweeps close that gap. Table~\ref{tab:v5_knee} extends the same protocol down to
$N=10$ on the same three models; Table~\ref{tab:v5_datasize} widens it to nine
models on the original grid, to test whether a model's appetite for data tracks
how much the warm start helps it. The answers separate by metric: the
\emph{standalone} bound has a real knee (at $N=150$--$250$), whereas the
\emph{iteration-10 improvement} is already saturated at $N=10$, and on two of the
three models it approaches its asymptote from above, so no knee exists. Across
the nine models, sensitivity to training-set size does not track the benefit the
warm start delivers (Spearman $\rho = +0.02$, $p = 0.97$). The honest reading is
early saturation on data, on every model tested.

Training cost is unaffected by the training-set size, which completes the
answer: because the number of gradient steps is held fixed at 500, wall-clock
across $N = 10$ to $N = 250$ varies by only $2\%$ (BN\_13, $147$--$150$\,s),
$2\%$ (Segmentation\_14, $305$--$312$\,s) and $9\%$ (Grids\_29,
$358$--$389$\,s) --- a $25\times$ change in data for no change in compute. The
binding constraint on per-graph training is therefore the gradient-step budget,
not the number of evidence configurations, and since the configurations are
generated self-supervised (no LP or JGLP solve in the loop) there is no label
budget to spend either.

\begin{table*}[t]\centering\scriptsize
\caption{\textbf{Training-set size below the P7 grid.} The sweep in Table~\ref{tab:p7_trainsize} starts at $N=250$ and is flat, which locates the plateau only as \emph{at or below} its smallest point. This extends it down to $N=10$ on the same three models, same protocol ($512/512/256$, $500$ gradient steps at every $N$), three replicates per cell differing in the drawn training pool, the batch stream and the weight init. Upper block: held-out standalone gap at iteration~0, the quantity that measures what the network itself learned. Lower block: paired iteration-10 improvement over zero-init. Median over $30$ held-out seeds, then mean over the three replicates. \emph{Knee} is the smallest $N$ within $5\%$ of the $N=16{,}000$ asymptote. The two blocks answer differently: the standalone gap falls steeply and has a locatable knee, while the iteration-10 improvement is already saturated at $N=10$ --- on BN\_13 and Grids\_29 it approaches the asymptote from \emph{above}, so no knee exists to find.}
\label{tab:v5_knee}
\resizebox{\textwidth}{!}{%
\begin{tabular}{@{}l|rrrrrr|r@{}}
\toprule
\textbf{PGM} & $N{=}10$ & $N{=}25$ & $N{=}50$ & $N{=}100$ & $N{=}150$ & $N{=}250$ & knee \\
\midrule
\multicolumn{8}{@{}l}{\emph{held-out standalone gap at iteration 0 (lower is tighter)}} \\
BN\_13 & 1.78 & 1.24 & 0.94 & 0.77 & 0.73 & 0.69 & $N{=}250$ \\
Seg\_14 & 2.89 & 1.50 & 1.04 & 0.65 & 0.54 & 0.43 & $N{=}250$ \\
Grids\_29 & 19.43 & 13.15 & 9.42 & 8.05 & 7.68 & 7.56 & $N{=}150$ \\
\midrule
\multicolumn{8}{@{}l}{\emph{paired iteration-10 improvement over zero-init, $\log_{10}$ (higher favours the network)}} \\
BN\_13 & 0.357 & 0.377 & 0.377 & 0.353 & 0.334 & 0.316 & already flat \\
Seg\_14 & 0.654 & 0.820 & 0.873 & 1.008 & 0.991 & 1.097 & already flat \\
Grids\_29 & 0.036 & 0.047 & 0.033 & 0.023 & 0.016 & 0.016 & already flat \\
\bottomrule
\end{tabular}}
\end{table*}

\begin{table*}[t]\centering\scriptsize
\caption{\textbf{Does sensitivity to training-set size track how much the warm start helps?} Paired iteration-10 improvement over zero-init ($\log_{10}$) against training-set size, on nine models spanning the full range of warm-start benefit. The six models of the first block were selected by a rule fixed and recorded \emph{before} the runs (rank every eligible model in Table~\ref{tab:table1_full} by its iteration-10 improvement, split into thirds, take the two nearest each third's median); the three of the second block are the P7 models, whose existing sweep covers the same grid. \emph{T4} is the model's improvement in Table~\ref{tab:table1_full}; \emph{range} is $\max-\min$ across the four sizes. Spearman $\rho$ between T4 and range is $+0.017$ ($p=0.97$), and dropping Seg\_14 --- the one model carrying it --- gives $-0.214$ ($p=0.61$): the curves are flat on every model, and sensitivity does \emph{not} track the benefit. grid80$\times$80.f5 is the clearest counterexample: the warm start hurts it by roughly six orders, yet its range is the largest in the table. Every model run is listed, none dropped or reordered.}
\label{tab:v5_datasize}
\resizebox{\textwidth}{!}{%
\begin{tabular}{@{}l|r|rrrr|r@{}}
\toprule
\textbf{PGM} & T4 & $N{=}250$ & $N{=}1{,}000$ & $N{=}4{,}000$ & $N{=}16{,}000$ & range \\
\midrule
Seg\_18 & +0.21 & 0.208 & 0.200 & 0.189 & 0.184 & 0.024 \\
BN\_14 & +0.17 & 0.144 & 0.146 & 0.152 & 0.152 & 0.008 \\
Grids\_23 & +0.02 & 0.035 & 0.029 & 0.031 & 0.029 & 0.006 \\
Grids\_18 & +0.01 & 0.015 & 0.018 & 0.016 & 0.017 & 0.002 \\
Seg\_19 & -2.26 & 0.086 & 0.081 & 0.091 & 0.106 & 0.025 \\
grid80$\times$80.f5 & -3.83 & -6.115 & -5.976 & -5.992 & -5.917 & 0.198 \\
\midrule
BN\_13 & +0.31 & 0.318 & 0.312 & 0.305 & 0.301 & 0.017 \\
Seg\_14 & +1.27 & 1.124 & 1.129 & 1.210 & 1.273 & 0.150 \\
Grids\_29 & +0.04 & 0.018 & 0.010 & 0.012 & 0.009 & 0.009 \\
\bottomrule
\end{tabular}}
\end{table*}

\begin{figure}[ht]
\centering
\includegraphics[width=\textwidth]{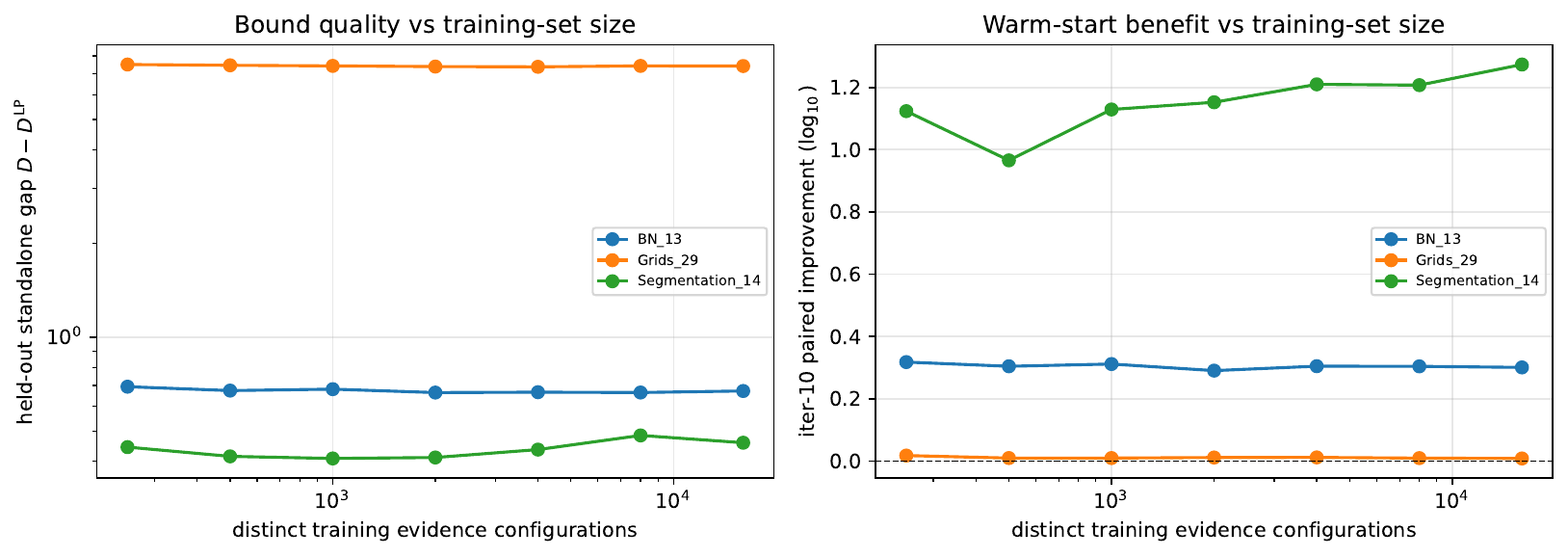}
\caption{Held-out bound quality (left) and iteration-10 paired improvement
(right) as a function of the number of distinct training evidence
configurations, with architecture and gradient-step count held fixed. Log
$x$-axis. Neither curve shows a knee above $250$ configurations.}
\label{fig:p7_learning_curve}
\end{figure}

\paragraph{Amortization gap.} Table~\ref{tab:p7_amortization} evaluates the
standalone bound on the network's \emph{own training} instances and on
held-out instances. The train gap is essentially as large as the test gap
(ratio $0.86$--$1.01$ across the three families), so the residual between the
network's bound and $D^{\mathrm{LP}}$ is not a generalization failure: the
network does not reach the optimal bound even on examples it was trained on.
This bounds what more data or better regularization can buy, and identifies the
parameterization and optimization of the message predictor---not the size of
the training set---as the limiting factor. The per-cell differences are noisy
($10$ instances per cell) and take both signs; the stable observation is that
train and test gaps coincide, not the value of any individual difference. Figure~\ref{fig:p7_learning_curve} plots the learning curves.

\begin{table}[t]\centering\scriptsize
\caption{\textbf{Amortization gap.} Standalone bound gap $D(\boldsymbol\theta, A f_\phi(\mathbf{e})) - D^{\mathrm{LP}}$ evaluated on the network's \emph{own training} instances versus held-out instances (median over 10 of each). A nonzero train gap means the network does not reach the optimal bound even on examples it was trained on; the difference between the two columns is the generalization component.}
\label{tab:p7_amortization}
\begin{tabular}{@{}lr|rrr@{}}
\toprule
\textbf{PGM} & train size & train gap & test gap & difference \\
\midrule
BN\_13 & 250 & 0.66629 & 0.72039 & +0.054108 \\
BN\_13 & 500 & 0.67301 & 0.68346 & +0.010447 \\
BN\_13 & 1000 & 0.62922 & 0.66779 & +0.038565 \\
BN\_13 & 2000 & 0.78199 & 0.68431 & -0.097674 \\
BN\_13 & 4000 & 0.62562 & 0.66762 & +0.042 \\
BN\_13 & 8000 & 0.80683 & 0.66225 & -0.14459 \\
BN\_13 & 16000 & 0.88751 & 0.66812 & -0.2194 \\
\midrule
Grids\_29 & 250 & 6.4504 & 7.5887 & +1.1383 \\
Grids\_29 & 500 & 7.1082 & 7.5449 & +0.43669 \\
Grids\_29 & 1000 & 6.4822 & 7.4777 & +0.99552 \\
Grids\_29 & 2000 & 7.8018 & 7.4802 & -0.32158 \\
Grids\_29 & 4000 & 7.2847 & 7.4539 & +0.16918 \\
Grids\_29 & 8000 & 7.7342 & 7.4877 & -0.24648 \\
Grids\_29 & 16000 & 6.3398 & 7.4623 & +1.1225 \\
\midrule
Segmentation\_14 & 250 & 0.5996 & 0.41282 & -0.18678 \\
Segmentation\_14 & 500 & 0.77982 & 0.40282 & -0.377 \\
Segmentation\_14 & 1000 & 0.3558 & 0.38301 & +0.027216 \\
Segmentation\_14 & 2000 & 0.71614 & 0.41137 & -0.30477 \\
Segmentation\_14 & 4000 & 0.34797 & 0.41848 & +0.070509 \\
Segmentation\_14 & 8000 & 0.27035 & 0.4596 & +0.18926 \\
Segmentation\_14 & 16000 & 0.31046 & 0.43529 & +0.12483 \\
\bottomrule
\end{tabular}
\end{table}

\subsection{LP-Supervised CMPE Pipeline (Alternative)}
\label{app:lp_supervised}

Tables~\ref{tab:cmpe_eps_v2}--\ref{tab:cmpe_eps_two_v2} use the no-LP pipeline above. The LP-supervised variant uses a hybrid loss $\mathcal{L} = c \cdot \mathrm{MSE}(\mathbf{m}, \mathbf{m}^{\mathrm{LP}}) + (1-c) D_\tau$ with $c=1.0$ (pure MSE against the LP-optimal edge message), and remains a viable alternative when an LP solver is already available in the user's pipeline. Empirically, it produces slightly tighter raw NN bounds at iter=0 on degenerate PGMs but is matched or exceeded by the no-LP variant on non-degenerate PGMs (Table~\ref{tab:nolp_pilot}).

\paragraph{Settings for Table~\ref{tab:cmpe_eps_two_v2} (K=2).} Both $K=1$ and
$K=2$ now run at $100$ $\mu$-gradient steps $\times$ $50$ test seeds, i.e.\
$5{,}000$ (seed, $\mu$-step) trials per cell. This table has been through three
settings and we give the progression, because the trend across them is itself
informative. Our first evaluation reported 13 of the 14 $K=2$ PGMs at a reduced
$50 \times 5$ setting, because re-running the full three-arm evaluation is
dominated by roughly a thousand LP solves per PGM. That turned out to be
avoidable: a full $100 \times 10$ three-arm run already existed, and the only
column missing at full settings was JGLP+WS, which solves no LPs and has no
learnable parameters, so computing that one column brought every PGM to
$100 \times 10$. We then raised the seed count to $50$ for more
replicates.

Each raise moved the NN+TTA column in its favour on the grid and segmentation
families while leaving JGLP+WS essentially unchanged, and the effect has now
saturated. Grids\_29's mean gap at iteration~10 went $6.4$ at $50\times5$ to
$1.8$ at $100\times10$ to $1.60$ at $100\times50$; Segmentation\_19's went
$0.89$ to $0.41$ to $0.436$. The first raise is the large one and the second
changes these cells by under $10\%$, which is what one wants from a replicate
count: it says the earlier numbers were noisy rather than wrong. This is expected
rather than surprising --- test-time
adaptation accumulates across $\mu$-steps, so doubling the trajectory doubles
the adaptation budget, and the reduced setting therefore understated the
method. One consequence is worth stating explicitly because it cuts across a
claim we have made elsewhere: at the full setting JGLP+WS still beats NN+TTA at
every budget on BN\_12, but on BN\_13 the ordering \emph{reverses} at all three
budgets (iteration~10: $0.065$ for JGLP+WS against $0.044$ for NN+TTA). The
$\lambda$-continuity advantage is therefore specific to BN\_12 among the
Bayesian networks, not a property of the family.

\subsection{Numerical Floor of the Reparameterization}
\label{app:numerics}
The reparameterization is exact in real arithmetic
but computed in floating point, and $A$ must be applied in double precision: in
single precision the antisymmetric cancellation carries a $\sim\!10^{-7}$ error,
negligible against the $10^{-2}$--$10^{0}$ gaps targeted but dominant in the
converged regime.

\subsection{Cross-Topology Transfer with a Join-Graph GNN (Negative Result)}
\label{app:gnn}
The per-graph MLP of Section~\ref{sec:method} can be replaced by a graph neural
network that operates on the join graph itself, with a per-edge readout emitting
$\mathbf{m}$; the scatter $A$ is determined entirely by the join-graph
structure, so Proposition~\ref{prop:validity} applies unchanged and such a model
is valid by construction on topologies it has never seen. Whether it is
\emph{tight} on them is a separate question, and in our experiments it was not.
We trained a $770{,}689$-parameter GNN with $6$ message-passing layers on five
grid topologies and evaluated it on held-out $80\times 80$ grids, and separately
on two segmentation topologies evaluated on a third. It improved on the
moment-matched mini-bucket start --- the quantity it must beat to be useful ---
on $0$ of $54$ held-out instances, with median excesses of $+12.7$ and $+11.7$
nats on the grids and $+7.0$ nats on Segmentation\_18, and the deficit grew with
held-out graph size, the opposite of what transfer requires. Validation loss
plateaued by epoch~6 of~150, an underfitting signature, which indicts this
architecture rather than the idea; we report it as a negative result and name
cross-topology transfer as the primary open problem.

\clearpage
\end{document}